\documentclass[Journal]{IEEEtai}
\usepackage{times}
\usepackage{epsfig}
\usepackage{graphicx}
\usepackage{amsmath}
\usepackage{amssymb}
\usepackage{mathtools}
\usepackage{multirow}
\usepackage{subfig}
\usepackage{fancyhdr}
\usepackage{soul}
\usepackage[ruled,vlined]{algorithm2e}
\usepackage{alphalph}
\usepackage{comment}
\usepackage{hyperref}
\usepackage[utf8]{inputenc}
\usepackage{tabularx}
\usepackage{booktabs}
\usepackage[table,xcdraw]{xcolor}
\usepackage{xcolor}
\usepackage{ragged2e}
\usepackage{footnote}
\usepackage{tablefootnote}
\makesavenoteenv{tabular}
\ifCLASSOPTIONcompsoc
  \usepackage[nocompress]{cite}
\else
  \usepackage{cite}
\fi

\begin{document}
\title{Progressive Update Guided Interdependent Networks for Single Image Dehazing}

\author{Aupendu~Kar\IEEEauthorrefmark{2},~\IEEEmembership{Student~Member,~IEEE,}
        Sobhan~Kanti~Dhara\IEEEauthorrefmark{2},
        Debashis~Sen,~\IEEEmembership{Senior~Member,~IEEE,}
        and~Prabir~Kumar~Biswas,~\IEEEmembership{Senior~Member,~IEEE}
\IEEEcompsocitemizethanks{\IEEEcompsocthanksitem \IEEEauthorrefmark{2}Aupendu Kar and Sobhan Kanti Dhara share equal contribution 
\IEEEcompsocthanksitem Aupendu Kar,  Debashis Sen and Prabir Kumar Biswas  are associated with the department of Electronics and  Electrical Communication Engineering, Indian Institute of Technology Kharagpur, India-721302. email: {mailtoaupendu@gmail.com, dsen@ece.iitkgp.ac.in, pkb@ece.iitkgp.ac.in. Sobhan Kanti Dhara is associated with the Department of Electronics a Communication Engineering, National Institute of Technology Rourkela, India-769008. email: {dhara.sk@gmail.com}  (Corresponding author: Sobhan Kanti Dhara.) }\protect\\}}

\markboth{Journal of IEEE Transactions on Artificial Intelligence, Vol. 00, No. 0, Month 2020}
{Kar \MakeLowercase{\textit{et al.}}: Progressive Update Guided Interdependent Networks for Single Image Dehazing}

\maketitle

\begin{abstract}
\justifying
Images with haze of different varieties often pose a significant challenge to dehazing. Therefore, guidance by estimates of haze parameters related to the variety would be beneficial, and their progressive update jointly with haze reduction will allow effective dehazing. To this end, we propose a multi-network dehazing framework containing novel interdependent dehazing and haze parameter updater networks that operate in a progressive manner. The haze parameters, transmission map and atmospheric light, are first estimated using dedicated convolutional networks that allow color-cast handling. The estimated parameters are then used to guide our dehazing module, where the estimates are progressively updated by novel convolutional networks. The updating takes place jointly with progressive dehazing using a network that invokes inter-step dependencies. The joint progressive updating and dehazing gradually modify the haze parameter values toward achieving effective dehazing. Through different studies, our dehazing framework is shown to be more effective than image-to-image mapping and predefined haze formation model based dehazing. The framework is also found capable of handling a wide variety of hazy conditions wtih different types and amounts of haze and color casts. Our dehazing framework is qualitatively and quantitatively found to outperform the state-of-the-art on synthetic and real-world hazy images of multiple datasets with varied haze conditions.
\end{abstract}

\begin{IEEEImpStatement}
Haze in the atmosphere due to several environmental conditions degrade the visibility of the captured scenes in images. Image dehazing diminishes haze in the captured scenes and improves the visibility of the scene content. As a result, it helps to achieve better performance in recreational photography, scene surveillance, autonomous driving, intelligent transportation, and many more applications. Most recent dehazing techniques produce satisfactory results in simulated hazy images but they do not usually perform well in handling a wide variety of hazy conditions, including the presence of color cast in real-world images. We propose a framework that strategically uses deep neural networks to update haze-defining parameters
and dehaze progressively so that real-world hazy image conditions are handled appropriately. Our framework implements an interdependent multi-network system, with each having dedicated roles imparting specific advantages while collectively working towards effective dehazing. Extensive studies on simulated and real-world hazy images show that our proposed framework handles a wide variety of hazy conditions with different degrees of haze and color casts, unlike the state-of-the-art.

\end{IEEEImpStatement}

\begin{IEEEkeywords}
Single Image Dehazing, Progressive Dehazing, Haze Parameter Guidance, Interdependent Networks.
\end{IEEEkeywords}

\section{Introduction}
\setlength{\textfloatsep}{0.25cm}
\label{Intro}

\IEEEPARstart{A} hazy image captured in the hazy environment suffers from obscured visibility, reduced contrast, color cast and many other degradations due to the scattering and absorption of light by fog, aerosols, sands and mists present in the atmosphere\cite{he2010single, berman2018single, 4288166, 9878218}. Such distorted images hinder the performance of several computer vision tasks. Therefore, dehazing in such cases is essential for producing images of good perceptual quality. Such enhanced quality image improves the performance of subsequent computer vision tasks on them~\cite{li2018benchmarking, 9514465, 9423576}.

\begin{figure}[ht]
    \centering
       \includegraphics[width=0.8\linewidth]{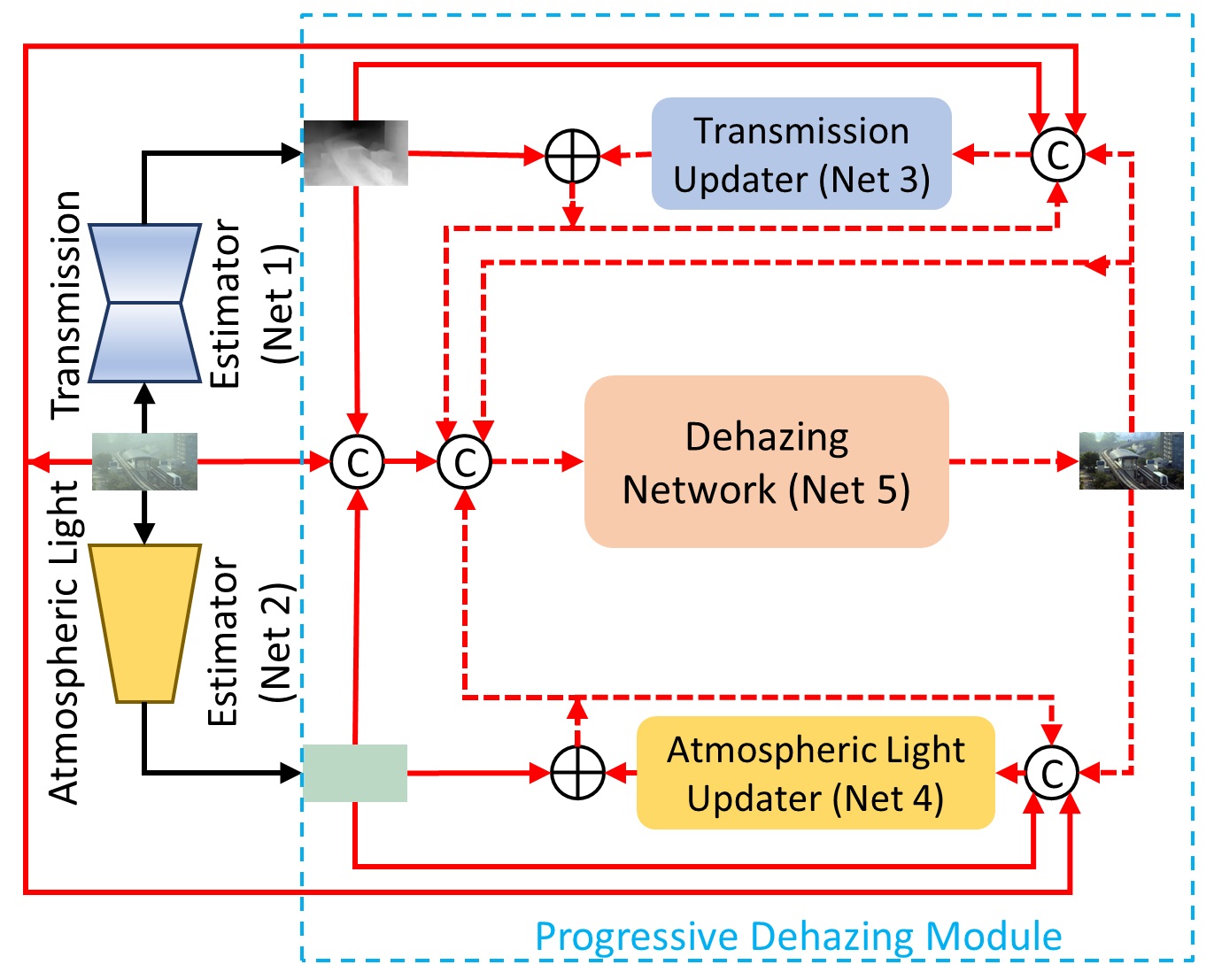}
  \caption{A schematic diagram of our proposed multi-network framework `PUG-D'  for dehazing. (Dotted red lines indicate progressively updated content)}
  \label{ga} 
\end{figure} 

Although initial works on dehazing considered multiple images of the same scene~\cite{nayar1999vision}, later, single image dehazing gained popularity, which aims at producing a dehazed image from the single hazy image at hand. While many single image dehazing techniques like~\cite{berman2018single,peng2019image, kim2019fast,8931242,bui2017single, choi2015referenceless,ancuti2013single,tan2008visibility,fattal2008single,he2010single,li2015edge,fattal2014dehazing, 9385131} are based on estimating haze parameters, namely, transmission map and atmospheric light, and using them in a predefined model, quite a few attempts have been made at end-to-end single image dehazing~\cite{Dong_2020_CVPR,8897130,8902220,qu2019enhanced, dudhane2019ryf,liu2019griddehazenet,zhang2018densely, ren2018gated,li2017aod,cai2016dehazenet}.

Accurate estimation of transmission map helps in proper dehazing reversing the effects of absorption and scattering~\cite{ 8695091}. Several approaches estimate the transmission map using different image priors /characteristics such as the dark channel prior (DCP)~\cite{he2010single}, color attenuation prior, haze-lines, etc. \cite{berman2018single,peng2019image, kim2019fast,8931242,bui2017single,li2017single, choi2015referenceless,ancuti2013single,tan2008visibility,fattal2008single,he2010single,li2015edge,fattal2014dehazing}. Further, accurate atmospheric light estimation is also crucial for recovering the appropriate illumination condition during dehazing. Many techniques estimate the atmospheric light from the bright pixels of DCP~\cite{he2010single}. In contrast, a few approaches use deep neural networks to estimate both transmission map and atmospheric light directly from the hazy image~\cite{cai2016dehazenet,zhang2018densely,liu2019learning} and use them in a predefined light scattering model to perform dehazing. 

To avoid the use of predefined models, single image dehazing frameworks based on end-to-end deep learning have been proposed lately that do not perform separate transmission map and atmospheric light estimations~\cite{Dong_2020_CVPR,8897130,8902220,qu2019enhanced, dudhane2019ryf,liu2019griddehazenet, ren2018gated,li2017aod, du2018recursive}.
By doing so they prevent the sub-optimal restoration that may occur due to the estimation of transmission map and atmospheric light disjoint from the dehazing model estimation~\cite{li2017aod}, but at the cost of not using the guidance of haze parameters.

\begin{figure}[ht]
    \centering
       \includegraphics[width=\linewidth]{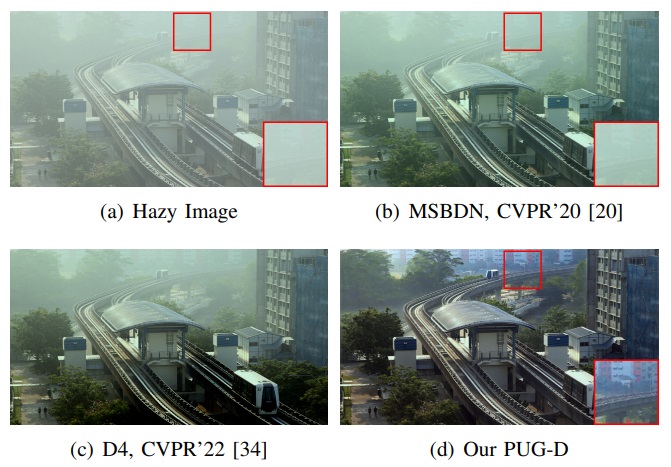}
     \caption{Subjective comparison of our result on a real hazy image \cite{peng2019image} with the state-of-the-art MSBDN~\cite{Dong_2020_CVPR} and D4~\cite{yang2022self}  methods for dehazing. Cropped regions in  boxes are for detailed inspection.}
  \label{fig: teaser} 
\end{figure}

Guidance by an appropriate transmission map can help in effective dehazing for a wide range of haze density because a transmission map essentially provides the amount of haze at image pixels as a function of the scene depth. Guidance by a suitable atmospheric light can also prove to be useful in avoiding color distortions as an atmospheric light represents the illumination associated with the haze.
So guidance from transmission map and atmospheric light in an end-to-end deep dehazing model could prove effective, which we investigate in our work. However, such an approach must not compute the haze parameters disjoint from the image dehazing process so that the end-to-end framework can avoid sub-optimal dehazing by adapting to the guidance appropriately. Therefore, after an initial estimation of the transmission map and atmospheric light based on losses specific to them, the haze parameters must be updated jointly with the dehazing operation. This will make the multiple deep networks that jointly perform the two updates and the dehazing interdependent among themselves. Further, the end-to-end framework that jointly updates the haze parameters and dehazes can be designed to do so progressively so that the updating and dehazing processes feed from each other obtaining refined estimates targeted toward a better dehazed output.

In this paper, we propose a multi-network dehazing framework that includes transmission map and atmospheric light updater networks for progressively guiding a dehazing network with the networks being interdependent among themselves. A schematic representation of our framework, Progressive Update Guided Dehazing (PUG-D), having five separate networks (Net 1-Net 5) is shown in Fig.~\ref{ga}. Initial estimates of transmission map and atmospheric light from dedicated estimator networks (Net 1 and Net 2) are used in the framework. The framework contains two separate novel updater networks (Net 3 and Net 4) that are used to progressively update the initial estimates of transmission map and atmospheric light. The updating happens jointly with the progressive refinement of the dehazed image estimated by the novel dehazing network (Net 5) making the networks interdependent. Our dehazing module comprising of the updater and dehazing networks (Net 3, 4, 5) is trained end-to-end while leveraging guidance by the progressively updated haze parameters. So, our approach neither uses any predefined model nor ignores transmission map and atmospheric light, and also it neither uses any image prior nor subjects itself to sub-optimal dehazing.

The proposed approach is extensively evaluated on standard and recent datasets, and compared qualitatively and quantitatively with the state-of-the-art considering both synthetic and natural hazy images in order to demonstrate its superior performance. With the help of ablation studies and auxiliary experiments, we empirically demonstrate the vital roles of the haze parameter guidance, the progressive haze parameter updating jointly with refined dehazing, and the end-to-end learning in our approach. Some of these investigations are reported in a \textit{supplementary} document. Our system is found to handle a wide variety of hazy conditions, ranging from low to high density, with or without color cast, producing detail-preserving, visibility enhanced and visually pleasing dehazed images. We also find that our approach possesses a unique image artifact-handling capability.

To summarize, the novel contributions of our paper are:
\begin{enumerate}[noitemsep]
\item We propose an end-to-end dehazing module that progressively dehazes hazy images using interdependent dehazing and updater networks.
\item We introduce novel haze parameter updater networks that update initial estimates of transmission map and atmospheric light to guide the dehazing process.
\item We propose a dehazing network that performs the refined dehazing jointly with the progressive updating by the updaters while invoking inter-step dependencies.
\item To handle color cast in hazy images, channel-wise atmospheric light is initially estimated using a novel deep network and then updated in the dehazing module.
\end{enumerate}
The rest of the paper is organized as follows. Section II discusses the related work. Section III elaborately describes our proposed image dehazing framework, PUG-D. Section IV presents the results of extensive experiments, and the qualitative and quantitative comparisons of our approach with the state-of-the-art. Section V discusses additional studies of our proposed approach. Section VI concludes the paper and discusses the future scope. Further, the supplementary of the paper presents  additional studies and experiments to show the efficacy of the proposed framework. Our paper's web-page is {\color{magenta}\href{https://aupendu.github.io/progressive-dehaze}{aupendu.github.io/progressive-dehaze}}.
\section{Related Work}
We categorize single image dehazing approaches into hand-crafted prior based and learning-based data-driven solutions. Most of the state-of-the-art approaches are deep learning based and belong to the second category.

\subsection{Handcrafted Prior based Image Dehazing}

Single image dehazing is an ill-posed problem in computer vision. Different astutely considered priors or assumptions have been used to solve this problem. Tan et al.~\cite{tan2008visibility} performed dehazing by maximizing contrast. Fattal et al.~\cite{fattal2008single} proposed a dehazing technique based on the relationship between surface shading and transmission map. He et al.~\cite{he2010single} proposed the dark channel prior (DCP), which is the most popular prior and has been employed in recent image dehazing techniques like in~\cite{7984895, 9134933}. Later, many priors have been proposed for dehazing~\cite{omer2004color, fattal2014dehazing, zhu2015fast, bui2017single, berman2018single, 9594664}.
Fattal et al. exploited the deviation of color-lines~\cite{omer2004color} from the origin due to the presence of haze to perform dehazing based on color-lines prior. 
Zhu et al. propose color attenuation prior~\cite{zhu2015fast} and use the same in a linear model  for image dehazing.  Bui et. al~\cite{bui2017single} fitted the hazy pixel clusters of RGB space in color ellipsoid and then calculate a prior vector to estimate the transmission map using color ellipsoid geometry for dehazing. Ju et al.~\cite{8931242} proposed a gamma correction prior based image dehazing model which generates a gamma corrected image from the input hazy image. Both the images are then used to compute the scene depth for performing dehazing. Later, Ju et al.~\cite{9502671} discussed on the limitations of single local/global prior and then proposed a model that blended non-local, local and global priors. 
Berman et. al proposed haze-lines~\cite{berman2018single} which are computed to estimate the transmission map for dehazing. Ju et. al~\cite{9594664} proposed region line prior for dehazing. In~\cite{9332266}, the authors introduced an enhanced atmospheric light scattering to handle the dim light and perform better dehazing. Ganguly et al.~\cite{9354831} introduced sparse haze model which was combined with atmospheric light scattering for image dehazing. Zhang et al.~\cite{9445080} proposed fish-retina inspired model to include the wavelength dependence of the scattering during haze formation. 

Effective atmospheric light estimation is an essential task for good quality dehazing~\cite{he2010single}. Tang et al.~\cite{tang2014investigating} proposed an optimization framework to estimate the atmospheric light while ensuring the quality of the dehazed image. Kim et al.~\cite{kim2011single} proposed a quadtree subdivision approach for the estimation. Sulami et al.~\cite{sulami2014automatic} exploited the distribution of pixels in small image patches to estimate the atmospheric light. However, the most popular way of estimating atmospheric light is by averaging the top 0.1\% of brightest pixels in the dark channel~\cite{he2010single}.   

Hazy images are prone to color cast issues in the presence of different atmospheric conditions like the sandstorm~\cite{huang2014visibility}.  Huang et al. proposed a dehazing technique which handles color cast due to sandstorms. Ancuti et al.~\cite{ancuti2013single} introduced white balancing in image dehazing for handling color cast. Choi et al.~\cite{choi2015referenceless} suggested a similar approach, where they introduced haze density weight along with white balancing for haze aware image dehazing. Peng et al.~\cite{peng2019image} proposed image dehazing through saturation correction that handles color cast. Recently, Kim et al.~\cite{kim2019fast} proposed saturation-based transmission map estimation for dehazing which performs color correction  using  white balancing  approach. 

\subsection{Learning based Image Dehazing}
The unprecedented success of deep convolution neural networks (CNN) in different computer vision tasks motivated the community to apply it for image dehazing. Cai et al.~\cite{cai2016dehazenet} introduced DehazeNet to estimate the transmission map for image dehazing. Li et al.~\cite{li2017aod} introduced AODNet, an end-to-end CNN model on a light scattering model to generate a dehazed image. Zhang et al.~\cite{zhang2018densely} proposed  DCPDN, a dense encoder-decoder model that follows a atmospheric light scattering model  to produce the haze-free image. Pang et al.~\cite{8529212} proposed HRGAN based on a generator network and a discriminator network to estimate atmospheric light and transmission map for dehazing. Qu et al.~\cite{qu2019enhanced} proposed a generative adversarial network (GAN) based Pix2pix  network that considers image-to-image translation over the light scattering model expression for dehazing. Zhao et. al.~\cite{9366772} proposed a weakly supervised framework, RefineDNet, for dehazing. Li et al.~\cite{8902220} exploited the idea of DCP along with a gradient prior to propose a semi-supervised deep model to dehaze images.   Golts et al.~\cite{8897130} used DCP loss for training an unsupervised deep network for image dehazing. Liu et al.~\cite{liu2019griddehazenet} introduced GridDNet, an end-to-end deep learning architecture with pre- and post-processing blocks for the same. Ren et al.~\cite{ren2018gated} proposed a multi-scale gated fusion network using an encoder-decoder architecture that produced haze-free images. Dudhane et al.~\cite{dudhane2019ryf} proposed RYF-Net for image dehazing. Dong et al.~\cite{Dong_2020_CVPR} proposed a multi-scale deep network which works on a strengthen-operate-subtract-boosting strategy for image dehazing.  Liu et al.~\cite{9125952} derived a trainable Hadamard-Product-Propagation model to recover the image details during dehazing.   Zhang et al.~\cite{9381290} estimated the haze density and image details by multi-scale hierarchical features and then, restore the hazy image. Zhao et al.~\cite{9252912} modeled  context dependencies using pyramid structure for dehazing. Zhang et al.~\cite{9302656} introduced an attention mechanism for emphasizing the important features during learning. Song et al.~\cite{9893197} proposed WSAMF-Net that utilizes both spatial and frequency domain information for image Reconstruction. Bai et. al~\cite{9677961} proposed a dehazing model based on a self-guided feature fusion framework. Recently, Yang et.al~\cite{yang2022self} propose a dehazing approach based on density and depth Decomposition.

\section{Progressive Update Guided Dehazing (PUG-D)}
We present our proposed framework PUG-D in two parts, with one of them comprising of the transmission map and atmospheric light estimators, and the other containing our novel dehazing module that includes the transmission map and atmospheric light updater networks and the dehazing network. We discuss the motivation of our work leading to the design of the dehazing module in Section~\ref{ModelMotiv}. We then describe our transmission map and atmospheric light estimator models in Section~\ref{sec: TA}. In Section~\ref{Sec: DehazingBlock}, we present our novel transmission map and atmospheric light updater networks and the proposed dehazing network, which together form our dehazing module.  In Section~\ref{sec: stage}, we discuss the training strategy for our framework.

\subsection{The Haze Formation Model and Our Motivation}
\label{ModelMotiv}
In the dehazing literature~\cite{li2018benchmarking}, it is well-accepted that haze in images can be explained through the following atmospheric light scattering model of haze formation~\cite{koschmieder1924theorie}:
\begin{equation}
    I(x)=J(x) t(x)+(1-t(x))A
    \label{eq_HazeForm}
\end{equation}
where $I(x)$ is the value at the pixel $x$ in the hazy image channel, $J(x)$ is the corresponding scene radiance, $A$ is the  atmospheric light and $t(x)$ is the transmission map. $t(x)$ is depth-dependent and it is defined as $t(x)=e^{-\beta d(x)}$, where $\beta$ is the attenuation coefficient, which is related to haze density, and $d(x)$ is the distance between the camera and the scene. Therefore, generating the dehazed image $J(x)$ from the hazy image $I(x)$ requires the estimation of depth-dependent parameter $t(x)$ along with the parameter $A$.

As mentioned in Section~\ref{Intro}, many contributions in the related literature estimate the transmission map $t(x)$ and the atmospheric light $A$ separately, and perform the dehazing operation as follows:
\begin{equation}
    J(x)=\frac{I(x)-A}{t(x)}+A
    \label{eq_dehaze}
\end{equation}
Estimation of $t(x)$ and $A$ using separate objective functions or using the reconstruction loss of the dehazed output through (\ref{eq_dehaze}) may restrict the image reconstruction that results in dehazing~\cite{li2017aod}. This is because the dehazing process using a predefined model does not possess the flexibility to adapt to the possible inaccuracies in the estimates of $t(x)$ and $A$. Therefore, in spite of the fact that haze in images is due to the model in (\ref{eq_HazeForm}), a dehazing model that is estimated jointly with $t(x)$ and $A$ may be better suited as it would be capable of adapting to the errors in $t(x)$ and $A$.

A few recently developed deep learning based methods consider end-to-end training for dehazing without estimating the transmission map and atmospheric light. Such techniques \cite{qu2019enhanced, ren2018gated} have been found to perform very well on images with lower amounts of haze, but they often do not work satisfactorily for a wide range of haze density. The absence of guidance by an appropriate transmission map may be a pivotal reason for this observation, as a transmission map essentially provides the amount of haze at image pixels and its use might make a model aware of the degree of dehazing required.

With the motivation to address the above issues, we use a separate end-to-end deep dehazing module along with deep estimators of haze parameters, that is, transmission map and atmospheric light (See Fig.~\ref{ga}). The advantage reaped is empirically demonstrated in Section~\ref{tab: formulavsIPUDN}. Our dehazing module takes a hazy image along with its estimated transmission map and atmospheric light as inputs, and updates the two estimates using a couple of updater networks. Then the dehazing network in the module progressively reconstructs the optimal dehazed image with guidance from the updated transmission map and atmospheric light. While the progressive updating of the haze parameters facilitates rectification of their estimates with the target to yield better dehazed outputs, the progressive dehazing jointly with the updating allows its adaptation to inaccuracies in the computed haze parameters. The advantages of the various components of our approach are shown in Section~\ref{sec: ablation}.

\subsection{Initial Estimates of the Haze Parameters}
\label{sec: TA}
\subsubsection{The Transmission Map Estimation Network} 
\label{sec: Tmap}
The transmission map provides useful information about haze density, which helps in the proper dehazing of an image. We use the densely connected encoder-decoder network of~\cite{zhang2018densely} to estimate the transmission map. We train the transmission map estimation model using structural similarity (SSIM)~\cite{wang2004image} as the loss function instead of the mean-squared error (MSE) loss function. \cite{zhang2018densely} shows that the use of SSIM loss gives sharper edges retaining structural information resulting in reduction of halo artifacts, one of the main issues associated with image dehazing. Once the model is trained, we use it as a transmission map estimator while training networks in the dehazing module, and later during testing. Note that the transmission map estimation does not employ any predefined image prior.

\subsubsection{Our Atmospheric Light Estimation Network}
\label{Atmospheric_Light_Estimation}
\setlength{\textfloatsep}{0.25cm}
\begin{figure}[ht]
    \centering
       \includegraphics[width=1\linewidth]{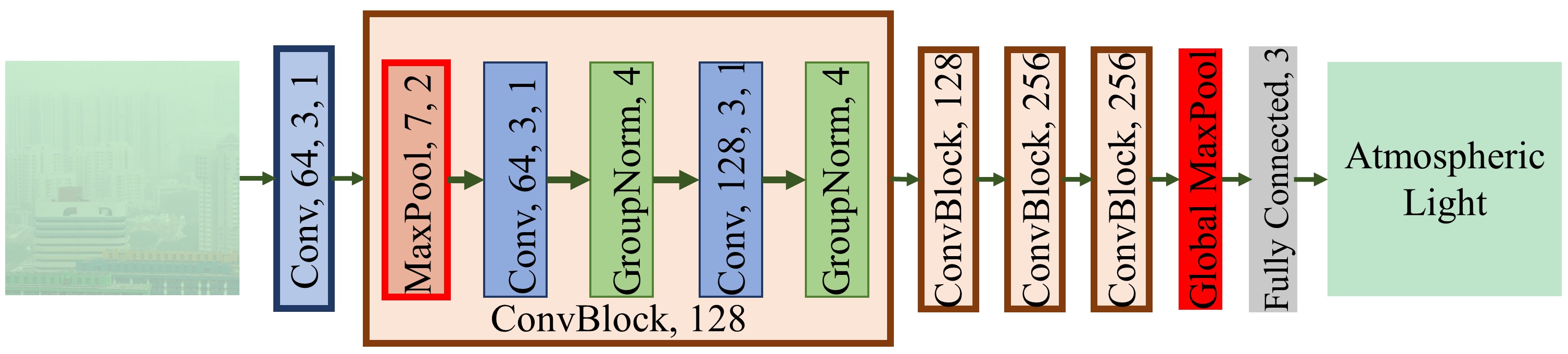}
  \caption{Atmospheric light estimation model to predict atmospheric light across $3$ color channels. The convolutional blocks hierarchically extract regional contributions to atmospheric light, which is pooled globally to get the maximum contribution as the estimate.}
  \label{fig: ambarchi} 
\end{figure}

\begin{figure}[ht]
    \centering
       \includegraphics[width=\linewidth]{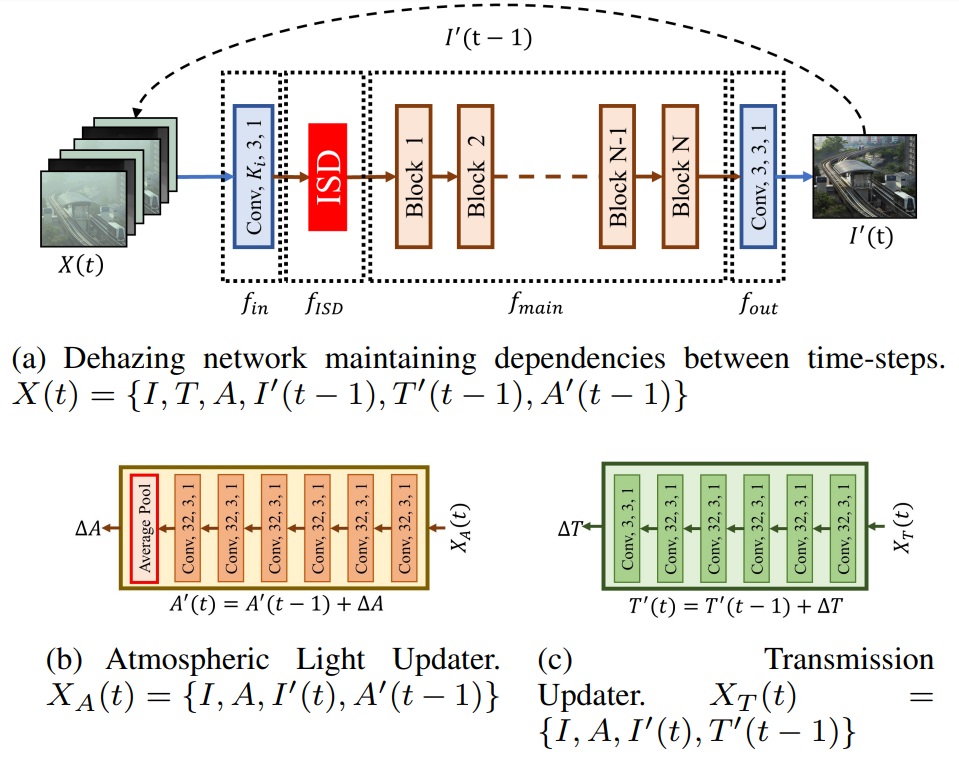}
    \caption{Our proposed  dehazing module with the interdependent dehazing and transmission map \& atmospheric light updater networks. 
    Here, $I'(t)$, $A'(t)$ and $T'(t)$ are the dehazed output, updated atmospheric light and updated transmission map, respectively, at time step $t$. Estimates from previous time-steps are fed back to the current time-step resulting in progressive updating and dehazing. The $f_{ISD}$ block maintains an inter-step dependency (ISD). The inputs to the dehazing model are $X(t)=\{I, T, A, I'(t-1), T'(t-1), A'(t-1)\}$, to the atmospheric light updater are $X_{A}(t)=\{I, A, I'(t), A'(t-1)\}$ and to the transmission map updater are $X_{T}(t)=\{I, A, I'(t), T'(t-1)\}$, where $I$, $A$ and $T$ are the hazy input image, and initial estimates of atmospheric light and transmission map, respectively. $f_{main}$ contains consecutive high-level feature learning blocks. The updater networks employ hierarchical feature extraction, with the atmospheric light updater using average pooling to aggregate all the pixel-level contributions.} 
  \label{fig: dehazing} 
\end{figure} 

Atmospheric light is a critical factor for generating dehazed outputs with proper lighting condition. Inaccurate estimation of atmospheric light may lead to under or overexposed images with color distortions. For atmospheric light estimation, we propose a convolutional neural network architecture as shown in Fig.~\ref{fig: ambarchi}. As evident, we use sequentially stacked convolutional layers, where each of them is followed by group normalization~\cite{wu2018group} and ReLU non-linearity. $7\times7$ max-pool layers with stride $2$ help in reducing spatial dimension in subsequent pairs of convolutional blocks. The feature extracted by the stacked layers is discussed in Section~IV-E of the \textit{supplementary}. Our model computes a three-element atmospheric light vector, an element corresponding to each color channel. As the presence of color cast in a hazy image affects one or more color channels, estimating channel-wise atmospheric light facilitates color cast reduction. The large max-pooling kernel size reduces the effect of local factors like object color while estimating the atmospheric light, which is just a single value for a color channel. We also use a global max-pooling at the end, which pools a single maximum intensity across the spatial dimension in a channel, as our target is to estimate the intensity of ambient light in the image channels. The intuition of the global max-pool is inspired from \cite{berman2018single}, where the atmospheric light is intuitively associated with the higher intensity pixels. To validate our usage, we performed experiments comparing global average-pooling to global max-pooling as shown in Section~III-D of the \textit{supplementary} and found that global max-pooling gives a far better estimate of the atmospheric light. We trained our atmospheric light estimator using the mean-squared error as the loss function. The proposed atmospheric light estimation does not employ any predefined image prior.

\subsection{The Dehazing Module Performing Joint Progressive Haze Parameter Updating and Dehazing} 
\label{Sec: DehazingBlock}
In this section, we describe our primary novel contribution, the dehazing module comprising of networks to progressively update the transmission map and atmospheric light estimates and a network to perform progressive dehazing using the updated haze parameters. All the three networks in the module are trained jointly making them interdependent allowing optimal dehazing. Our dehazing module comprising of the two updater networks and the dehazing network is shown in Fig.~\ref{fig: dehazing}. We first describe here the proposed processes of progressive updating and progressive dehazing, which are followed by the corresponding network architectures and loss functions.

\subsubsection{The Progressive Haze Parameter Updating Jointly with Dehazing}
The entire algorithm of the proposed dehazing system having five networks is given as Algorithm~\ref{algo}. We describe here the information flow during the training and the testing of our dehazing system. During forward propagation, the already trained transmission map estimation model $\Gamma$ (Fig~\ref{ga}: Net 1, Section~\ref{sec: Tmap}) and atmospheric light estimation model $\Lambda$ (Fig~\ref{ga}: Net 2, Section~\ref{Atmospheric_Light_Estimation}) take the hazy image $I$ as input and outputs initial estimates of transmission map $T$ and atmospheric light $A$, respectively.

The dehazing module takes in the estimated transmission map $T$ and atmospheric light $A$ along with the hazy image $I$ and progressively dehazes the image, where the transmission map and atmospheric light are also updated. The inputs involved in our dehazing module at time step $t$ of the progressive updating and dehazing are of two types: static and dynamic. Static inputs are the hazy image $I$, the initially estimated transmission map $T$ and the initially estimated atmospheric light $A$, which are time-independent. The time-dependent dynamic inputs contain the dehazed image $I'(t-1)$, the updated transmission map $T'(t-1)$ and the updated atmospheric light $A'(t-1)$, where $t-1$ represents the previous time step. At the first timestep $t=1$, we take $I'(t-1)=I$, $T'(t-1)=T$, and $A'(t-1)=A$. In each time step, the transmission map and atmospheric light are updated and a new dehazed output is computed. 

At a time step $t$, the three static inputs and the three dynamic inputs from the previous time step, which we represent together as $X(t)$, are fed into the dehazing network $H$. The dehazing network (Fig~\ref{ga}: Net 5) is shown elaborately in Fig.~\ref{fig: dehazing}(a) and its architecture is described in Section~\ref{sec: dehazingarchi}. $H$ provides a dehazed output $I'(t)$ at time step $t$. The input $X_{T}(t)$ at a time step $t$ to the transmission map updater $\cup_{\Gamma}$ contains the hazy image $I$, initially estimated transmission map $T$, the dehazed image $I'(t)$ from the network $H$ and the updated transmission map $T'(t-1)$ from previous time step $t-1$. In a similar manner, the hazy image $I$, initially estimated atmospheric light $A$, the dehazed image $I'(t)$ from the network $H$ and the updated atmospheric light $A'(t-1)$ from previous time step $t-1$ form the input $X_{A}(t)$ at a time step $t$ to the atmospheric light updater $\cup_{\Lambda}$. The transmission map and atmospheric light updater networks (Fig~\ref{ga}: Net 3 \& 4) are shown elaborately in Figs.~\ref{fig: dehazing} (c) and~\ref{fig: dehazing} (b), respectively, and their architectures are described in Section~\ref{sec: updater_network}. The transmission map and atmospheric light updater networks output the respective updates $\Delta T$ and $\Delta A$. Then, the updated transmission map $T'(t)$ at timestep $t$ is obtained as $T'(t-1) + \Delta T$ and the updated atmospheric light $A'(t)$ at timestep $t$ is obtained as $A'(t-1) + \Delta A$. 

\setlength{\textfloatsep}{0pt}
\begin{algorithm}[t]
\SetAlgoLined
\KwResult{$I'$=dehazed image}
\KwData{$I$=hazy image}
 $maxtimestep=t_{e}$ (Total number of steps)\;
 $T=\Gamma(I)$ (Estimate transmission map)\;
 $A=\Lambda(I)$ (Estimate atmospheric map)\;
 $I'(0)=I, T'(0)=T, A'(0)=A$\;
 \While{$t<t_{e}$}{
  $I'(t)=H(I, T, A, I'(t-1), T'(t-1), A'(t-1))$\;
  $\Delta T = \cup_{\Gamma}(I, T, I'(t), T'(t-1))$\;
  $\Delta A = \cup_{\Lambda}(I, A, I'(t), A'(t-1))$\;
  $T'(t)=T'(t-1)+\Delta T$\;
  $A'(t)=A'(t-1)+\Delta A$\;
 }
$I'(t_e)=H(I, T, A, I'(t_e-1), T'(t_e-1), A'(t_e-1))$\;
 \If {is train}{
 Calculate loss between $I_{gt}$ (ground truth) $\& \ I'(t_{e})$\;
 Calculate gradients\;
 Update parameters\;
 }
 \caption{The proposed dehazing algorithm}
 \label{algo}
\end{algorithm}

The dehazing network and the transmission map and atmospheric light updaters are trained jointly by back-propagating the loss between the dehazed output after finishing all the time steps and the ideal hazy-free image. The training process involving all the five networks of our approach is elaborated in Section~\ref{sec: implement}. In the case of testing, we obtain the required dehazed image from the network $H$ after finishing all the time steps, that is, $\hat{J}=I'(t_e)$, where $\hat{J}$ is the estimated haze-free image (see (\ref{eq_HazeForm})) and $t_e$ is the total number of step (a hyper-parameter discussed in Section~\ref{sec: dehazingarchi}).

\subsubsection{The Progressive Dehazing Approach}
\label{sec: dehaze_itr}

Our dehazing network in Fig.~\ref{fig: dehazing} (a) consists of four main parts: (a) Input feature extraction, $f_{in}$, (b) Inter-step dependency layer, $f_{ISD}$, (c) Consecutive trainable blocks for higher-level feature extraction, $f_{ftr}$, (d) Output layer for dehazed image reconstruction, $f_{out}$. The dehazing network can be mathematically described as:
\begin{equation}
\begin{aligned}
    y(t) = f_{in}(X(t)),\\
    z(t) = f_{ISD}(z(t-1), y(t)),\\
    I'(t) = f_{out}(f_{ftr}(z(t)))
    \label{RDnet_eq}
\end{aligned}
\end{equation}
The architectures of the said parts are described in Section~\ref{sec: dehazingarchi}. While $f_{in}$ extracts features $y(t)$ from the input $X(t)$, $f_{out}$ reconstructs the dehazed output $I'(t)$ from the features at its input. We use simple convolutional blocks as $f_{in}$ and $f_{out}$. $f_{ISD}$ takes the features $y(t)$ as inputs and outputs $z(t)$. The $z(t-1)$ output from $f_{ISD}$ in the previous step of the entire dehazing module is used at the input of $f_{ISD}$ in the current step along with $y(t)$. This allows us to effectively maintain dependencies between the steps enabling interaction among intermediate features from different time steps. We simply use a convolutional LSTM block as $f_{ISD}$ with $z(t)$ and $z(t-1)$ forming its current and previous states, respectively.
The high-level feature extractor $f_{ftr}$ contains consecutive trainable blocks as shown in Fig.~\ref{fig: dehazing} (a), which can be viewed to be performing feature-to-feature translation. A simple architecture of consecutive residual blocks can form $f_{ftr}$. The deep networks with sophisticated architectures used in several end-to-end dehazing methods such as MSBDN~\cite{Dong_2020_CVPR} and GridDNet~\cite{liu2019griddehazenet} can also act as  appropriate $f_{ftr}$ architectures. We use a couple of different $f_{ftr}$ architectures as mentioned next in Section~\ref{sec: dehazingarchi}.

\subsubsection{Architectures of the Networks in the Dehazing Module}

\paragraph{Dehazing Network} 
\label{sec: dehazingarchi}
In our dehazing network architecture shown in Fig.~\ref{fig: dehazing} (a), $f_{in}$ and $f_{out}$ are single layer convolutional networks. All the convolutional layer filters have $3\times3$ size and padding $1\times1$ with $ReLU$ non-linearity. There are $14$ input channels in $f_{in}$ due to the concatenation of the RGB hazy and intermediate dehazed images, the transmission map and the $3$-channel atmospheric light with both of them in static and dynamic forms, and there are $K_i$ output channels. $f_{out}$ takes the output of $f_{ftr}$ with $K_o$ channels as the input and outputs a $3$-channel RGB dehazed image. In $f_{ISD}$,  all the convolutional layer filters again have $3\times3$ size and padding $1\times1$ with $ReLU$ non-linearity and there are $K_i$ input channels and $K_i$ output channels. $f_{ftr}$ takes the $K_i$ channels at the output of $f_{ISD}$ and provides a $K_o$ channel output. As said earlier, $f_{ftr}$ can be any group of consecutive trainable blocks that performs feature-to-feature translation extracting high-level features. We consider three such different architectures for $f_{ftr}$. A simple group of 6 consecutive residual blocks is considered as the $f_{ftr}$ architecture with $K_i=K_o=32$, which is employed in ablation studies of our approach. For comparative experiments and performance analysis, along with the simple architecture, we also consider the sophisticated feature extraction architectures used in the dehazing approaches of MSBDN and GridDNet as $f_{ftr}$. While MSBDN's encoder-boosted decoder architecture with multi-scale blocks is used as $f_{ftr}$, GridDNet's grid network with attention-based multi-scale consecutive blocks is employed for the same. 

\paragraph{Updater Networks} 
\label{sec: updater_network}
We use six consecutive convolutional blocks in both the updater networks shown in Figs.~\ref{fig: dehazing} (c) and \ref{fig: dehazing} (b) to estimate the changes /updates required in the transmission map and atmospheric light at each time step. Both the updater networks have consecutive convolutional blocks with parametric $ReLU$ non-linearity, and the filters are of $3\times3$ size and $1\times1$ padding is used. We use the hyperbolic tangent non-linearity in the last convolutional layer so that the changes can be in both positive and negative directions. In the case of the atmospheric light updater, we use global average pooling after the last convolutional layer to get a single global update instead of pixel-wise updates, as we found the former to be empirically superior 

\subsubsection{Loss Functions}
\label{loss_function}
A combination of different loss functions like $MSE$, $L_{1}$, $SSIM$, adversarial~\cite{ledig2017photo} loss and perceptual~\cite{johnson2016perceptual} loss has usually been used in literature for training dehazing models. In a similar manner, we consider two different loss functions for training our dehazing module. We impose the supervision on the final dehazed output $I'(t_e)$ with the number of time steps as $t_e$. Hence, the losses are computed between $I'(t_e)$ and $I_{gt}$, which is the ground truth haze-free image. One of the losses used is the $L_{1}$ loss, which we empirically find to be superior to the $MSE/L_{2}$ loss for training our system.
The other loss used by us is the perceptual  loss~\cite{johnson2016perceptual}. The perceptual loss function used by us is as follows
\begin{equation}
    \mathcal{L}_{P}=\frac{1}{CHW}\sum_{c=1}^C\sum_{h=1}^H\sum_{w=1}^W\mid \phi_{c,h,w}(I'(t_e)) - \phi_{c,h,w}(I_{gt})\mid
    \label{vgg_eq}
\end{equation}
Unlike~\cite{johnson2016perceptual}, in our perceptual loss computation we use absolute error between the high-level features extracted by a VGG network from the ground truth haze-free and the dehazed image. In the above, we consider the high features at the \textit{relu2\_2} layer of the vgg19 architecture~\cite{simonyan2014very}. So, the total reconstruction loss $\mathcal{L}$ is defined as
\begin{equation}
    \mathcal{L}=\mathcal{L}_{L1}+\lambda\mathcal{L}_{P}
    \label{loss_eq}
\end{equation}
where $\mathcal{L}_{L1}$ is the mean absolute difference loss and $\mathcal{L}_{P}$ is the perceptual loss. $\lambda$ is a hyper-parameter and we use $\lambda=0.8$.

\subsection{Stage-wise Training and Fine Tuning of Our Entire Dehazing System}
\label{sec: stage}
Our proposed dehazing system is trained in three stages with the last stage being a fine-tuning stage. The transmission map and atmospheric light estimators are trained first and then the dehazing module with the updater and dehazing networks is trained end-to-end. Instead of training the whole framework in one stage, we divide the training procedure into three stages. This is done as we experimentally found that training the whole system as one from the beginning makes the convergence slow, and the training gets stuck in poor local minima. The different objective functions involved possibly push the training in different directions producing detrimentally small gradient magnitudes.  

So, in the first stage, the transmission map and atmospheric light estimators are trained separately with their respective loss functions. In the second stage, the updater networks and the dehazing network in the  dehazing module are trained jointly using the reconstruction loss $\mathcal{L}$ in an end-to-end manner. In the third stage, all five trained networks in the entire system are fine-tuned considering them together with the three different objective functions. This fine-tuning, which is carried out at a lower learning rate, is performed to introduce intricate dependencies between all the five networks performing the initial estimations of transmission map and atmospheric light, their updating and the dehazing. Fine-tuning them together allows us to achieve the best performance from our entire dehazing system, while remaining in the local vicinity of the solutions provided by the three individual trainings. The training details of our approach are provided in Section~\ref{sec: implement}.

\section{Experimental Results}

In this section, we evaluate the performance of the proposed approach (PUG-D) by comparing its dehazing results to that of   EPix2Pix~\cite{qu2019enhanced}, GridDNet~\cite{liu2019griddehazenet}, ALC~\cite{peng2019image}, Haze-Lines~\cite{berman2018single}, MSBDN~\cite{Dong_2020_CVPR},   FSID~\cite{kim2019fast}, IDE~\cite{9332266}, IDRLP~\cite{9594664}, RefineDNet~\cite{9366772} and D4~\cite{yang2022self}. The publicly available implementation codes provided by the respective authors of the above state-of-the-art approaches are used. The three variants of the proposed approach resulting from the different $f_{ftr}$ blocks employed (See Sections~\ref{sec: dehaze_itr},~\ref{sec: dehazingarchi}), which we refer to as PUG-D(R6), PUG-D(G) and PUG-D(M), are compared with the existing approaches on $5$ different datasets. PUG-D(R6) is the baseline where $f_{ftr}$ blocks consist of six consecutive residual blocks. PUG-D(G) and PUG-D(M) use the blocks of GridDNet~\cite{liu2019griddehazenet} and MSBDN~\cite{Dong_2020_CVPR}, respectively in the $f_{ftr}$ blocks. 

\begin{table*}[t]
\caption{Comparison of different state-of-the-art approaches with the proposed for dehazing on images from NR-haze dataset with varying degrees of synthetic haze and different color casts. All learned models are trained on NR-haze dataset. \newline (Best in bold red, second best in bold blue, and third best in bold black. $*$ signifies unpaired/ unsupervised techniques and $\S$ signifies supervised techniques).}
\resizebox{1\textwidth}{!}{%
\centering
\begin{tabular}{|l|c|c|c|c|c|c|c|}
\hline
\multicolumn{1}{|c|}{\multirow{3}{*}{ Techniques}} & 
\multicolumn{6}{c|}{Non-cast Hazy Images}  & Color Cast Hazy Image   \\ \cline{2-8} 

\multicolumn{1}{|c|}{}    
& \begin{tabular}[c]{@{}c@{}}Outdoor\\ Low Haze\\(LSOT)\end{tabular}                                 & \begin{tabular}[c]{@{}c@{}}Outdoor\\ Mid Haze\\(MSOT)\end{tabular} 
& \begin{tabular}[c]{@{}c@{}}Outdoor\\ High Haze\\(HSOT)\end{tabular}                                & \begin{tabular}[c]{@{}c@{}}Indoor\\ Low Haze\\(LSIT)\end{tabular}        
& \begin{tabular}[c]{@{}c@{}}Indoor\\ Mid Haze\\(MSIT)\end{tabular}                                  & \begin{tabular}[c]{@{}c@{}}Indoor\\ High Haze\\(HSIT)\end{tabular} 
& \begin{tabular}[c]{@{}c@{}}Color Cast\\ Random Haze\\(SCHT)\end{tabular} \\ \cline{2-8} 

\multicolumn{1}{|c|}{} 
& \begin{tabular}[c]{@{}c@{}}PSNR/SSIM/CIEDE\end{tabular} 
& \begin{tabular}[c]{@{}c@{}}PSNR/SSIM/CIEDE\end{tabular} 
& \begin{tabular}[c]{@{}c@{}}PSNR/SSIM/CIEDE\end{tabular} 
& \begin{tabular}[c]{@{}c@{}}PSNR/SSIM/CIEDE\end{tabular} 
& \begin{tabular}[c]{@{}c@{}}PSNR/SSIM/CIEDE\end{tabular} 
& \begin{tabular}[c]{@{}c@{}}PSNR/SSIM/CIEDE\end{tabular} 
& \begin{tabular}[c]{@{}c@{}}PSNR/SSIM/CIEDE\end{tabular} \\ \hline

\begin{tabular}[c]{@{}l@{}}EPix2Pix$^\S$~\cite{qu2019enhanced}\end{tabular} & 
\begin{tabular}[c]{@{}c@{}}20.40/0.8793/36.62\end{tabular}  & 
\begin{tabular}[c]{@{}c@{}}19.60/0.8508/39.57\end{tabular}  & 
\begin{tabular}[c]{@{}c@{}}18.20/0.7860/44.25\end{tabular}  & 
\begin{tabular}[c]{@{}c@{}}21.00/0.8441/33.02\end{tabular}  & 
\begin{tabular}[c]{@{}c@{}}18.30/0.7790/38.75\end{tabular}  & 
\begin{tabular}[c]{@{}c@{}}16.57/0.7171/43.46\end{tabular}  & 
\begin{tabular}[c]{@{}c@{}}20.97/0.8299/45.83\end{tabular}  \\ \hline

\begin{tabular}[c]{@{}l@{}}GridDNet$^\S$~\cite{liu2019griddehazenet}\end{tabular} & \begin{tabular}[c]{@{}c@{}}21.36/0.8820/34.73\end{tabular}  & 
\begin{tabular}[c]{@{}c@{}}20.67/0.8545/36.15\end{tabular}  & 
\begin{tabular}[c]{@{}c@{}}19.31/0.8013/39.77\end{tabular}   & 
\begin{tabular}[c]{@{}c@{}}23.11/0.8916/27.89\end{tabular}   & 
\begin{tabular}[c]{@{}c@{}}21.34/0.8552/31.28\end{tabular}  & 
\begin{tabular}[c]{@{}c@{}}17.91/0.7793/38.02\end{tabular}  & 
\begin{tabular}[c]{@{}c@{}}22.69/0.8642/36.65\end{tabular}   \\ \hline

\begin{tabular}[c]{@{}l@{}}ALC$^*$~\cite{peng2019image}\end{tabular} & 
\begin{tabular}[c]{@{}c@{}}19.45/0.8500/27.67\end{tabular}  & 
\begin{tabular}[c]{@{}c@{}}16.89/0.7438/36.33\end{tabular}  & 
\begin{tabular}[c]{@{}c@{}}14.56/0.6289/44.25\end{tabular}  & 
\begin{tabular}[c]{@{}c@{}}19.94/0.8725/22.59\end{tabular}  & 
\begin{tabular}[c]{@{}c@{}}16.27/0.7778/31.98\end{tabular}  & 
\begin{tabular}[c]{@{}c@{}}13.50/0.6763/42.38\end{tabular}  & 
\begin{tabular}[c]{@{}c@{}}12.26/0.3148/79.15\end{tabular}  \\ \hline

\begin{tabular}[c]{@{}l@{}}Haze-Lines$^*$~\cite{berman2018single}\end{tabular}  & \begin{tabular}[c]{@{}c@{}}16.59/0.7941/41.83\end{tabular}  & 
\begin{tabular}[c]{@{}c@{}}13.44/0.6350/58.06\end{tabular}  & 
\begin{tabular}[c]{@{}c@{}}11.16/0.4972/76.17\end{tabular}  & 
\begin{tabular}[c]{@{}c@{}}15.62/0.7303/45.88\end{tabular}  & 
\begin{tabular}[c]{@{}c@{}}11.75/0.6020/62.87\end{tabular}  & 
\begin{tabular}[c]{@{}c@{}}8.48/0.4493/79.11\end{tabular}  & 
\begin{tabular}[c]{@{}c@{}}11.05/0.2749/77.04\end{tabular}  \\ \hline

\begin{tabular}[c]{@{}l@{}}FSID$^*$~\cite{kim2019fast}\end{tabular} & 
\begin{tabular}[c]{@{}c@{}}17.83/0.8107/28.49\end{tabular}  & 
\begin{tabular}[c]{@{}c@{}}13.36/0.6720/39.12\end{tabular}  & 
\begin{tabular}[c]{@{}c@{}}10.69/0.5514/47.01\end{tabular}  & 
\begin{tabular}[c]{@{}c@{}}18.51/0.7900/29.63\end{tabular}  & 
\begin{tabular}[c]{@{}c@{}}13.90/0.7265/35.89\end{tabular}  & 
\begin{tabular}[c]{@{}c@{}}11.29/0.6317/43.20\end{tabular}  & 
\begin{tabular}[c]{@{}c@{}}10.80/0.3521/75.30\end{tabular}  \\ \hline

\begin{tabular}[c]{@{}l@{}}MSBDN$^\S$~\cite{Dong_2020_CVPR}\end{tabular}  & 
\begin{tabular}[c]{@{}c@{}}\textcolor{red}{\textbf{27.93}}/\textcolor{blue}{\textbf{0.9597}}/\textbf{21.96}\end{tabular}  & 
\begin{tabular}[c]{@{}c@{}}\textcolor{blue}{\textbf{26.55}}/\textcolor{blue}{\textbf{0.9480}}/23.94\end{tabular}  & 
\begin{tabular}[c]{@{}c@{}}\textcolor{blue}{\textbf{24.49}}/\textcolor{blue}{\textbf{0.9147}}/28.01\end{tabular}  & 
\begin{tabular}[c]{@{}c@{}}\textcolor{blue}{\textbf{30.67}}/\textcolor{blue}{\textbf{0.9721}}/13.89\end{tabular}  & 
\begin{tabular}[c]{@{}c@{}}\textcolor{blue}{\textbf{27.37}}/\textcolor{blue}{\textbf{0.9477}}/17.05\end{tabular}  & 
\begin{tabular}[c]{@{}c@{}}23.76/\textcolor{blue}{\textbf{0.9033}}/\textbf{22.39}\end{tabular}  & \begin{tabular}[c]{@{}c@{}}\textcolor{blue}{\textbf{27.65}}/\textcolor{blue}{\textbf{0.9422}}/\textcolor{blue}{\textbf{26.78}}\end{tabular}  \\ \hline

\begin{tabular}[c]{@{}l@{}}IDRLP$^*$~\cite{9594664}\end{tabular} & 
\begin{tabular}[c]{@{}c@{}}16.61/0.8143/36.17\end{tabular}  & 
\begin{tabular}[c]{@{}c@{}}13.28/0.6451/47.43\end{tabular}  & 
\begin{tabular}[c]{@{}c@{}}11.48/0.5258/52.94\end{tabular}  & 
\begin{tabular}[c]{@{}c@{}}17.30/0.7800/40.70\end{tabular}  & 
\begin{tabular}[c]{@{}c@{}}14.45 /0.6695/48.35\end{tabular}  & 
\begin{tabular}[c]{@{}c@{}} 10.09/0.4590/59.09\end{tabular}   & 
\begin{tabular}[c]{@{}c@{}} 8.96/0.1260/77.99\end{tabular}   \\ \hline

\begin{tabular}[c]{@{}l@{}}IDE$^*$~\cite{9332266}\end{tabular} & 
\begin{tabular}[c]{@{}c@{}}15.66/0.7601/39.03\end{tabular}  & 
\begin{tabular}[c]{@{}c@{}}12.91/0.6518/47.35\end{tabular}  & 
\begin{tabular}[c]{@{}c@{}}10.74/0.5548/51.63\end{tabular}  & 
\begin{tabular}[c]{@{}c@{}}14.14/0.7512/36.08\end{tabular}  & 
\begin{tabular}[c]{@{}c@{}}11.52/0.6541/44.80\end{tabular}  & 
\begin{tabular}[c]{@{}c@{}} 9.65/0.5569/54.82\end{tabular}   & 
\begin{tabular}[c]{@{}c@{}} 12.20/0.4273/70.47\end{tabular}   \\ \hline

\begin{tabular}[c]{@{}l@{}}Ours PUG-D(R6)\end{tabular}  & 
\begin{tabular}[c]{@{}c@{}}\textbf{25.83}/\textbf{0.9430}/22.44\end{tabular}  & 
\begin{tabular}[c]{@{}c@{}}\textbf{24.43}/0.9220/\textbf{22.77}\end{tabular}  & 
\begin{tabular}[c]{@{}c@{}}22.81/0.8822/\textbf{26.76}\end{tabular}  & 
\begin{tabular}[c]{@{}c@{}}\textbf{30.02}/\textbf{0.9645}/\textcolor{blue}{\textbf{12.44}}\end{tabular}  & 
\begin{tabular}[c]{@{}c@{}}\textbf{27.17}/\textbf{0.9401}/\textcolor{blue}{\textbf{16.04}}\end{tabular}  & 
\begin{tabular}[c]{@{}c@{}}\textcolor{blue}{\textbf{23.94}}/\textbf{0.8902}/\textcolor{blue}{\textbf{21.95}}\end{tabular}  & 
\begin{tabular}[c]{@{}c@{}}\textbf{26.74}/\textbf{0.9157}/\textbf{28.92}\end{tabular}  \\ \hline 

\begin{tabular}[c]{@{}l@{}}Ours PUG-D(G)\end{tabular}  & 
\begin{tabular}[c]{@{}c@{}}24.70/0.9401/\textcolor{blue}{\textbf{21.19}}\end{tabular}  & \begin{tabular}[c]{@{}c@{}}24.30/0.9294/\textcolor{blue}{\textbf{22.24}}\end{tabular}  & \begin{tabular}[c]{@{}c@{}}\textbf{23.23}/\textbf{0.8934}/\textcolor{blue}{\textbf{25.66}}\end{tabular}  & 
\begin{tabular}[c]{@{}c@{}}29.35/0.9630/13.15\end{tabular}  & \begin{tabular}[c]{@{}c@{}}26.67/0.9347/\textbf{17.00}\end{tabular}  & 
\begin{tabular}[c]{@{}c@{}}\textbf{23.94}/0.8850/22.42\end{tabular}  & 
\begin{tabular}[c]{@{}c@{}}26.23/0.9129/30.48\end{tabular}  \\ \hline 

\begin{tabular}[c]{@{}l@{}}Ours PUG-D(M)\end{tabular}  & \begin{tabular}[c]{@{}c@{}}\textcolor{blue}{\textbf{27.40}}/\textcolor{red}{\textbf{0.9607}}/\textcolor{red}{\textbf{17.36}}\end{tabular}  & \begin{tabular}[c]{@{}c@{}}\textcolor{red}{\textbf{26.79}}/\textcolor{red}{\textbf{0.9541}}/\textcolor{red}{\textbf{18.41}}\end{tabular}  & \begin{tabular}[c]{@{}c@{}}\textcolor{red}{\textbf{25.38}}/\textcolor{red}{\textbf{0.9289}}/\textcolor{red}{\textbf{22.17}}\end{tabular}  & \begin{tabular}[c]{@{}c@{}}\textcolor{red}{\textbf{31.66}}/\textcolor{red}{\textbf{0.9793}}/\textcolor{red}{\textbf{9.70}} \end{tabular}  & \begin{tabular}[c]{@{}c@{}}\textcolor{red}{\textbf{29.01}}/\textcolor{red}{\textbf{0.9593}}/\textcolor{red}{\textbf{13.12}}\end{tabular}  & \begin{tabular}[c]{@{}c@{}}\textcolor{red}{\textbf{25.56}}/\textcolor{red}{\textbf{0.9192}}/\textcolor{red}{\textbf{18.53}}\end{tabular}  & \begin{tabular}[c]{@{}c@{}}\textcolor{red}{\textbf{28.99}}/\textcolor{red}{\textbf{0.9460}}/\textcolor{red}{\textbf{23.51}}\end{tabular}  \\ \hline   
\end{tabular}
}
\label{tab:table1}
\end{table*} 

\subsection{Datasets} 
\label{sec: dataset}
The datasets used for comparing our dehazing approach to the state-of-the-art and for a detailed study of our approach are discussed here. 
This includes the well-known RESIDE dataset~\cite{li2018benchmarking} along with the BeDDe~\cite{9099036}, O-Haze~\cite{ancuti2018Ohaze} and I-Haze~\cite{ancuti2018Ihaze} datasets.
We also mention here the image sets used for training while comparing on each of these datasets. 

\subsubsection{Outdoor and Indoor Images with Synthetic Haze (for both Training and Testing) - RESIDE dataset}$\ $\\
 \label{RSIDE}
The popular RESIDE dataset~\cite{li2018benchmarking} containing both indoor and outdoor images is one of the two synthetic hazy image datasets that we consider to perform quantitative evaluation. In this quantitative comparison of our approach with the state-of-the-art, all the learning-based approaches including ours are trained and tested considering the corresponding image sets (as mentioned in Section~\ref{sec: implement}2 \& Table~\ref{tab:sotssyn}), namely, Indoor Training Set (ITS), Outdoor Training Set (OTS) and Synthetic Objective Testing Set (SOTS) provided in the RESIDE dataset.

\subsubsection{Outdoor and Indoor Images with Low, Medium and High Levels of Synthetic Haze and Different Color Casts (for both Training and Testing) - NR-haze dataset}$\ $\\
\label{OurDataset}
 The second synthetic hazy image dataset that we consider is one that we created such that not only both outdoor and indoor images are present but also the haze in them are in different categorized amounts and some of the outdoor images contain different types of color cast. Our procedure of forming the new NR-haze dataset from the NYU depth dataset~\cite{silberman2012indoor} and the RESIDE dataset is explained elaborately in Section~I of the \textit{supplementary}. Its training set contains $2378$ outdoor and $1349$ indoor images with random amounts of haze and color cast. It's testing set contains $350$ images categorized into $7$ sets, namely, Low-haze Synthetic Indoor Test (LSIT) and Outdoor Test (LSOT) sets, Mid-haze Synthetic Indoor Test (MSIT) and Outdoor Test (MSOT) sets, High-haze Synthetic Indoor Test (HSIT) and Outdoor Test (HSOT) sets, and Synthetic Color-cast Haze Test-set (SCHT), with $50$ hazy images in each of them. This dataset is employed for both qualitative and subjective performance evaluation. 
 All the learning-based approaches compared on the NR-haze dataset are trained and tested on the corresponding image sets provided in the dataset (mentioned in Section~\ref{sec: implement}2 \& Table~\ref{tab:table1}). This dataset is also employed to perform the ablation studies and additional experiments on the proposed approach, which are discussed in Section~\ref{sec: ablation}, and in Sections III and IV of the \textit{supplementary}. The models obtained by training the three variants of our approach on this dataset are considered as our pre-trained models for evaluation wherever applicable.  

\subsubsection{Outdoor and Indoor Images with Real Haze Generated by Haze Machines (for only Testing) - O-Haze and I-Haze datasets}$\ $\\
\label{RealDataset1}
The popular O-Haze~\cite{ancuti2018Ohaze} and I-Haze~\cite{ancuti2018Ihaze} datasets, which respectively contain $45$ outdoor and $35$ indoor images with real haze generated using haze machines along with the corresponding haze-free images, are also considered to perform quantitative evaluation. The two entire sets of images provided in the two datasets are considered for the evaluation and the pre-trained models of all the learning-based approaches provided by the corresponding authors are used for performance comparison with the pre-trained models of our approach.

\subsubsection{Outdoor Real Hazy and non-Hazy Image Pairs (for only Testing) - BeDDE dataset}$\ $\\
\label{RealDataset2}
BeDDE~\cite{9099036} dataset containing $208$ outdoor images with real haze and corresponding haze-free images is also considered for comparing dehazing results quantitatively. All the images in the dataset are considered for the quantitative performance comparison, for which again the pre-trained models of all the learning-based approaches provided by the corresponding authors are used along with the pre-trained models of our approach.

\subsubsection{Real-world Outdoor Hazy Images}
\label{RealDataset3}
Real-world outdoor hazy images from \cite{choi2015referenceless, peng2019image, fattal2014dehazing, li2018benchmarking} are considered for the subjective performance comparison of the proposed approach with the state-of-the-art. The dehazed results obtained from all the learning-based approaches including ours are again generated considering the corresponding pre-trained models.

\subsection{Model Training Details}
\label{sec: implement}
The training details of the variants of the proposed approach and the state-of-the-art approaches requiring paired supervision are given here. As mentioned earlier, for the comparative performance evaluation in this section, and additional studies of our approach in the next section and in the \textit{supplementary}, we consider the dehazing models trained on the relevant image sets in the RESIDE dataset~\cite{li2018benchmarking} and the NR-haze dataset of Section~\ref{OurDataset}. 

\subsubsection{Proposed Approach}
\label{Proposed Approach}

\paragraph{Training Setup}
\label{TrainSetup}
We follow the standard practice~\cite{li2018benchmarking} of synthetically adding random amounts of haze to the haze-free images from the training set using (\ref{eq_HazeForm}). A random amount of haze is generated through a random selection of the attenuation coefficient $\beta$ and the atmospheric light $A$. The values of depth $d(x)$ at image pixels available from depth maps are used for the same. We extract $16$ patches of size $224\times224$ for each batch update, and we augment those images using random horizontal and vertical flipping along with $90^{\circ}$ rotation. The Adam optimizer~\cite{kingma2014adam} with the default settings in the PyTorch environment is employed for the learning with the initial learning rate as  $10^{-4}$. If a validation set of images is available in the dataset, we validate the model on it after each iteration comprising of $2000$ batch updates, to obtain the best model for testing. Although we extract patches during training, we feed the hazy images as a whole into the network while performing dehazing on the testing set of hazy images in the dataset.

\paragraph{The 3-stage Training}
As discussed in Section~\ref{sec: stage}, we train our dehazing framework in three stages, for which we use the above training setup in each stage. In the first stage, we train the transmission map and atmospheric light estimators separately. We train both the transmission and atmospheric light estimation models for $250$ iterations ($2000$ batch updates each) using the Adam optimizer with the initial learning rate and a batch size of $16$, where we decay the learning rate by half after every $50$ iterations.

In the second stage, we use the already trained transmission and atmospheric light estimators to compute the initial transmission map and atmospheric light for each hazy image. We do not train the transmission and atmospheric light estimators during this stage. We train the dehazing module comprising of the transmission map and atmospheric light updaters and the dehazing network jointly using the hazy images of the training set, with the corresponding initial transmission map and atmospheric light as inputs. We train for $500$ iterations ($2000$ batch updates each) using the Adam optimizer with the initial learning rate, where a batch size of $2$ is used and the learning rate is decayed by half after every $100$ iterations.

At the final stage, we fine-tune the whole dehazing system together. We find that the fine-tuning results in performance improvement as some dependency between the initial prior estimators and the dehazing module is invoked. During fine-tuning, we keep the learning rate of transmission map and atmospheric light estimation models $10$ times lesser than the dehazing module and train the whole network for another $500$ iterations with the same training setup.
\begin{table*}[t]
\centering
\caption{Comparison of different state-of-the-art approaches with the proposed for single image dehazing on the hazy image from SOTS of the RESIDE dataset. All learned models are trained on RESIDE dataset. (Best in bold black)}
\label{tab:sotssyn}
\resizebox{\textwidth}{!}{%
\begin{tabular}{|c|c|c|c|c|c|c|c|c|c|c|c|c|c|}
\hline
\multirow{2}{*}{Measures} &
\multicolumn{13}{c|}{Techniques} \\ \cline{2-14} &
\multicolumn{7}{c|}{Unpaired/ Unsupervised} & \multicolumn{6}{c|}{Supervised} \\ \cline{2-14} &
  
  \begin{tabular}[c]{@{}c@{}}ALC\\ ~\cite{peng2019image}\end{tabular} & 
  \begin{tabular}[c]{@{}c@{}}Haze-Lines\\ ~\cite{berman2018single}\end{tabular} & 
  \begin{tabular}[c]{@{}c@{}}FSID\\ ~\cite{kim2019fast}\end{tabular} & 
  \begin{tabular}[c]{@{}c@{}}IDRLP\\ ~\cite{9594664}\end{tabular} & 
  \begin{tabular}[c]{@{}c@{}}IDE\\ ~\cite{9332266}\end{tabular} & 
  \begin{tabular}[c]{@{}c@{}}RefineDNet\\ ~\cite{9366772}\end{tabular} & 
  \begin{tabular}[c]{@{}c@{}}D4\\ ~\cite{yang2022self}\end{tabular} & 
  
  \begin{tabular}[c]{@{}c@{}}EPix2Pix\\ ~\cite{qu2019enhanced}\end{tabular} & 
  \begin{tabular}[c]{@{}c@{}}GridDNet\\ ~\cite{liu2019griddehazenet}\end{tabular} & 
  \begin{tabular}[c]{@{}c@{}}MSBDN\\ ~\cite{Dong_2020_CVPR}\end{tabular} & 
  \begin{tabular}[c]{@{}c@{}}Ours\\ PUG-D(R6)\end{tabular} & 
  \begin{tabular}[c]{@{}c@{}}Ours\\ PUG-D(G)\end{tabular} & 
  \begin{tabular}[c]{@{}c@{}}Ours\\ PUG-D(M)\end{tabular}  \\ \hline 
  
  \begin{tabular}[c]{@{}c@{}}PSNR\\ SSIM\end{tabular} &
  \begin{tabular}[c]{@{}c@{}}20.59\\ 0.87\end{tabular} & 
  \begin{tabular}[c]{@{}c@{}}16.86\\ 0.77\end{tabular} & 
  \begin{tabular}[c]{@{}c@{}}20.38\\ 0.82\end{tabular} & 
  \begin{tabular}[c]{@{}c@{}}18.46\\ 0.82\end{tabular} & 
  \begin{tabular}[c]{@{}c@{}}15.08\\ 0.76\end{tabular} & 
  \begin{tabular}[c]{@{}c@{}}20.52\\ 0.88\end{tabular} & 
  \begin{tabular}[c]{@{}c@{}}17.39\\ 0.69\end{tabular} & 
  
  \begin{tabular}[c]{@{}c@{}}22.72\\ 0.88\end{tabular} & 
  \begin{tabular}[c]{@{}c@{}}31.47\\ \textbf{0.98}\end{tabular} & 
  \begin{tabular}[c]{@{}c@{}}33.69\\ \textbf{0.98}\end{tabular} & 
  \begin{tabular}[c]{@{}c@{}}32.95\\ \textbf{0.98}\end{tabular} & 
  \begin{tabular}[c]{@{}c@{}}\textbf{33.92}\\ \textbf{0.98}\end{tabular} & 
  \begin{tabular}[c]{@{}c@{}}33.81\\ \textbf{0.98}\end{tabular}  \\ \hline 
\end{tabular}
}
\end{table*}

\begin{table*}[t]
\centering
\caption{Comparison of different state-of-the-art approaches with the proposed for single image dehazing on real-world hazy images. All learned models are the pre-trained ones (in datasets different from the testing) provided by the authors. \newline (Best in bold red, second best in bold blue, and third best in bold black)}
\label{tab:real_quanti}
\resizebox{\textwidth}{!}{%
\begin{tabular}{|c|c|c|c|c|c|c|c|c|c|c|c|c|c|c|}
\hline
\multirow{4}{*}{Dataset} &
  \multirow{4}{*}{Measures} &
  \multicolumn{13}{c|}{Techniques} \\ \cline{3-15} 
 & &
  \multicolumn{7}{c|}{Unpaired/ Unsupervised} & \multicolumn{6}{c|}{Supervised} \\ \cline{3-15} 
 &
   &
  \begin{tabular}[c]{@{}c@{}}ALC\\ ~\cite{peng2019image}\end{tabular} & 
  \begin{tabular}[c]{@{}c@{}}Haze-Lines\\ ~\cite{berman2018single}\end{tabular} & 
  \begin{tabular}[c]{@{}c@{}}FSID\\ ~\cite{kim2019fast}\end{tabular} & 
  \begin{tabular}[c]{@{}c@{}}IDRLP\\ ~\cite{9594664}\end{tabular} & 
  \begin{tabular}[c]{@{}c@{}}IDE\\ ~\cite{9332266}\end{tabular} & 
  \begin{tabular}[c]{@{}c@{}}RefineDNet\\ ~\cite{9366772}\end{tabular} & 
  \begin{tabular}[c]{@{}c@{}}D4\\ ~\cite{yang2022self}\end{tabular} & 
  
  \begin{tabular}[c]{@{}c@{}}EPix2Pix\\ ~\cite{qu2019enhanced}\end{tabular} & 
  \begin{tabular}[c]{@{}c@{}}GridDNet\\ ~\cite{liu2019griddehazenet}\end{tabular}  & 
  \begin{tabular}[c]{@{}c@{}}MSBDN\\ ~\cite{Dong_2020_CVPR}\end{tabular} & 
  \begin{tabular}[c]{@{}c@{}}Ours\\ PUG-D(R6)\end{tabular} & 
  \begin{tabular}[c]{@{}c@{}}Ours\\ PUG-D(G)\end{tabular} & 
  \begin{tabular}[c]{@{}c@{}}Ours\\ PUG-D(M)\end{tabular} \\ \hline  
BeDDE &
  \begin{tabular}[c]{@{}c@{}}VI\\ RI\end{tabular} &
  
  \begin{tabular}[c]{@{}c@{}}0.8620\\ \textbf{0.9696}\end{tabular} & 
  \begin{tabular}[c]{@{}c@{}}0.8715\\ 0.9589\end{tabular} & 
  \begin{tabular}[c]{@{}c@{}}0.8991\\ 0.9683\end{tabular} & 
  \begin{tabular}[c]{@{}c@{}}0.8914\\ 0.9683\end{tabular} & 
  \begin{tabular}[c]{@{}c@{}}0.8461\\ 0.9654\end{tabular} & 
  \begin{tabular}[c]{@{}c@{}}\textbf{0.9073}\\ 0.9707\end{tabular} & 
  \begin{tabular}[c]{@{}c@{}}0.8128\\ 0.9707\end{tabular} & 
  
  \begin{tabular}[c]{@{}c@{}}0.8956\\ 0.9640\end{tabular} & 
  \begin{tabular}[c]{@{}c@{}}0.8909\\ 0.9682\end{tabular} & 
  \begin{tabular}[c]{@{}c@{}}0.7688\\ 0.9039\end{tabular} & 
  \begin{tabular}[c]{@{}c@{}}0.9065\\ {\textbf{0.9711}}\end{tabular} & 
  \begin{tabular}[c]{@{}c@{}}\textcolor{blue}{\textbf{0.9074}}\\ \textcolor{red}{\textbf{0.9717}}\end{tabular} & 
  \begin{tabular}[c]{@{}c@{}}\textcolor{red}{\textbf{0.9075}}\\ 0.9687\end{tabular} \\ \hline  
O-Haze &
  \begin{tabular}[c]{@{}c@{}}VI\\ RI\\ PSNR\\ SSIM\end{tabular} &
  
  \begin{tabular}[c]{@{}c@{}}0.8666\\ 0.9623\\ 16.06\\ 0.45\end{tabular} & 
  \begin{tabular}[c]{@{}c@{}}0.8797\\ 0.9623\\ 15.81\\ 0.52\end{tabular} & 
  \begin{tabular}[c]{@{}c@{}}0.8946\\ 0.9582\\ 16.81\\ 0.52\end{tabular} & 
  \begin{tabular}[c]{@{}c@{}}0.8988\\ 0.9605\\ 14.77\\0.48\end{tabular} & 
  \begin{tabular}[c]{@{}c@{}}0.8862\\ 0.9570\\ 13.42\\ 0.44\end{tabular} & 
  \begin{tabular}[c]{@{}c@{}}0.8800\\ 0.9645\\ 17.12\\ 0.54\end{tabular} &  
  \begin{tabular}[c]{@{}c@{}}0.8485\\ 0.9540\\ 14.45\\ 0.35\end{tabular} & 
  
  \begin{tabular}[c]{@{}c@{}}0.9078\\ 0.9715\\ 17.38\\ 0.61\end{tabular} & 
  \begin{tabular}[c]{@{}c@{}}0.7653\\ 0.9150\\ 13.54\\ 0.37\end{tabular} & 
  \begin{tabular}[c]{@{}c@{}}0.8151\\ 0.9627\\ 16.83\\ 0.45\end{tabular} & 
  \begin{tabular}[c]{@{}c@{}}\textcolor{red}{\textbf{0.9164}}\\ \textcolor{blue}{\textbf{0.9737}}\\ \textbf{19.39}\\ \textcolor{blue}{\textbf{0.64}}\end{tabular} & 
  \begin{tabular}[c]{@{}c@{}}\textcolor{blue}{\textbf{0.9122}}\\ \textcolor{red}{\textbf{0.9749}}\\ \textcolor{red}{\textbf{20.00}}\\ \textcolor{red}{\textbf{0.64}}\end{tabular} & 
  \begin{tabular}[c]{@{}c@{}}\textbf{0.9113}\\ \textbf{0.9736}\\ \textcolor{blue}{\textbf{19.50}}\\ \textbf{0.63}\end{tabular} \\ \hline 
I-Haze &
  \begin{tabular}[c]{@{}c@{}}VI\\ RI\\ PSNR\\ SSIM\end{tabular} &
  
  \begin{tabular}[c]{@{}c@{}}0.9064\\ 0.9724\\ 14.34\\ 0.55\end{tabular} & 
  \begin{tabular}[c]{@{}c@{}}0.9083\\ 0.9685\\ 15.48\\ 0.60\end{tabular} & 
  \begin{tabular}[c]{@{}c@{}}0.9309\\ 0.9711\\ \textcolor{red}{\textbf{17.21}}\\ \textbf{0.61}\end{tabular} & 
  \begin{tabular}[c]{@{}c@{}}0.9273\\ 0.9752\\ 17.20\\ 0.62\end{tabular} & 
  \begin{tabular}[c]{@{}c@{}}0.9329\\ 0.9693\\ 15.37\\ 0.52\end{tabular} & 
  \begin{tabular}[c]{@{}c@{}}0.9283\\ 0.9702\\ 15.79\\ 0.64\end{tabular} &  
  \begin{tabular}[c]{@{}c@{}}0.9016\\ 0.9590\\ 13.60\\ 0.43\end{tabular} & 
  
  \begin{tabular}[c]{@{}c@{}}0.9336\\ 0.9727\\ 15.80\\ \textbf{0.61}\end{tabular} & 
  \begin{tabular}[c]{@{}c@{}}0.8693\\ 0.9197\\ 12.24\\ 0.47\end{tabular} & 
  \begin{tabular}[c]{@{}c@{}}0.9125\\ 0.9744\\ \textcolor{blue}{\textbf{16.57}}\\ \textcolor{red}{\textbf{0.64}}\end{tabular} & 
  \begin{tabular}[c]{@{}c@{}}\textcolor{red}{\textbf{0.9418}}\\ \textcolor{blue}{\textbf{0.9782}}\\ 16.21\\ \textcolor{blue}{\textbf{0.62}}\end{tabular} & 
  \begin{tabular}[c]{@{}c@{}}\textcolor{blue}{\textbf{0.9402}}\\ \textcolor{red}{\textbf{0.9785}}\\ 16.02\\ \textbf{0.61}\end{tabular} & 
  \begin{tabular}[c]{@{}c@{}}\textbf{0.9394}\\ \textbf{0.9776}\\ \textbf{16.26}\\ 0.60\end{tabular} \\ \hline 
\end{tabular}%
}
\end{table*}

\begin{figure*}
    \centering
       \includegraphics[width=\linewidth]{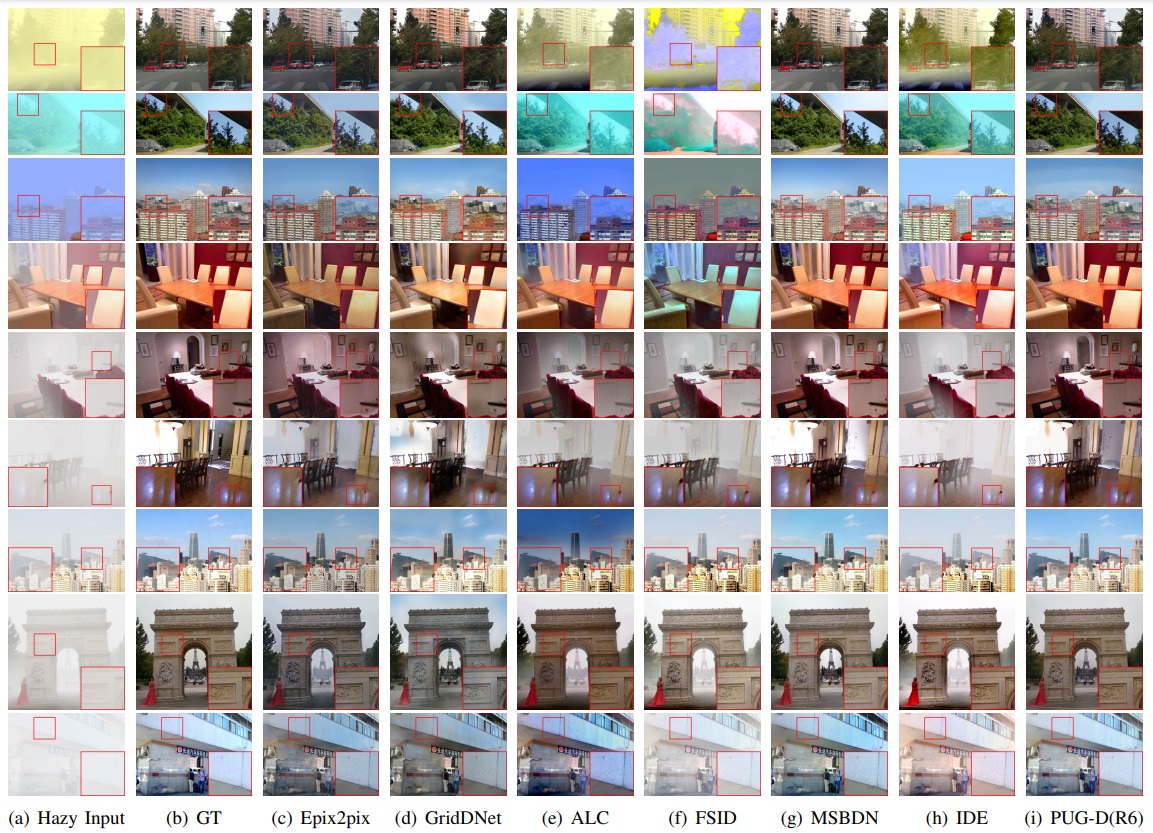}
     \caption{Subjective evaluation of the different methods on hazy images with synthetically generated haze. Zooming into  image regions like the cropped ones in boxes will show the effectiveness of our method. The 1st-3rd rows: Results on hazy images with color cast. The 4th-6th rows: Results on indoor images with low, medium, and high haze without color cast, respectively. The 7th-9th rows: Same as 4th-6th rows except the use of outdoor images. (See Fig. 3 in supplementary for results of PUG-D variants)}
\label{fig: sythetic}
\end{figure*}

\begin{figure*}
    \centering
       \includegraphics[width=\linewidth]{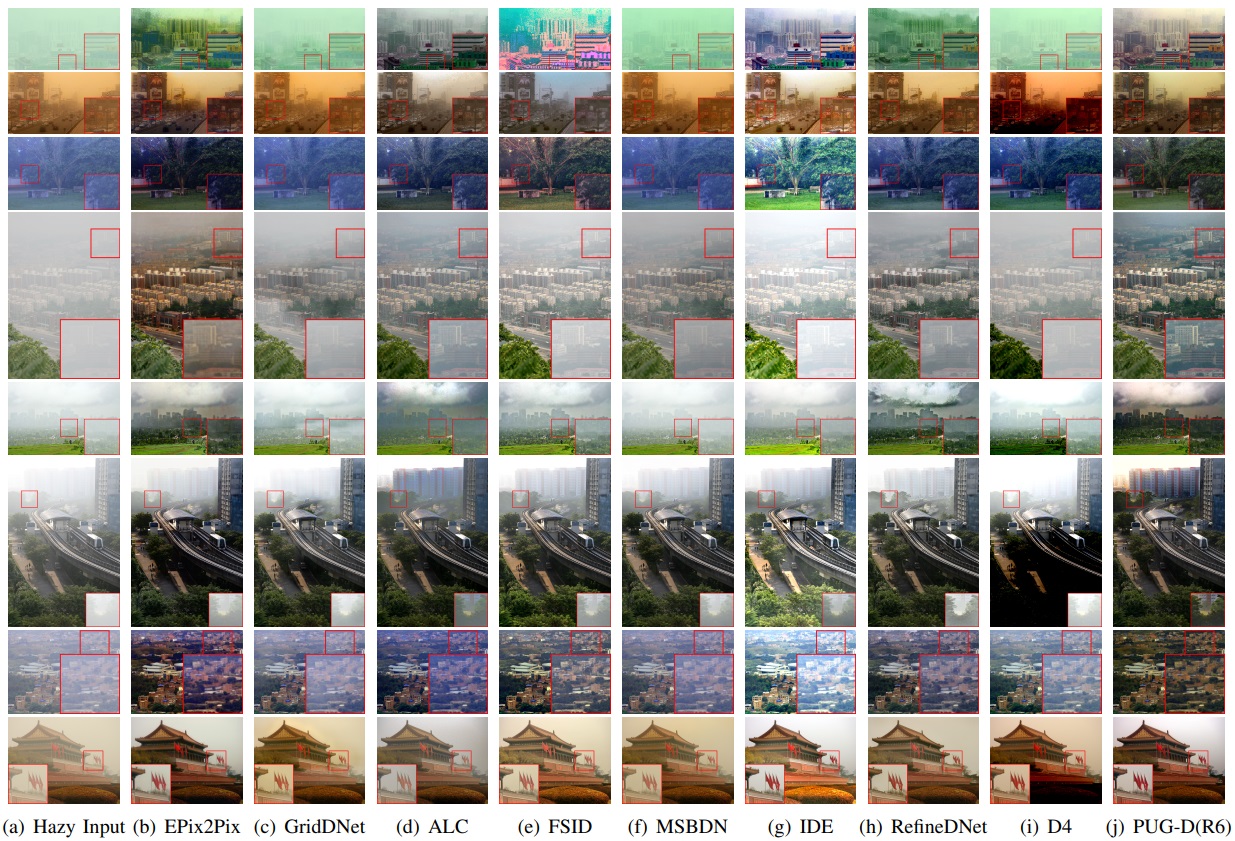}
    \caption{Subjective evaluation of the different methods on real-world hazy images. Zooming into image regions like the cropped ones in boxes will show the effectiveness of our method. The 1st-3rd rows: Results on hazy images with noticeable color cast. The 4th-8th rows: Results on hazy images with varying amounts of haze. (See Fig. 4 in supplementary for results of PUG-D variants)}
    \label{fig: real}
\end{figure*} 

\subsubsection{Existing Approaches}
\label{Existing}
The learning based models trained on the RESIDE dataset have already been provided by the corresponding authors. Therefore, to compare the performance of these approaches with the variants of our approach on the testing image set of the RESIDE dataset, we employ the relevant pre-trained models provided and our models trained on the training images of the RESIDE dataset as explained in Section~\ref{Proposed Approach}. 

In order to compare the performance of the learning based existing approaches with those of ours on the testing set of $350$ hazy images of the NR-haze dataset, we train the approaches using the corresponding training image set. The training of the variants of our approach and the existing approaches are done exactly with the same setup as mentioned in Section~\ref{TrainSetup}. Further, just like the second stage of training in our models, the models of existing approaches are trained for $500$ iterations ($2000$ batch updates each) with a batch size of $2$ using the Adam optimizer with the initial learning rate of $10^{-4}$, where the learning rate is decayed by half after every $100$ iterations.

\subsection{Quantitative Evaluation}
\label{sec: Quant_eval}
Here, we show the effectiveness of our PUG-D framework through quantitative evaluation of the $3$ variants of our dehazing approach. As we shall see below, while all the variants perform well quantitatively compared to the existing, PUG-D(M) consistently outperforms the state-of-the-art while dehazing both synthetic and real hazy images from different datasets.

\subsubsection{Evaluation on Synthetic Hazy Images}
\label{sec: Synthetic_Datasets}
The synthetic hazy images in LSIT, MSIT, HSIT, LSOT, MSOT, HSOT and SCHT sets of the NR-haze dataset presented  in Section~\ref{OurDataset}  and in SOTS of the popular RESIDE dataset are used to quantitatively compare the performance of the variants of the proposed approach with that of the state-of-the-art approaches mentioned earlier. As elaborated in Section~\ref{RSIDE},~\ref{OurDataset} and \ref{Existing}, the learning based approaches among those considered for the comparison are trained on the corresponding training sets.
The evaluation is performed based on the well-accepted structural similarity measure (SSIM) and peak-signal-to-noise ratio (PSNR), which are computed using the generated dehazed images and corresponding non-hazy references. We also use the CIEDE2000~\cite{sharma2005ciede2000} measure for the evaluation on our NR-haze dataset, which computes the pixel-wise color discrepancy between the dehazed and corresponding non-hazy images. While higher values of SSIM and PSNR indicate better performance, lower values of CIEDE2000 do the same.

Table~\ref{tab:table1} lists the measures obtained for the different approaches on the various test subsets of the NR-haze dataset. It is evident that the PUG-D(M) variant of our approach convincingly outperforms all the other approaches in terms of all the three measures, except the solitary case of PSNR on LSOT set. We also see that all the three variants of our approach score high in quantitative performance when tested in a broad spectrum of hazy images with varying levels of haze and color cast. The approach MSBDN of \cite{Dong_2020_CVPR} that uses end-to-end training is found to be the best existing approach by far, and the use of our PUG-D framework that employs the architecture of MSBDN in its feature extraction block produces even better results. Following existing literature, Table~\ref{tab:sotssyn} shows the PSNR and SSIM measures for the different approaches on SOTS of the RESIDE dataset. As can be seen, MSBDN and GridDNet perform significantly better than the other existing approaches. A similar trend is also seen in the performances of the $3$ different variants of our approach. Most importantly, our approaches PUG-D(M) and PUG-D(G) slightly outperform the state-of-the-art in terms of PSNR. In terms of SSIM, our approach's variants, MSBDN and GridDNet perform equally well producing results very close to the highest possible score.
\begin{table*}[t]
\centering
\caption{Ablation study of our PUG-D framework (using only $\mathcal{L}_{L1}$ in (\ref{loss_eq})) showing the performance improvements achieved on the NR-haze dataset by including its various components successively.  (Best: Bold red highlight, Second best: Blue highlight)}
\label{ablationModel}
\resizebox{\textwidth}{!}{%
\begin{tabular}{|l|c|c|c|c|c|c|c|}
\hline
\multicolumn{1}{|c|}{\multirow{3}{*}{Techniques}} &
  \multicolumn{6}{c|}{Non-cast Hazy Images} &
  Color Cast Hazy Image \\ \cline{2-8} 
\multicolumn{1}{|c|}{}                            & \begin{tabular}[c]{@{}c@{}}Outdoor\\ Low Haze\\(LSOT)\end{tabular}                                                   & \begin{tabular}[c]{@{}c@{}}Outdoor\\ Mid Haze\\(MSOT)\end{tabular} & \begin{tabular}[c]{@{}c@{}}Outdoor\\ High Haze\\(HSOT)\end{tabular}                                                  & \begin{tabular}[c]{@{}c@{}}Indoor\\ Low Haze\\(LSIT)\end{tabular}                                                   & \begin{tabular}[c]{@{}c@{}}Indoor\\ Mid Haze\\(MSIT)\end{tabular}                                                   & \begin{tabular}[c]{@{}c@{}}Indoor\\ High Haze\\(HSIT)\end{tabular} & \begin{tabular}[c]{@{}c@{}}Color Cast\\ Random Haze\\(SCHT)\end{tabular}                                                        \\ \cline{2-8}   
\multicolumn{1}{|c|}{} &
  \begin{tabular}[c]{@{}c@{}}PSNR/SSIM/\\ CIEDE2000\end{tabular} &
  \begin{tabular}[c]{@{}c@{}}PSNR/SSIM/\\ CIEDE2000\end{tabular} &
  \begin{tabular}[c]{@{}c@{}}PSNR/SSIM/\\ CIEDE2000\end{tabular} &
  \begin{tabular}[c]{@{}c@{}}PSNR/SSIM/\\ CIEDE2000\end{tabular} &
  \begin{tabular}[c]{@{}c@{}}PSNR/SSIM/\\ CIEDE2000\end{tabular} &
  \begin{tabular}[c]{@{}c@{}}PSNR/SSIM/\\ CIEDE2000\end{tabular} &
  \begin{tabular}[c]{@{}c@{}}PSNR/SSIM/\\ CIEDE2000\end{tabular} \\ \hline
\begin{tabular}[c]{@{}l@{}}Baseline-1\\ RESNet16\end{tabular} &
  \begin{tabular}[c]{@{}c@{}}17.43/0.7047\\ /53.24\end{tabular} &
  \begin{tabular}[c]{@{}c@{}}17.05/0.6867/\\ 53.66\end{tabular} &
  \begin{tabular}[c]{@{}c@{}}16.00/0.6188/\\ 55.84\end{tabular} &
  \begin{tabular}[c]{@{}c@{}}17.75/0.7014/\\ 46.70\end{tabular} &
  \begin{tabular}[c]{@{}c@{}}15.99/0.6258/\\ 50.92\end{tabular} &
  \begin{tabular}[c]{@{}c@{}}13.85/0.5100/\\ 58.11\end{tabular} &
  \begin{tabular}[c]{@{}c@{}}18.73/0.7167/\\ 52.50\end{tabular} \\ \hline
\begin{tabular}[c]{@{}l@{}}Baseline-2 \\ RESNet6+A+TM\end{tabular} &
  \begin{tabular}[c]{@{}c@{}}22.99/0.8737/\\ 38.48\end{tabular} &
  \begin{tabular}[c]{@{}c@{}}19.19/0.7704/\\ 47.90\end{tabular} &
  \begin{tabular}[c]{@{}c@{}}15.69/0.6640/\\ 55.88\end{tabular} &
  \begin{tabular}[c]{@{}c@{}}26.14/0.9244/\\ 24.85\end{tabular} &
  \begin{tabular}[c]{@{}c@{}}20.11/0.8071/\\ 38.30\end{tabular} &
  \begin{tabular}[c]{@{}c@{}}16.40/0.6853/\\ 49.94\end{tabular} &
  \begin{tabular}[c]{@{}c@{}}23.35/0.8287/\\ 44.65\end{tabular} \\ \hline
\begin{tabular}[c]{@{}l@{}}Baseline-3\\ RESNet6+A+TM+ISD\end{tabular} &
  \begin{tabular}[c]{@{}c@{}}23.16/0.8832/\\ 35.21\end{tabular} &
  \begin{tabular}[c]{@{}c@{}}19.76/0.7907/\\ 44.05\end{tabular} &
  \begin{tabular}[c]{@{}c@{}}17.37/0.7003/\\ 52.86\end{tabular} &
  \begin{tabular}[c]{@{}c@{}}\color{blue}27.44/0.9378/\\ \color{red}\textbf{21.63}\end{tabular} &
  \begin{tabular}[c]{@{}c@{}}22.28/0.8394/\\ 33.34\end{tabular} &
  \begin{tabular}[c]{@{}c@{}}18.25/0.7258/\\ 44.51\end{tabular} &
  \begin{tabular}[c]{@{}c@{}}23.99/0.8401/\\ 42.38\end{tabular} \\ \hline
\begin{tabular}[c]{@{}l@{}}Baseline-4\\ RESNet6+A+TM+PUD\end{tabular} &
  \begin{tabular}[c]{@{}c@{}}\color{blue}23.54/0.8934/\\ \color{blue}34.38\end{tabular} &
  \begin{tabular}[c]{@{}c@{}}\color{blue}22.17/0.8670/\\ \color{blue}38.27\end{tabular} &
  \begin{tabular}[c]{@{}c@{}}\color{blue}20.76/0.8214/\\ \color{blue}41.50\end{tabular} &
  \begin{tabular}[c]{@{}c@{}}27.37/0.9230/\\ 25.06\end{tabular} &
  \begin{tabular}[c]{@{}c@{}}\color{blue}25.08/0.8953/\\ \color{blue}30.44\end{tabular} &
  \begin{tabular}[c]{@{}c@{}}\color{blue}21.80/0.8172/\\ \color{blue}39.95\end{tabular} &
  \begin{tabular}[c]{@{}c@{}}\color{blue}24.78/0.8670/\\ \color{blue}39.67\end{tabular} \\ \hline
\begin{tabular}[c]{@{}l@{}}Baseline-5\\ RESNet6+A+TM+ISD+PUD\end{tabular} &
  \begin{tabular}[c]{@{}c@{}}\color{red}\textbf{23.85}/\textbf{0.9096}/\\ \color{red}\textbf{33.26}\end{tabular} &
  \begin{tabular}[c]{@{}c@{}}\color{red}\textbf{22.57}/\textbf{0.8883}/\\ \color{red}\textbf{36.67}\end{tabular} &
  \begin{tabular}[c]{@{}c@{}}\color{red}\textbf{21.61}/\textbf{0.8431}/\\ \color{red}\textbf{39.61}\end{tabular} &
  \begin{tabular}[c]{@{}c@{}}\color{red}\textbf{28.24}/\textbf{0.9405}/\\ \color{blue}22.04\end{tabular} &
  \begin{tabular}[c]{@{}c@{}}\color{red}\textbf{25.16}/\textbf{0.9070}/\\ \color{red}\textbf{27.38}\end{tabular} &
  \begin{tabular}[c]{@{}c@{}}\color{red}\textbf{21.90}/\textbf{0.8362}/\\ \color{red}\textbf{35.42}\end{tabular} &
  \begin{tabular}[c]{@{}c@{}}\color{red}\textbf{25.08}/\textbf{0.8818}/\\ \color{red}\textbf{38.19}\end{tabular} \\ \hline
\end{tabular}%
}
\end{table*}

\subsubsection{Evaluation on Images with Real Haze} 
\label{sec: real_dataset}

The images containing real haze from three different real-world benchmark datasets, O-Haze~\cite{ancuti2018Ohaze}, I-Haze~\cite{ancuti2018Ihaze}, and BeDDE~\cite{9099036} datasets discussed in Sections~\ref{RealDataset1} and \ref{RealDataset2}  are also used for quantitative comparison of the performance of the proposed approach's variants with that of the state-of-the-art approaches. As mentioned in those subsections, the pre-trained models provided by the respective authors of the learning-based approaches including ours are considered for the comparison.  Along with the PSNR and SSIM measures, the Visibility Index (VI) and the Realness Index (RI) by \cite{9099036} are also used, which are specifically designed to analyze the performance of dehazing algorithms in real hazy images. For both VI and RI, a higher value signifies a better performance.

Table~\ref{tab:real_quanti} gives the values of the measures obtained for the different approaches on the three datasets. For BeDDE dataset only VI and RI are shown, as typical ground truths required to calculate PSNR and SSIM are not available in this dataset~\cite{9099036}. It is apparent from the table that all the variants of our approach perform better than the state-of-the-art in all the cases, except PSNR and SSIM on the I-Haze dataset where our approach is found superior in terms of VI and RI. The values in the table indicate the effective dehazing performance of all the variants of our approach in terms of both visibility and artifact-free reconstruction. Among the proposed approach's variants which use different feature extraction blocks, we find that PUG-D(G) is slightly ahead in performance.
\begin{figure*}
    \centering
    \includegraphics[width=\linewidth]{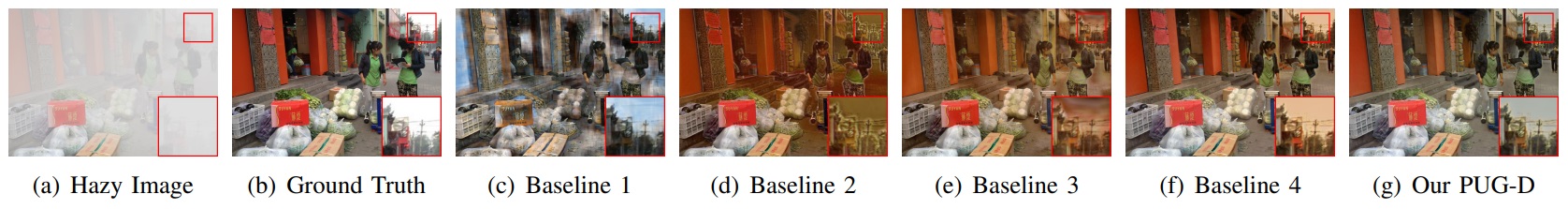}
    \caption{Ablation study of our PUG-D framework demonstrating the contributions of its components. (a) Hazy input, (b) Ground truth, (c) Baseline-1: RESNet16, (d) Baseline-2: RESNet6+A+TM, (e) Baseline-3: RESNet6+A+TM+ISD, (f) Baseline-4: RESNet6+A+TM+PUD, (e) Baseline-5 (our PUG-D framework): RESNet6+A+TM+ISD+PUD. (zoom for the best view)}
    \label{view}
\end{figure*} 

\subsection{Subjective Evaluation}
\label{sec: subjective_eval}
Here, we compare the dehazing results of our approach with a few effective state-of-the-art approaches qualitatively. We only consider the results of the PUG-D(R6) variant of our approach here, which uses the simplest feature extraction block among all the variants. In Section IV-B of the \textit{supplementary}, we compare the dehazing results of the three variants qualitatively with each other, where we find them to be almost equally good.

\subsubsection{Evaluation on Synthetic Hazy Images} 
\label{sec: subjective_synthetic}

In Fig.~\ref{fig: sythetic}, we show the dehazing results of the various approaches including ours on synthetically generated hazy images. Among the approaches for comparison mentioned at the beginning of this section, we consider EPix2Pix, GridDNet, ALC, FSID, MSBDN, and IDE in the figure as they perform the best qualitatively on images of dataset considered. The images in Fig.~\ref{fig: sythetic} contain synthetic haze of varying amounts and a few of them also contain synthetic color cast. The images are taken from the NR-haze dataset considering one each from its $6$ test image sets having indoor and outdoor images with different haze densities and $3$ from its SCHT set having different kinds of color cast. As mentioned in Section~\ref{OurDataset}, the learning-based approaches, whose dehazing results are shown, are trained on the related training sets of the NR-haze dataset. The first three images in Fig.~\ref{fig: sythetic} are hazy images with yellow, cyan and blue casts, respectively. These color casts are shown, as similar casts naturally occur in images captured in haze (see first three row images in Fig.~\ref{fig: real}). The remaining six images are non-cast hazy images, where the first three are of indoor scenes, and the last three are of outdoor scenes.

From the results generated on the color cast hazy images by the existing approaches, we see that many of them do not remove the color casts by a significant amount and in all image regions. As can be seen from the dehazed results obtained using our PUG-D(R6) on the color-cast hazy images, color casts and haze are satisfactorily removed and visually realistic dehazed images close to the ground truths are generated, outperforming the others. 

Considering the dehazing performance of the approaches on the rest of the hazy images in the figure, we can see that the amount of haze reduced by the existing techniques is limited as compared to that of our PUG-D(R6). A few existing approaches perform quite well for indoor images, however our approach does better. So, for the synthetic indoor and outdoor hazy images in Figure~\ref{fig: sythetic} having different amounts of haze, our approach is found to remove haze substantially without introducing noticeable distortion and it produces images close to the ground truths. MSBDN from the state-of-the-art is the closest to our approach in dehazing performance as per visual observations with respect to the ground truths.

\subsubsection{Evaluation on Real-world Hazy Images}
\label{sec: subjective_real}

In Fig.~\ref{fig: real}, we qualitatively compare the performance of our PUG-D(R6) with a few other state-of-the-art techniques on real-world hazy images. We show the results of EPix2Pix, GridDNet, ALC, FSID, MSBDN, IDE, RefineDNet, and D4 among the approaches mentioned earlier, as they perform the best qualitatively on images with real haze. The real-world hazy images are considered from the various sources as mentioned in Section~\ref{RealDataset3}, where it is also mentioned that the results of learning-based approaches considered in Fig.~\ref{fig: real} are generated using their pre-trained models provided by the authors. The first row shows results on a real hazy image having a green color cast~\cite{peng2019image}. The results in the second row are for a real hazy image with a yellow-red color cast due to sandstorm~\cite{peng2019image}. The third row shows results on a real hazy image with a bluish color cast~\cite{choi2015referenceless}. The rest of the real hazy images in the figure contain varied amounts of haze without significant color cast. 

As evident from the dehazed results of the existing approaches on real color cast hazy images, most of them do not remove color casts substantially. Color distortion is also introduced by a few of them, and in a couple of dehazed results, we see loss of object color. Our PUG-D(R6) removes color casts substantially along with haze and maintains visually realistic object color without introducing visible color distortion or loss.
 
Considering the dehazing performance  of  all  the  approaches on all the images in the figure, we see that the amount of dehazing by the existing approaches, particularly in regions with thick haze, is limited compared to our PUG-D(R6). In a few cases, color artifacts are evident in the results by the existing approaches, which include non-realistic reproduction of color (like bluish color in place of green), unlike the results of our method. Our approach produces better dehazing results for all the images reducing haze substantially and producing visually realistic output with faithful color reconstruction. This is true for the critical hazy images in the fourth and seventh rows as well, where dense haze is present in distant areas in the former and the atmospheric light estimation is difficult in the latter owing to the absence of sky region. 

From the above analysis on a variety of real-world hazy images, we can see that our PUG-D(R6) performs effective dehazing and color cast removal in a variety of hazy conditions outperforming the other techniques. We also discuss the model complexity of PUG-D(R6) in Section IV-A of the \textit{supplementary}.

\section{Additional Study}
\subsection{Ablation study of our PUG-D framework}
\label{sec: ablation}

We present an ablation study of our dehazing framework using the NR-haze dataset in Table~\ref{ablationModel}. We use five different complete dehazing model baselines whose outputs are dehazed images. They are Baseline-1 (RESNet16): sixteen consecutive residual blocks forming an image-to-image mapping network that takes the hazy image as the input, Baseline-2 (RESNet6+A+TM):  six consecutive residual blocks, which take the estimated transmission map and atmospheric light along with the hazy image as inputs, 
Baseline-3 (RESNet6+A+TM+ISD): Baseline 2 along with the ISD module, Baseline-4 (RESNet6+A+TM+PUD): Baseline-2 along with our proposed joint progressive updating \& dehazing (PUD), (V) Baseline-5 (RESNet6+A+TM+PUD+ISD): Baseline-2 along with ISD and PUD. For a fair comparison, we apply only mean absolute error ($L_1$) as the loss function to train the models, and we perform all the experiments using the first and second stages of our training process. We can see from Table~\ref{ablationModel} that Baseline-2 having the transmission map and atmospheric light with less number of residual blocks as compared to the Baseline-1 produces significantly improved results for different types of hazy conditions. These experimental findings support the use of estimated transmission map and atmospheric light for guidance while performing end-to-end training for dehazing. Baseline-3 with the additional ISD module over Baseline-2 provides a minor improvement in the results. ISD introduces inter-step dependencies in the feature layers to provide a performance improvement. However, in most cases, the improvements are much more prominent in Baseline-4, where PUD is used in addition to Baseline-2. Our PUD successfully restricts huge performance drop with the increase in haze density, and this experimentally proves the effectiveness of our proposed progressive updating and dehazing procedure. Baseline-5, therefore includes PUD with the Baseline-4 to provide the best performance in almost all the cases. Note that, Baseline-5 is essentially our proposed framework PUG-D(R6) trained using only $L_1$ loss and two of our three training stages.

We further show a qualitative evaluation of the different baselines of our model on a synthetic hazy image in Fig.~\ref{view}. We witness that, except Baseline-5, all the baselines suffer from unpleasant artifacts. Baseline-4 is relatively better in handling the artifacts but suffers from color distortion visible in the cropped regions. However, Baseline-5 representing our PUG-D framework produces a visually realistic dehazed output close to the ground truth.

\begin{table*}
\centering
\caption{Dehazing performance comparison of the atmospheric light scattering model (ALSM) with the dehazing model of our proposed PUG-D, where both take estimates of transmission map and atmospheric light as inputs along with the hazy image. (Best: Bold highlight)}
\label{tab: formulavsIPUDN}
\resizebox{\textwidth}{!}{%
\begin{tabular}{|c|c|c|c|c|c|c|c|}
\hline
\multirow{3}{*}{Techniques} &
  \multicolumn{6}{c|}{Non-cast Hazy Images} &
  Color Cast Hazy Image \\ \cline{2-8} 
 &
  \begin{tabular}[c]{@{}c@{}}Outdoor\\ Low Haze\end{tabular} &
  \begin{tabular}[c]{@{}c@{}}Outdoor\\ Mid Haze\end{tabular} &
  \begin{tabular}[c]{@{}c@{}}Outdoor\\ High Haze\end{tabular} &
  \begin{tabular}[c]{@{}c@{}}Indoor\\ Low Haze\end{tabular} &
  \begin{tabular}[c]{@{}c@{}}Indoor\\ Mid Haze\end{tabular} &
  \begin{tabular}[c]{@{}c@{}}Indoor\\ High Haze\end{tabular} &
  \begin{tabular}[c]{@{}c@{}}Color Cast\\ Random Haze\end{tabular} \\ \cline{2-8} 
 &
  \begin{tabular}[c]{@{}c@{}}PSNR/SSIM/\\ CIEDE2000\end{tabular} &
  \begin{tabular}[c]{@{}c@{}}PSNR/SSIM/\\ CIEDE2000\end{tabular} &
  \begin{tabular}[c]{@{}c@{}}PSNR/SSIM/\\ CIEDE2000\end{tabular} &
  \begin{tabular}[c]{@{}c@{}}PSNR/SSIM/\\ CIEDE2000\end{tabular} &
  \begin{tabular}[c]{@{}c@{}}PSNR/SSIM/\\ CIEDE2000\end{tabular} &
  \begin{tabular}[c]{@{}c@{}}PSNR/SSIM/\\ CIEDE2000\end{tabular} &
  \begin{tabular}[c]{@{}c@{}}PSNR/SSIM/\\ CIEDE200\end{tabular} \\ \hline
\begin{tabular}[c]{@{}c@{}}ALSM based\\ dehazing\end{tabular} &
  \begin{tabular}[c]{@{}c@{}}24.56/ 0.9086/\\ 29.96\end{tabular} &
  \begin{tabular}[c]{@{}c@{}}21.06/ 0.8214/\\ 40.68\end{tabular} &
  \begin{tabular}[c]{@{}c@{}}15.94/ 0.6739/\\ 50.22\end{tabular} &
  \begin{tabular}[c]{@{}c@{}}27.84/ 0.9264/\\ 20.60\end{tabular} &
  \begin{tabular}[c]{@{}c@{}}21.47/ 0.8230/\\ 34.26\end{tabular} &
  \begin{tabular}[c]{@{}c@{}}17.39/ 0.7012/\\ 45.23\end{tabular} &
  \begin{tabular}[c]{@{}c@{}}22.97/0.8021/\\ 48.17\end{tabular} \\ \hline
\begin{tabular}[c]{@{}c@{}}Our \\ PUG-D\end{tabular} &
  \textbf{\begin{tabular}[c]{@{}c@{}}25.83/0.9430/\\ 22.44\end{tabular}} &
  \textbf{\begin{tabular}[c]{@{}c@{}}24.43/0.9220/\\ 22.77\end{tabular}} &
  \textbf{\begin{tabular}[c]{@{}c@{}}22.81/ 0.8822/\\ 26.76\end{tabular}} &
  \textbf{\begin{tabular}[c]{@{}c@{}}30.02/0.9645/\\ 12.44\end{tabular}} &
  \textbf{\begin{tabular}[c]{@{}c@{}}27.17/0.9401/\\ 16.04\end{tabular}} &
  \textbf{\begin{tabular}[c]{@{}c@{}}23.94/0.8902/\\ 21.95\end{tabular}} &
  \textbf{\begin{tabular}[c]{@{}c@{}}26.74/ 0.9157/\\ 28.92\end{tabular}} \\ \hline
\end{tabular}%
}
\end{table*}

\begin{figure}
    \centering
    \includegraphics[width=\linewidth]{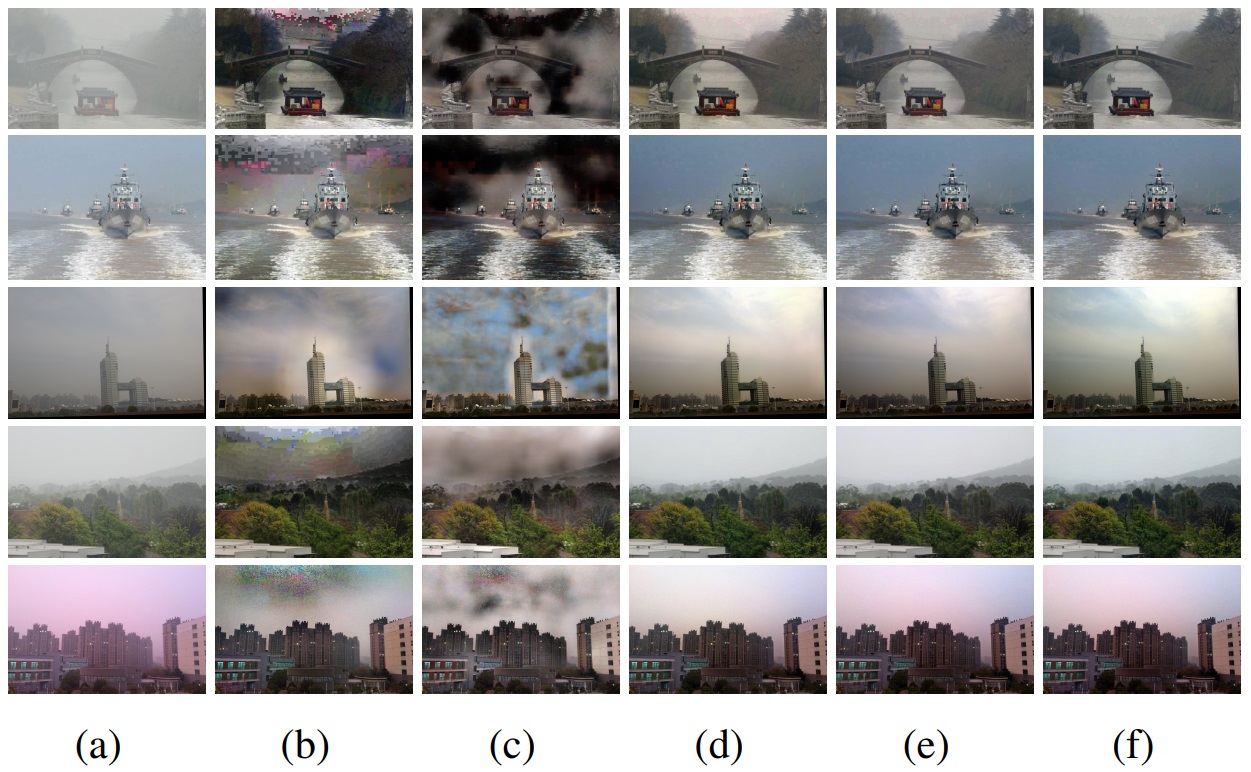}
    \caption{Appearence of artifacts during dehazing. (a) Real Hazy Image (b) MSBDN (c) GridDNet (d) PUG-D(M) (e) PUG-D(G) (f) PUG-D(R6)}
    \label{fig: artefacts}
\end{figure}

\subsection{Atmospheric Light Scattering Model (vs) Separate Dehazing Network}
\label{alsmvsDN}
Table~\ref{tab: formulavsIPUDN} shows results on the NR-haze dataset of a study justifying the use of a separate dehazing network in our proposed PUG-D framework instead of the Koschmieder's atmospheric light scattering model for dehazing. In the case of use of the atmospheric light scattering model for dehazing, we estimate the transmission map and atmospheric light using our densely connected encoder-decoder network and the proposed convolutional neural network based model, respectively. We then reconstruct the dehazed image using~\ref{eq_dehaze} and optimize both the network using the reconstruction loss along with transmission map and atmospheric light estimation loss, similar to the process employed by a few existing approaches \cite{zhang2018densely, Deng_2019_ICCV}. In the table, for a wide range of hazy conditions, we observe the superior dehazing performance of our proposed approach (PUG-D(R6)) of employing a separate dehazing module that takes initial estimates of transmission map and atmospheric light as inputs along with the hazy image. This signifies that our separate dehazing module containing the updater networks successfully handles any insufficiency in the initial estimates of transmission map and atmospheric light.

\subsection{PUG-D's Image Artifact Handling Capability} 
Here we discuss the substantial effectiveness demonstrated by our approach in handling image artifacts during dehazing. Fig.~\ref{fig: artefacts} shows a few real-world hazy images from Section~\ref{RealDataset3} in which the best existing dehazing approaches MSBDN and GridDNet produce or boost artifacts. The artifacts already present such as those in the images of the first two rows of the figure may be boosted by the two existing approaches and they may generate strong halo-like artifacts as well like those in the images of the figure's third and fourth rows (see Fig.~\ref{fig: artefacts}.(b) and (c)). As can be seen from Fig.~\ref{fig: artefacts}.(d), (e) and (f), all the three variants of our approach do not produce or boost such noticeable artifacts. This is in spite of considering the feature extraction architectures of MSBDN and GridDNet in two of the variants, which clearly demonstrates the significance and effectiveness of our PUG-D framework especially in handling image artifacts during dehazing.

A few more additional experiments and studies are given in the Sections~III and~IV of the \textit{supplementary}.

\section{Conclusion}
A single image dehazing framework is proposed in this paper which involves progressive dehazing jointly with the updating of the transmission map and atmospheric light values. Our unique multi-network framework contains a channel-wise atmospheric light estimation network that allows handling of color cast in hazy images. It also contains novel transmission map and atmospheric light updater networks that allow the handling of insufficiency in their initial estimates. A novel dehazing network interdependent with the updater networks outputs the dehazed image from the framework in a progressive manner, where inter-step dependencies are maintained. The dehazing module with the updater and dehazing networks is trained end-to-end allowing effective dehazing.

Our proposed progressive update guided dehazing framework is designed to work in a wide variety of real-world hazy conditions with different amounts of haze and color casts. Our dehazing approach is experimentally found to perform effectively for both indoor and outdoor hazy images. It is also found to outperform the state-of-the-art in general on real-world and synthetic hazy images both qualitatively and quantitatively. The dehazed results produced by our approach are seen to provide enhanced visibility by ensuring visually realistic and faithful restoration without the introduction of distortion. 
The studies also show that  that our progressive updating and dehazing strategy involving end-to-end training with haze parameter guidance is more effective than just end-to-end learning of an image-to-image mapping. We also find that our use of a separate dehazing module, with transmission map and atmospheric light guidance, is effective compared to the use of haze parameter estimates in a predefined model for dehazing. The proposed framework is also observed to be highly effective in managing image artifacts while dehazing.

\begin{center}
 {\larger[4] \color{red}Supplementary Material\newline}
\end{center}

\setcounter{section}{0}
\setcounter{figure}{0}
\setcounter{table}{0}

\section{Description of the NR-haze Dataset}

We form a new dataset of indoor and outdoor hazy images taking a cue from the procedure followed to build the RESIDE dataset~\cite{li2018benchmarking}. Indoor images associated with depth maps are considered from the NYU Depth dataset V2~\cite{silberman2012indoor}. Outdoor images are taken from the Outdoor Training Set (OTS) and Synthetic Objective Testing Set (SOTS) of the RESIDE dataset. As we employ parts of both the NYU and RESIDE datasets, we refer to the new dataset as the NR-haze
dataset, which is available at {\color{magenta}\href{https://aupendu.github.io/progressive-dehaze}{aupendu.github.io/progressive-dehaze}}. We discuss the motivation behind generating a separate dataset of images with synthetic haze below first, and then describe its formation elaborately.

\subsection{Motivation}
Popular existing synthetic hazy image datasets such as the RESIDE dataset use the Koschmieder's model (see Section~III-A of the main paper) on haze-free images to generate the synthetic haze. Atmospheric light and transmission map parameters of the model are provided for the generation, where the transmission map is computed based on the depth maps of the images. Therefore, accurate image depth maps are of utmost importance to generate an appropriate synthetic hazy image dataset.

\begin{figure*}
    \centering
    \includegraphics[width=0.8\linewidth]{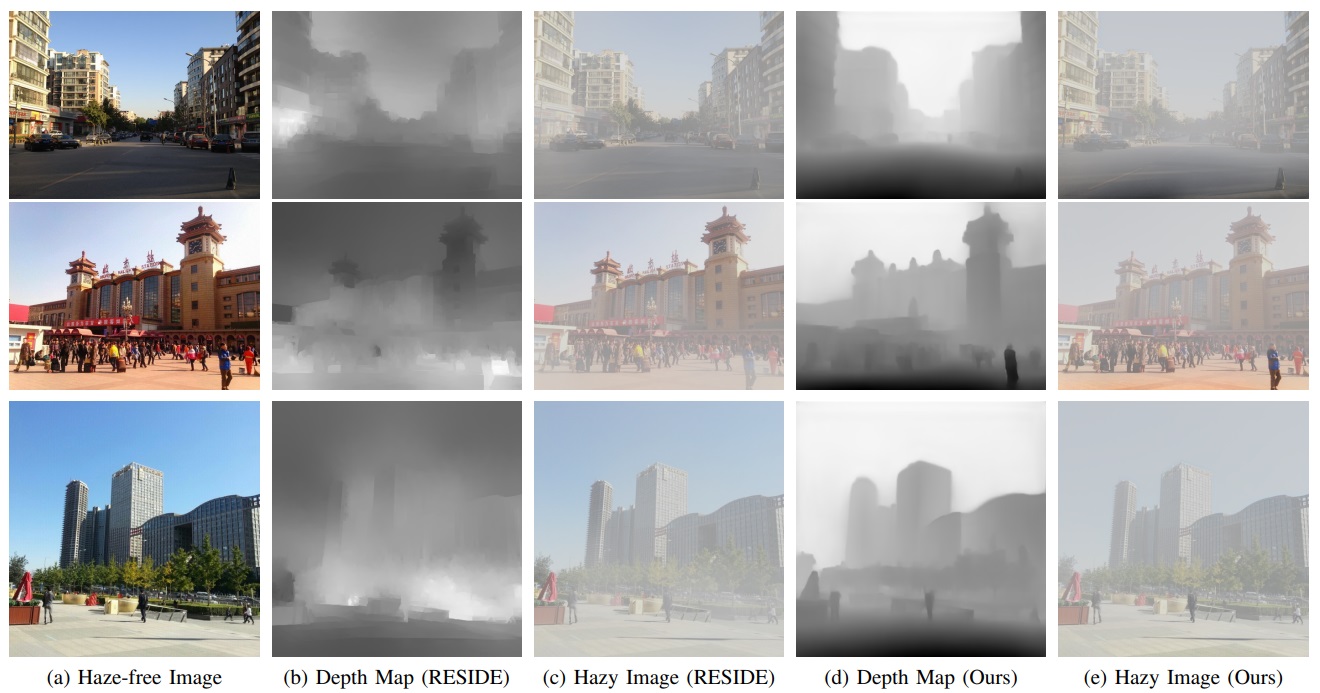}
    \caption{(c) \& (e) Hazy images generated from the (a) haze-free outdoor images applying the Koschmieder's model using the (b) \& (d) depth maps, respectively. Depth maps (RESIDE) are as given in the RESIDE dataset and Depth maps (Ours) are generated using~\cite{li2018megadepth} for our NR-haze dataset. [Depth map grading: darker is nearer, brighter is farther]}
    \label{fig: depth}
\end{figure*}

Classical image depth estimation approaches, such as the one used to generate the OTS of the RESIDE dataset, are not the best available and may produce erroneous depth maps. Such examples of erroneous depth maps are shown in Fig.~\ref{fig: depth}(b), where it is evident that lower depth values are assigned to distant regions. The resulting hazy images using the Koschmieder's model are shown in Fig.~\ref{fig: depth}(c). As can be seen, erroneous depth maps have produced improper transmission maps that resulted in uncharacteristic hazy images.

As the use of such uncharacteristic synthetic hazy images may result in improper training of a system, we create the NR-haze dataset, where we employ depth values estimated by the state-of-the-art approach of~\cite{li2018megadepth} for outdoor images. Fig.~\ref{fig: depth}(d) shows the depth maps used by us for the example outdoor images and Fig.~\ref{fig: depth}(e) shows corresponding appropriate hazy images generated. For indoor images, the sensor-captured depth data available is used.

In addition, existing synthetic hazy image datasets like the RESIDE dataset do not contain color cast hazy images. As color casts are sometimes associated with natural hazy images, for a comprehensive representation, we include synthetic hazy images with different color casts in the NR-haze dataset. Further, unlike any existing synthetic hazy image dataset, the synthetic hazy images for testing in the NR-haze dataset are classified in accordance to the amounts of haze in them for additional insight into an approach's quantitative performance.

The NR-haze dataset formation is elaborately described in the next subsection.

\begin{figure*}
    \centering
    \includegraphics[width=0.8\linewidth]{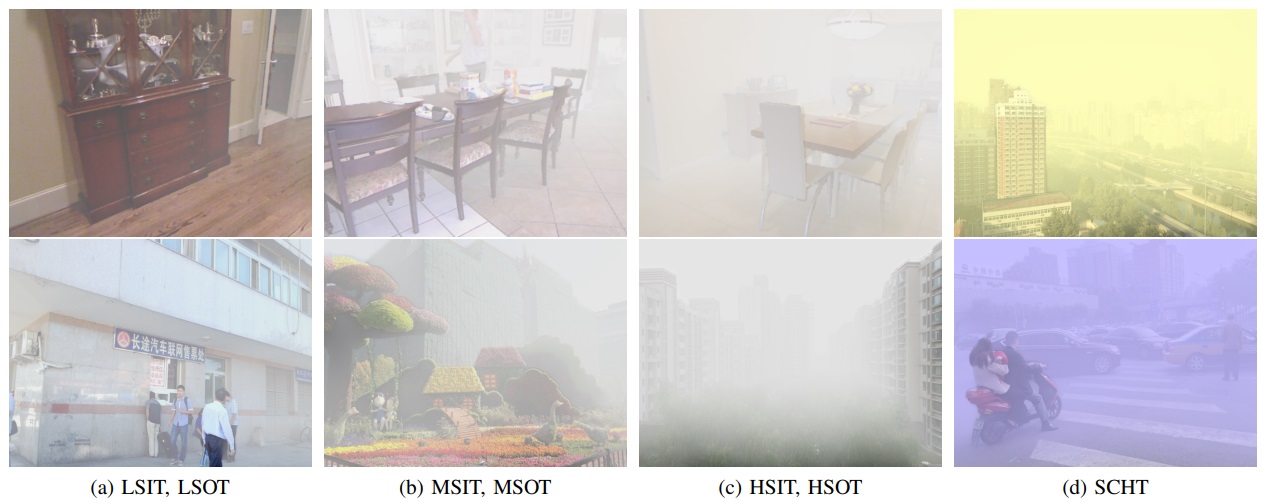}
    \caption{Sample hazy images from the NR-haze dataset. The first row images, except the rightmost, are from LSIT, MSIT and HSIT that represent indoor images with low, mid and high haze, respectively. The second row images, except the rightmost, are from LSOT, MSOT and HSOT that represent outdoor images with low, mid and high haze, respectively. The color cast hazy images in the last column are taken from SCHT.}
    \label{fig: sampledata}
\end{figure*}

\subsection{Dataset Generation}

\subsubsection{The Images}

The above discussion related to inaccurate depth maps and the absence of color cast hazy images motivates us to generate a separate dataset of synthetic hazy images, the NR-haze dataset. To do so, we collect indoor images from NYU Depth dataset V2~\cite{silberman2012indoor} and use the given depth maps which have been captured using a depth sensor. Outdoor images are collected from OTS and SOTS of the RESIDE dataset~\cite{li2018benchmarking} and we generate the depth maps using the state-of-the-art technique of~\cite{li2018megadepth}. Values in the depth maps range from $0$ to $\sim10$ for indoor images and range from $0$ to $1$ for outdoor images. In total, $1449$ indoor images and $2478$ outdoor images are collected from the said datasets. Among those images, $1349$ indoor images are used for the training set, $50$ for the validation set and $50$ to generate multiple testing sets. Similarly, $50$ outdoor images are used for the validation set, $50$ to generate multiple testing sets, and $2378$ are used for the training set.

\subsubsection{Training and Validation}
\label{training}
To generate hazy images for training, the indoor and outdoor images marked for the training set are considered. From the depth map $d$ associated with an image for training, we produce the transmission map $t$ as $t(x)=e^{-\beta d(x)}$ (see Section~III-A in the main paper), where $\beta$ is the attenuation coefficient that determines the amount of haze at a depth $d(x)$. To cover a wide range of hazy conditions, we choose $\beta$ randomly in the range of $0.2-0.8$ for indoor images and $2.0-5.0$ for outdoor images. Finally, we use the generated transmission map and a randomly chosen $A$ from the range of $0.3-1.0$ in the Koschmieder's atmospheric light scattering model (see Expression (1) in the main paper) to produce a hazy image channel $I$ from a non-hazy image channel $J$. Note that, the transmission map $t$ is reused to generate the three channels of the hazy image. But, the values of $A$ for two of the three channels are randomly chosen to deviate
within the range of $0-40\%$ from the value of $A$ chosen earlier for the third channel. When the random deviation results in a 3-channel atmospheric light vector with substantially different element values, color cast hazy images are produced. Hazy images for validation are generated in a way similar to those for training using the $50$ indoor and $50$ outdoor images marked for validation. 

\subsubsection{Testing}

The $50$ indoor and $50$ outdoor images marked for testing are used to $7$ different test sets. They are called the Low-haze Synthetic Indoor Test (LSIT) set, Mid-haze Synthetic Indoor Test (MSIT) set, High-haze Synthetic Indoor Test (HSIT) set, Low-haze Synthetic Outdoor Test (LSOT) set, Mid-haze Synthetic Outdoor Test (MSOT) set, High-haze Synthetic Outdoor Test (HSOT) set and Synthetic Color-cast Haze Test-set (SCHT) and they comprise of $350$ hazy images in total with $50$ in each of them. Among the $7$ test sets, only SCHT contains hazy images with color cast. As evident from the names, we produce three different levels of haze in both indoor and outdoor images to generate the $6$ test sets of hazy images without color cast. The Koschmieder's model is used to generate these hazy images as explained in Section~\ref{training}. The $\beta$ values are randomly chosen in the range of $2-2.5$, $3.25-3.75$ and $4.5-5.0$ to generate low, mid and high haze respectively in the outdoor images, and we randomly choose the $\beta$ values in the range of $0.2-0.3$, $0.45-0.55$ and $0.7-0.8$ to generate low, mid and high haze respectively in the indoor images. To ensure that the hazy images are cast-free, we randomly choose one value of $A$ from the range $0.85-0.95$ and use it for all the $3$ channels of the atmospheric light. SCHT is created by generating color-cast hazy images considering the $50$ outdoor images. Haze and color cast are produced using the Koschmieder's model where $\beta$ values are randomly chosen in the range of $3.0-4.0$. The value of $A$ is randomly chosen from the range $0.3-0.1$ for use in one of the atmospheric light channels. In the other two channels, values which deviate in the range of $10\%$ to $40\%$ from $A$ is randomly chosen. Examples of synthetic hazy images from the $7$ sets are shown in Fig.~\ref{fig: sampledata}.
\begin{table*}
\centering
\caption{Performance of our dehazing approach PUG-D for various pixel-wise loss functions used along with the perceptual loss. (Best: Bold highlight)}
\label{ablationLoss}
\resizebox{\textwidth}{!}{%
\begin{tabular}{|c|c|c|c|c|c|c|c|}
\hline
\multirow{3}{*}{\begin{tabular}[c]{@{}c@{}}\\Loss \\ Function\end{tabular}} &
  \multicolumn{6}{c|}{Non-cast Hazy Images} &
  Color Cast Hazy Image \\ \cline{2-8} 
 &
  \begin{tabular}[c]{@{}c@{}}Outdoor\\ Low Haze\end{tabular} &
  \begin{tabular}[c]{@{}c@{}}Outdoor\\ Mid Haze\end{tabular} &
  \begin{tabular}[c]{@{}c@{}}Outdoor\\ High Haze\end{tabular} &
  \begin{tabular}[c]{@{}c@{}}Indoor\\ Low Haze\end{tabular} &
  \begin{tabular}[c]{@{}c@{}}Indoor\\ Mid Haze\end{tabular} &
  \begin{tabular}[c]{@{}c@{}}Indoor\\ High Haze\end{tabular} &
  \begin{tabular}[c]{@{}c@{}}Color Cast\\ Random Haze\end{tabular} \\ \cline{2-8} 
 &
  \begin{tabular}[c]{@{}c@{}}PSNR/SSIM/\\ CIEDE2000\end{tabular} &
  \begin{tabular}[c]{@{}c@{}}PSNR/SSIM/\\ CIEDE2000\end{tabular} &
  \begin{tabular}[c]{@{}c@{}}PSNR/SSIM/\\ CIEDE2000\end{tabular} &
  \begin{tabular}[c]{@{}c@{}}PSNR/SSIM/\\ CIEDE2000\end{tabular} &
  \begin{tabular}[c]{@{}c@{}}PSNR/SSIM/\\ CIEDE2000\end{tabular} &
  \begin{tabular}[c]{@{}c@{}}PSNR/SSIM/\\ CIEDE2000\end{tabular} &
  \begin{tabular}[c]{@{}c@{}}PSNR/SSIM/\\ CIEDE2000\end{tabular} \\ \hline
MSE Loss &
  \begin{tabular}[c]{@{}c@{}}23.61/0.8939/\\ 36.32\end{tabular} &
  \begin{tabular}[c]{@{}c@{}}22.98/0.8708/\\ 39.82\end{tabular} &
  \begin{tabular}[c]{@{}c@{}}21.46/0.8043/ \\ 46.06\end{tabular} &
  \begin{tabular}[c]{@{}c@{}}27.54/0.9226/\\ 26.86\end{tabular} &
  \begin{tabular}[c]{@{}c@{}}24.79/0.8853/\\ 32.38\end{tabular} &
  \begin{tabular}[c]{@{}c@{}}\textbf{22.10}/0.8167/\\ 39.70\end{tabular} &
  \begin{tabular}[c]{@{}c@{}}24.88/0.8708/\\ 41.22\end{tabular} \\ \hline
\begin{tabular}[c]{@{}c@{}}Recursive\\ L1 Loss\end{tabular} &
\begin{tabular}[c]{@{}c@{}}\textbf{24.03}/\textbf{0.9141}/\\ 33.77\end{tabular} &
\begin{tabular}[c]{@{}c@{}}\textbf{23.15}/\textbf{0.8937}/\\ 37.80\end{tabular} &
\begin{tabular}[c]{@{}c@{}}21.09/0.8313/\\ 41.66\end{tabular} &
\begin{tabular}[c]{@{}c@{}}28.07/\textbf{0.9408}/\\ 22.46\end{tabular} &
\begin{tabular}[c]{@{}c@{}}25.08/0.9028/\\ 28.38\end{tabular} &
\begin{tabular}[c]{@{}c@{}}21.22/0.8257/\\ 37.64\end{tabular} &
\begin{tabular}[c]{@{}c@{}}\textbf{25.09}/0.8778/\\ 39.90\end{tabular} \\ \hline
L1 Loss &
  \begin{tabular}[c]{@{}c@{}}23.85/0.9096/\\ \textbf{33.26}\end{tabular} &
  \begin{tabular}[c]{@{}c@{}}22.57/0.8883/\\ \textbf{36.67}\end{tabular} &
  \begin{tabular}[c]{@{}c@{}}\textbf{21.61}/\textbf{0.8431}/ \\ \textbf{39.61}\end{tabular} &
  \begin{tabular}[c]{@{}c@{}}\textbf{28.24}/0.9405/\\ \textbf{22.04}\end{tabular} &
  \begin{tabular}[c]{@{}c@{}}\textbf{25.16}/\textbf{0.9070}/\\ \textbf{27.38}\end{tabular} &
  \begin{tabular}[c]{@{}c@{}}21.90/\textbf{0.8362}/\\ \textbf{35.42}\end{tabular} &
  \begin{tabular}[c]{@{}c@{}}25.08/\textbf{0.8818}/\\ \textbf{38.19}\end{tabular} \\ \hline
\end{tabular}
}
\end{table*}

\section{The Inter-step dependency (ISD)  Layer of Our Approach}
In our work, we use convolutional LSTM~\cite{xingjian2015convolutional} as shown in (\ref{LSTM_eq}) to implement the inter-step dependency (ISD) block. At time step $t$, the  layer receives features from the input feature extraction block and its previous state at time step $t-1$ of the entire module. It is well known that to obtain the state $h(t)$ from $h(t-1)$ and the input $X(t-1)$, an input gate $i(t)$, a forget gate $f(t)$, an output gate $o(t)$ and a cell state $c(t)$ are computed as follows:
\begin{equation}
\begin{aligned}
    y(t) = f_{in}(X(t)),\\
    i(t)=\sigma(W_{iy}\otimes y(t)+W_{is}\otimes h(t-1)+b_{i}),\\
    f(t)=\sigma(W_{fy}\otimes y(t)+W_{fs}\otimes h(t-1)+b_{f}),\\
    o(t)=\sigma(W_{oy}\otimes y(t)+W_{os}\otimes h(t-1)+b_{o}),\\
    g(t)=\tanh(W_{gy}\otimes y(t)+W_{gs}\otimes h(t-1)+b_{g}),\\
    c(t)=f(t)\circledcirc c(t-1)+i(t)\circledcirc g(t),\\
    h(t)=o(t)\circledcirc \tanh (c(t))
    \label{LSTM_eq}
\end{aligned}
\end{equation}
where $t$ indicates the current time step of the entire dehazing module and not the ISD block alone, $\sigma$ is the sigmoid function, $\tanh$ is the hyperbolic tangent function, $\circledcirc$ is element-wise multiplication, and $\otimes$ is the convolution operation.

\section{Additional Experiments on the Design of Our Approach}

\subsection{$L_{1}$ loss (vs) MSE:}
Table~\ref{ablationLoss} shows the comparison of MSE and $L_{1}$ loss functions for our approach using the NR-haze dataset leading to our choice in Section~III-C4 of the main paper. We impose both the losses on the final dehazed output. As can be seen, $L_{1}$ loss is experimentally found to be superior to MSE. We also perform experiments with recursive supervision (loss computation at every time-step), and notice that both $L_{1}$ and recursive $L_{1}$ losses are better than the use of MSE. However, we choose $L_{1}$ loss over recursive $L_{1}$ loss as it is computationally economical, and gives consistently better performance in terms of the CIEDE2000 measure. This possibly means supervision at the final iteration helps to preserve the color information in a better way than recursive supervision.
\begin{table*}
\centering
\caption{A study on the suitable number of time steps /iterations in our iterative dehazing framework PUG-D(R6). (Best: Bold highlight)}
\label{ablationLSTM}
\resizebox{\textwidth}{!}{%
\begin{tabular}{|c|c|c|c|c|c|c|c|}
\hline
\multirow{3}{*}{\begin{tabular}[c]{@{}c@{}}\\Recursive \\ Time-steps\end{tabular}} &
  \multicolumn{6}{c|}{Non-cast Hazy Images} &
  Color Cast Hazy Image \\ \cline{2-8} 
 &
  \begin{tabular}[c]{@{}c@{}}Outdoor\\ Low Haze\end{tabular} &
  \begin{tabular}[c]{@{}c@{}}Outdoor\\ Mid Haze\end{tabular} &
  \begin{tabular}[c]{@{}c@{}}Outdoor\\ High Haze\end{tabular} &
  \begin{tabular}[c]{@{}c@{}}Indoor\\ Low Haze\end{tabular} &
  \begin{tabular}[c]{@{}c@{}}Indoor\\ Mid Haze\end{tabular} &
  \begin{tabular}[c]{@{}c@{}}Indoor\\ High Haze\end{tabular} &
  \begin{tabular}[c]{@{}c@{}}Color Cast\\ Random Haze\end{tabular} \\ \cline{2-8} 
 &
  \begin{tabular}[c]{@{}c@{}}PSNR/SSIM/\\ CIEDE2000\end{tabular} &
  \begin{tabular}[c]{@{}c@{}}PSNR/SSIM/\\ CIEDE2000\end{tabular} &
  \begin{tabular}[c]{@{}c@{}}PSNR/SSIM/\\ CIEDE2000\end{tabular} &
  \begin{tabular}[c]{@{}c@{}}PSNR/SSIM/\\ CIEDE2000\end{tabular} &
  \begin{tabular}[c]{@{}c@{}}PSNR/SSIM/\\ CIEDE2000\end{tabular} &
  \begin{tabular}[c]{@{}c@{}}PSNR/SSIM/\\ CIEDE2000\end{tabular} &
  \begin{tabular}[c]{@{}c@{}}PSNR/SSIM/\\ CIEDE2000\end{tabular} \\ \hline
3 &
  \begin{tabular}[c]{@{}c@{}}25.73/0.9305/\\ 23.73\end{tabular} &
  \begin{tabular}[c]{@{}c@{}}23.60/0.9023/\\ 27.61\end{tabular} &
  \begin{tabular}[c]{@{}c@{}}22.07/0.8453/\\ 32.78\end{tabular} &
  \begin{tabular}[c]{@{}c@{}}\textbf{30.04}/0.9633/\\ 13.57\end{tabular} &
  \begin{tabular}[c]{@{}c@{}}25.52/0.9318/\\ 17.64\end{tabular} &
  \begin{tabular}[c]{@{}c@{}}21.69/0.8679/\\ 24.75\end{tabular} &
  \begin{tabular}[c]{@{}c@{}}\textbf{26.84}/0.9131/\\ 30.40\end{tabular} \\ \hline
6 &
  \begin{tabular}[c]{@{}c@{}}\textbf{25.83}/\textbf{0.9430}/\\ \textbf{22.44}\end{tabular} &
  \begin{tabular}[c]{@{}c@{}}\textbf{24.43}/\textbf{0.9220}/\\ \textbf{22.77}\end{tabular} &
  \begin{tabular}[c]{@{}c@{}}\textbf{22.81}/\textbf{0.8821}/\\ \textbf{26.75}\end{tabular} &
  \begin{tabular}[c]{@{}c@{}}30.02/\textbf{0.9645}/\\ \textbf{12.44}\end{tabular} &
  \begin{tabular}[c]{@{}c@{}}\textbf{27.16}/\textbf{0.9400}/\\ \textbf{16.04}\end{tabular} &
  \begin{tabular}[c]{@{}c@{}}\textbf{23.94}/\textbf{0.8901}/\\ \textbf{21.95}\end{tabular} &
  \begin{tabular}[c]{@{}c@{}}26.73/\textbf{0.9157}/\\ \textbf{28.92}\end{tabular} \\ \hline
9 &
  \begin{tabular}[c]{@{}c@{}}25.73/0.9421/\\ 24.02\end{tabular} &
  \begin{tabular}[c]{@{}c@{}}24.21/0.9099/\\ 28.57\end{tabular} &
  \begin{tabular}[c]{@{}c@{}}22.32/0.8617/\\ 33.20\end{tabular} &
  \begin{tabular}[c]{@{}c@{}}29.88/0.9588/\\ 16.34\end{tabular} &
  \begin{tabular}[c]{@{}c@{}}26.78/0.9306/\\ 21.65\end{tabular} &
  \begin{tabular}[c]{@{}c@{}}23.44/0.8867/\\ 26.04\end{tabular} &
  \begin{tabular}[c]{@{}c@{}}26.00/0.9044/\\ 31.77\end{tabular} \\ \hline
\end{tabular}
}
\end{table*}

\subsection{Time-steps: }
We perform a detailed study on the number of time steps for our progressive update and dehazing process to decide the value of $t_e$ in Section~III-C3a of the main paper. We use three different number of time steps to analyze the performance of the model with respect to them. We performed experiments with the number of time steps as $3, 6$ and $9$ using the same experimental set up as discussed in Section~IV-B of the main paper. Table~\ref{ablationLSTM} presents the performance evaluation of our PUG-D(R6) models that are trained with the different number of time steps on the NR-haze dataset. The experimental results show that number of time steps equal to $6$ is the best suited for our approach among the three, which we use in our work.
\begin{table*}[t]
\centering
\caption{Performance (of our PUG-D(R6) with only $L_1$ loss) comparison between global update and local updates in our atmospheric light updater network of PUG-D. (Best: Bold highlight)}
\label{tab: localvsglobal}
\resizebox{\textwidth}{!}{%
\begin{tabular}{|c|c|c|c|c|c|c|c|}
\hline
\multirow{3}{*}{\begin{tabular}[c]{@{}c@{}}\\Atmospheric\\ Light \\ Update\end{tabular}} &
  \multicolumn{6}{c|}{Non-cast Hazy Images} &
  Color Cast Hazy Image \\ \cline{2-8} 
 &
  \begin{tabular}[c]{@{}c@{}}Outdoor\\ Low Haze\\(LSOT)\end{tabular}                                                   & \begin{tabular}[c]{@{}c@{}}Outdoor\\ Mid Haze\\(MSOT)\end{tabular} & \begin{tabular}[c]{@{}c@{}}Outdoor\\ High Haze\\(HSOT)\end{tabular}                                                  & \begin{tabular}[c]{@{}c@{}}Indoor\\ Low Haze\\(LSIT)\end{tabular}                                                   & \begin{tabular}[c]{@{}c@{}}Indoor\\ Mid Haze\\(MSIT)\end{tabular}                                                   & \begin{tabular}[c]{@{}c@{}}Indoor\\ High Haze\\(HSIT)\end{tabular} & \begin{tabular}[c]{@{}c@{}}Color Cast\\ Random Haze\\(SCHT)\end{tabular}                                                        \\ \cline{2-8}  
 &
  \begin{tabular}[c]{@{}c@{}}PSNR/SSIM/\\ CIEDE2000\end{tabular} &
  \begin{tabular}[c]{@{}c@{}}PSNR/SSIM/\\ CIEDE2000\end{tabular} &
  \begin{tabular}[c]{@{}c@{}}PSNR/SSIM/\\ CIEDE2000\end{tabular} &
  \begin{tabular}[c]{@{}c@{}}PSNR/SSIM/\\ CIEDE2000\end{tabular} &
  \begin{tabular}[c]{@{}c@{}}PSNR/SSIM/\\ CIEDE2000\end{tabular} &
  \begin{tabular}[c]{@{}c@{}}PSNR/SSIM/\\ CIEDE2000\end{tabular} &
  \begin{tabular}[c]{@{}c@{}}PSNR/SSIM/\\ CIEDE2000\end{tabular} \\ \hline
Local &
  \begin{tabular}[c]{@{}c@{}}23.18/0.8875/\\ 35.67\end{tabular} &
  \begin{tabular}[c]{@{}c@{}}20.12/0.8037/\\ 44.41\end{tabular} &
  \begin{tabular}[c]{@{}c@{}}17.77/0.7482/\\ 48.61\end{tabular} &
  \begin{tabular}[c]{@{}c@{}}\textbf{28.35}/\textbf{0.9475}/\\ \textbf{20.23}\end{tabular} &
  \begin{tabular}[c]{@{}c@{}}22.98/0.8550/\\ 32.05\end{tabular} &
  \begin{tabular}[c]{@{}c@{}}18.74/0.7423/\\ 43.84\end{tabular} &
  \begin{tabular}[c]{@{}c@{}}24.22/0.8451/\\ 43.70\end{tabular} \\ \hline
Global &
  \begin{tabular}[c]{@{}c@{}}\textbf{23.85}/\textbf{0.9096}/\\ \textbf{33.26}\end{tabular} &
  \begin{tabular}[c]{@{}c@{}}\textbf{22.57}/\textbf{0.8883}/\\ \textbf{36.67}\end{tabular} &
  \begin{tabular}[c]{@{}c@{}}\textbf{21.61}/\textbf{0.8431}/ \\ \textbf{39.61}\end{tabular} &
  \begin{tabular}[c]{@{}c@{}}28.24/0.9405/\\ 22.04\end{tabular} &
  \begin{tabular}[c]{@{}c@{}}\textbf{25.16}/\textbf{0.9070}/\\ \textbf{27.38}\end{tabular} &
  \begin{tabular}[c]{@{}c@{}}\textbf{21.90}/\textbf{0.8362}/\\ \textbf{35.42}\end{tabular} &
  \begin{tabular}[c]{@{}c@{}}\textbf{25.08}/\textbf{0.8818}/\\ \textbf{38.19}\end{tabular} \\ \hline
\end{tabular}
}
\end{table*}

\subsection{Global vs. Local Atmospheric Light Updater Network}
\label{LGupdate}
As discussed in Section~III-C3b of the main paper, we globally update the atmospheric light in our dehazing framework. We use average pooling at the end of the atmospheric light updater network to get a single overall update in each color channel. Average pooling is considered to ensure that all the pixel values contribute to the global update. We adopt the global update instead of a local update based on our experimental findings. Using the NR-haze dataset, Table~\ref{tab: localvsglobal} shows the experimental results of our PUG-D(R6) without the average pooling where the update happens locally in comparison to when the pooling is used resulting in global update. Similar to Section V-A and Table IV in the main paper, only mean absolute error ($L_1$) loss function, and only the first and second training stages are used for our PUG-D(R6) in this study. It is clearly evident that in most cases there is performance degradation when the local update is considered. This may indicate that the atmospheric light in our approach represents the haze illumination as a global quantity of an image.

\subsection{Pooling in Atmospheric Light Estimation Network}
\label{sec: pooling_exp}
In Section~III-B2 of the main paper, we have discussed the intuition behind our max-pooling layers, including the global max-pooling, in our atmospheric light estimation network. To validate this, we experiment with two different pooling mechanisms, max- and average pooling, in the atmospheric light estimation network. In Table~\ref{ablationPooling}, we present the quantitative evaluation of our PUG-D(R6) on the NR-haze dataset. We show the mean squared error between the actual and estimated atmospheric lights of the two trained networks with max-pooling and average-pooling. We experimentally find that max-pooling performs far better with a substantial margin. Moreover, as described in Section~III-B2 of the main paper, max-pooling fits well with the popular idea of atmospheric light estimation using DCP.
\begin{table}
\centering
\caption{MSE comparison of max-pooling and average pooling for the atmospheric light estimation network in our PUG-D. All the values are to multiplied by $10^{-4}$. \newline (Best: Bold highlight)}
\label{ablationPooling}
\begin{tabular}{|c|c|c|c|}
\hline
\multirow{2}{*}{Image Type}                                                      & \multirow{2}{*}{Haze Density} & \multicolumn{2}{c|}{Pool Type} \\ \cline{3-4} 
 &                   & Max-Pool      & Average Pool  \\ \hline
\multirow{6}{*}{\begin{tabular}[c]{@{}c@{}}Non-cast \\ Hazy Images\end{tabular}} & Outdoor Low Haze              & \textbf{9.7}       & 26.7      \\ \cline{2-4} 
 & Outdoor Mid Haze  & \textbf{7.8} & 24.7          \\ \cline{2-4} 
 & Outdoor High Haze & \textbf{6.8} & 11.4          \\ \cline{2-4} 
 & Indoor Low Haze   & 25.2         & \textbf{22.2} \\ \cline{2-4} 
 & Indoor Mid Haze   & \textbf{9.1} & 22.4          \\ \cline{2-4} 
 & Indoor High Haze  & \textbf{5.2} & 17.8          \\ \hline
\begin{tabular}[c]{@{}c@{}}Color Cast \\ Hazy Images\end{tabular}                & Random Haze                   & \textbf{0.97}      & 1.69      \\ \hline
\end{tabular}
\end{table}

\begin{table*}
\centering
\caption{The Model Complexity of PUG-D(R6) along with that of the state-of-the-art deep learning models.}
\label{tab: time}
\resizebox{\textwidth}{!}{%
\begin{tabular}{|l|c|c|c|c|c|}
\hline
\multicolumn{1}{|c|}{Model Name} &
  \begin{tabular}[c]{@{}c@{}}Parameters\\ (Million)\end{tabular} &
  \begin{tabular}[c]{@{}c@{}}Memory\\ Consumed (MB)\end{tabular} &
  \begin{tabular}[c]{@{}c@{}}Images per\\ Second\end{tabular} &
  \begin{tabular}[c]{@{}c@{}}O-Haze/ I-Haze\\ (PSNR)\end{tabular} &
  \begin{tabular}[c]{@{}c@{}}BeDDE\\ (VI/ RI)\end{tabular} \\ \hline
EPDN         & 17.38 & 977  & 127.05 & 17.38/ 15.80 & 0.8956/ 0.9640 \\ \hline
GridDNet         & 0.958 & 963  & 75.82  & 13.54/ 12.24 & 0.8909/ 0.9682 \\ \hline
MSBDN        & 31.35 & 1051 & 56.30  & 16.83/ 16.57 & 0.7688/ 0.9039 \\ \hline
RefineDNet   & 65.80 & 3361 & 30.30  & 17.12/ 15.79 & 0.9073/ 0.9707 \\ \hline
D4           & 22.90 & 1715 & 40.12  & 14.45/ 13.60 & 0.8128/ 0.9707 \\ \hline
PUG-D(R6)    & 14.48 & 1213 & 32.27  & 19.39/ 16.21 & 0.9065/ 0.9711 \\ \hline
\end{tabular}%
}
\end{table*}

\begin{figure*}
    \centering
    \includegraphics[width=0.6\linewidth]{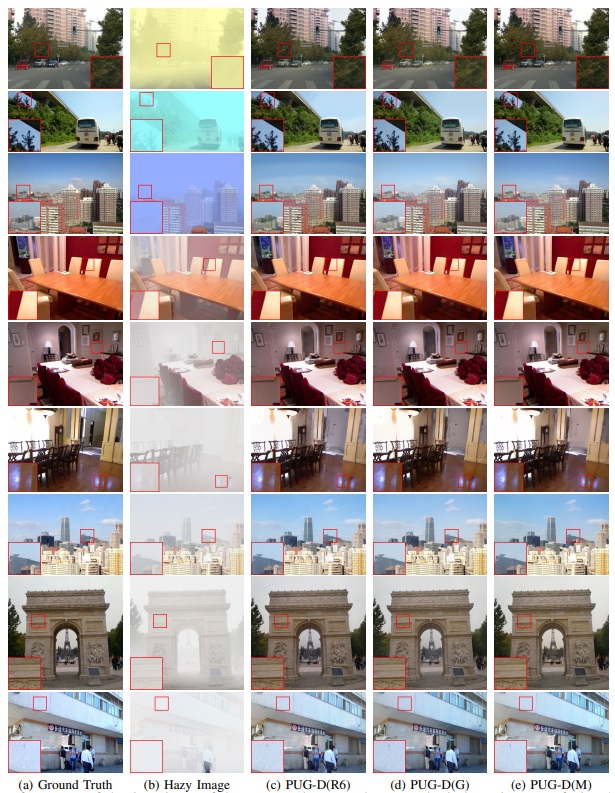}
    \caption{Dehazing performance of the 3 variants of the proposed approach on synthetic hazy images of Fig. 5 in the main paper.}
    \label{fig: syncomp}
\end{figure*}

\begin{figure*}
    \centering
    \includegraphics[width=0.6\linewidth]{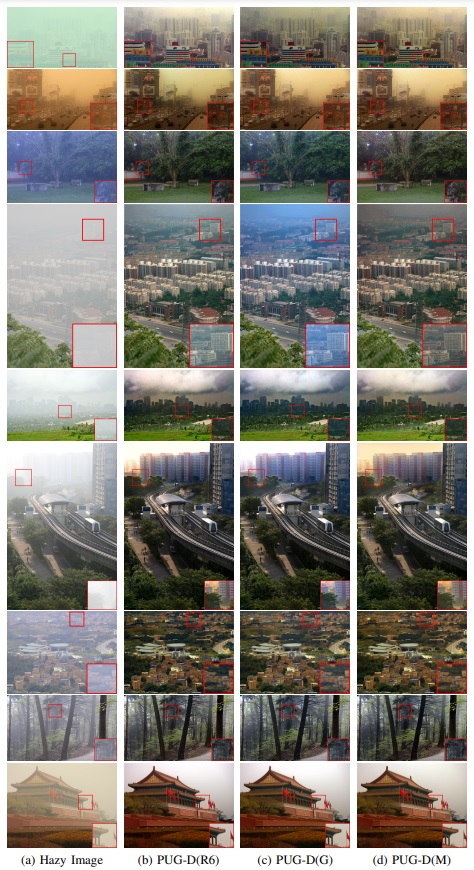}
    \caption{Dehazing performance of the 3 variants of the proposed approach on real-life hazy images of Fig. 6 in the main paper.}
    \label{fig: comp}
\end{figure*}

\section{Additional Analysis of Our Approach}

\subsection{Model Complexity}
\label{sec: time}
In this section, we present the complexity of our approach in terms of model parameters, memory consumed, and computation speed. The complexity of our proposed framework PUG-D(R6) is compared to the best performing state-of-the-art deep network based dehazing approaches, EPDN, D4, RefineDNet, GridDNet and MSBDN, in Table~\ref{tab: time}. It further lists the dehazing performance of the above pre-trained models on the datasets of real hazy images, O-Haze, I-Haze and BeDDE, which are discussed in Sections~IV-A3 and~IV-A4 of the main paper. While PSNR is shown for the O-Haze and I-Haze datasets, VI and RI are shown for BeDDE datasets. The models have been implemented on an nVIDIA 2080Ti GPU. Table~\ref{tab: time} shows that although the complexity of our PUG-D(R6) is comparable to other models, the performance of our PUG-D(R6) is in general much better than others. 

\begin{figure*}
    \centering
    \includegraphics[width=0.8\linewidth]{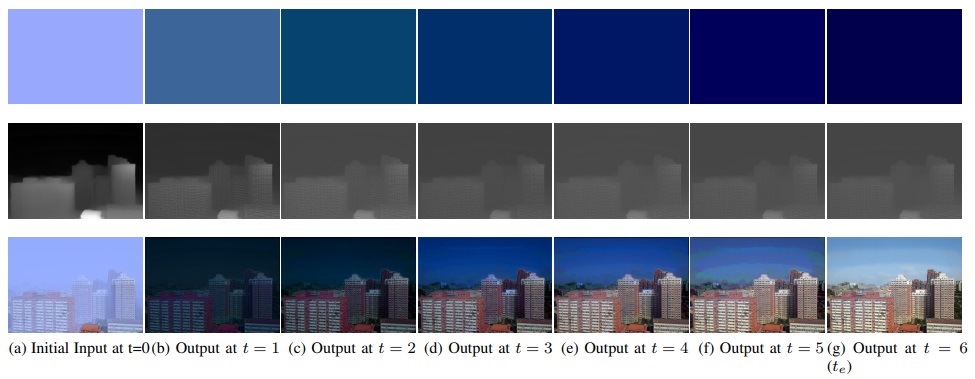}
    \caption{Intermediate outputs of our PUG-D(R6). Initially estimated transmission map and atmospheric light along with the hazy image are shown at $t=0$. The updated map and the dehazed image in each time steps are shown. First row shows atmospheric light and its updates. Second row shows transmission map and its updated maps. Third row shows dehazed outputs in all the time steps. (zoom for the best view)}
    \label{viewIntermediate}
\end{figure*}

\begin{figure*}
    \centering
    \includegraphics[width=0.8\linewidth]{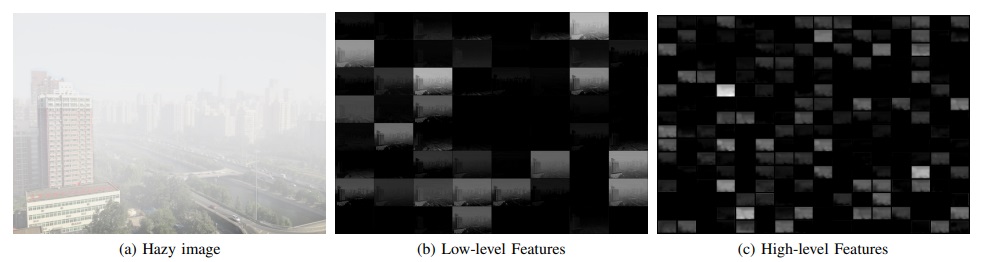}
    \caption{Low and high-level features extracted from a hazy image using trained atmospheric light estimation model.}
    \label{fig: atm_feat}
\end{figure*}

\subsection{Subjective Evaluation of the Proposed Approach's Variants}
In  Figs.~\ref{fig: syncomp} and~\ref{fig: comp}, we show the dehazed images obtained using the variants of the proposed approach on the synthetic and real hazy images of Figs. 5 and 6 in the main paper, respectively. As can be seen from the results on the synthetic hazy images, there is hardly any visual difference in their dehazing performance. The dehazing performance of the variants on the real hazy images varies slightly in a few cases, where the PUG-D(R6) variant with the simplest feature extraction architecture seems to perform a little better qualitatively by almost completely removing color cast and by avoiding color fading.  Fig.~\ref{figsupp/RealResults} shows the dehazing results of the PUG-D(R6) on quite a few additional real hazy images, where we see that the  dehazed images are visibility-enhanced versions of the corresponding hazy images with no noticeable artifacts.

\begin{figure*}
    \centering
    \includegraphics[width=0.8\linewidth]{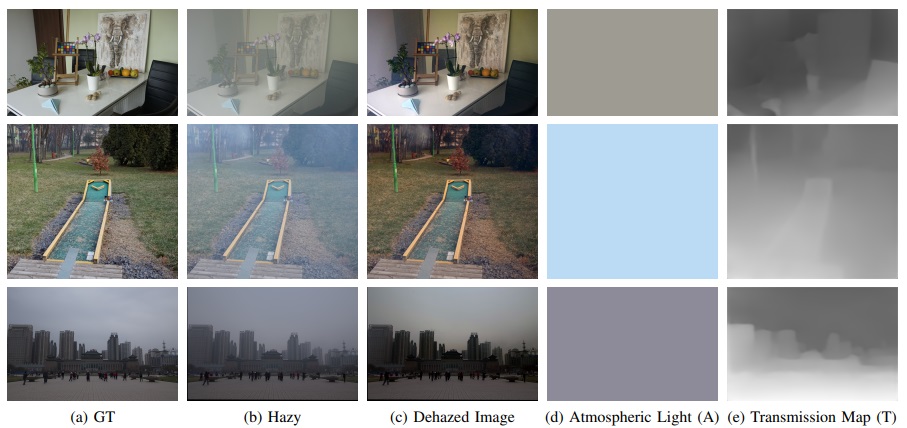}
    \caption{Atmospheric Light ($A$) and Transmission Map ($T$) of Hazy Input Image ($I$) and Corresponding Dehazed Image ($I'$). First, second and third row hazy images are from I-Haze, O-Haze and BeDDE datasets.}
    \label{fig: atm_trans_map}
\end{figure*} 

\subsection{Visualisation of Estimated Prior Maps}
Fig.~\ref{fig: atm_trans_map} presents three hazy images from I-Haze (top row), O-Haze (middle row), and BeDDE (last row) datasets and their corresponding atmospheric lights and transmission maps. As evident from the figure, the transmission map is a structure-aware and locally smooth map that quantifies the transmission of scene radiance from each pixel. The atmospheric light is an RGB value that indicates the light intensity of the source.  Achromatic atmospheric light produces a non-cast hazy image. The first and last row images are non-cast; therefore, atmospheric lights are also achromatic. On the other hand,    the middle-row image is color-cast hazy image with a chromatic  atmospheric light.    

\subsection{Intermediate Outputs of Our Dehazing Approach}
Fig.~\ref{viewIntermediate} shows the progressive updates of the atmospheric light and transmission map, and the generated dehazed output at each time step considering our PUG-D(R6). In our dehazing network, after each time step, the atmospheric light updater updates the initial estimated atmospheric light, and we observe that the updated color value becomes purer (higher saturation) with the increase in iteration.
On the other hand, the transmission map estimator network initially generates a smoothed map of the image structure. After that, in the dehazing module, the transmission map updater updates that map, upon which image structure details appear in the map, as seen by zooming into the relevant maps in Fig.~\ref{viewIntermediate}. These structures details are then diminished a little with the increase in the time steps, possibly striking a fine balance between structure preservation and noise reduction as required to achieve optimal dehazing. It is also evident from the figure that the dehazed output is the best at $t=6$.

\subsection{Feature Extraction in Our Atmospheric Light Estimation Model}
Fig.~\ref{fig: atm_feat} shows the extracted features by our trained atmospheric light estimation model from a hazy image in the initial layer and final layer before the global max-pooling. In the initial layer, instead of learning standard low-level kernels to obtain features like edges and orientations, we observe that the learned low-level kernels operate on input images to give outputs with different intensity shifts. Progressing to the final layer, the high-level feature contents become very smooth, possibly converging to provide a single value per color channel as the estimated atmospheric light after the global max-pooling.

\section{Distinction from Existing Methods}
Use of atmospheric light and transmission map as priors to guide dehazing~\cite{liu2019learning, guo2019dense, zhang2018densely, cai2016dehazenet} and iterative mechanisms~\cite{liu2019learning, hua2020iterative, du2018recursive} to perform dehazing have been considered quite a few times earlier. Here, we emphasize the novelty in the concepts of our progressive mechanism and of our use of the priors, in light of such existing methods.

Atmospheric light and transmission map have been used in literature to perform dehazing in conjunction with Koschmieder's atmospheric light scattering model. Once these priors are estimated using deep networks, they are used in Koschmieder's atmospheric light scattering model to achieve dehazing. We too estimate the priors using deep networks. However, we do not use any predefined light scattering model, but consider a dehazing module that progressively updates the priors using deep networks and  performs the dehazing using the updated priors in another deep network. Hence, we avoid any model-oriented restriction to the image reconstruction. We experimentally show  that the use of deep networks for dehazing rather than the Koschmieder's model after the priors are estimated improves performance.

In literature, iterative mechanisms have been used on estimator networks that compute the priors for use in Koschmieder's model based dehazing. Iterative mechanisms have been used on end-to-end deep dehazing networks as well. However, given the novel framework for dehazing, our  mechanism used in it is fundamentally different. The progressive mechanism is applied on the dehazing module which comprises of the two prior updater networks and the dehazing network.

\begin{figure*}
\centering
\captionsetup[subfloat]{labelformat=empty}
\subfloat[Hazy image]
{
    \includegraphics[width=6.5cm]{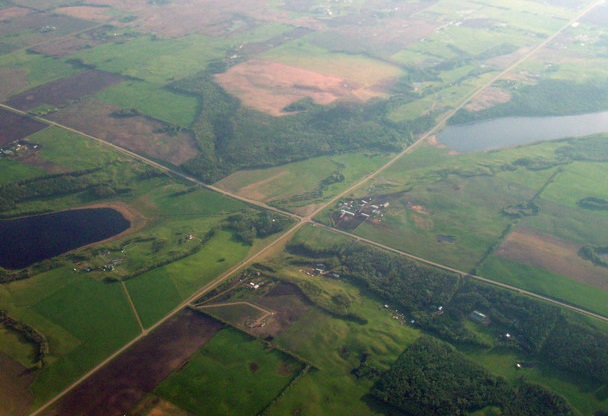}
}\hspace{-0.2cm}
\subfloat[Dehazed Output (Proposed)]
{
    \includegraphics[width=6.5cm]{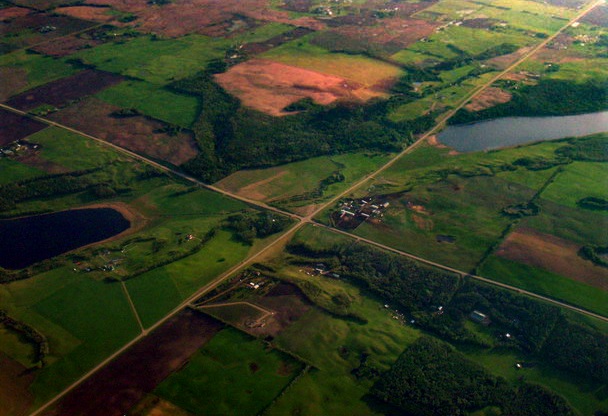}
}\hspace{-0.2cm}
\captionsetup{labelformat=empty}
\end{figure*}

\begin{figure*}
\centering \captionsetup[subfloat]{labelformat=empty}
\subfloat[Hazy image]
{
    \includegraphics[width=6.5cm]{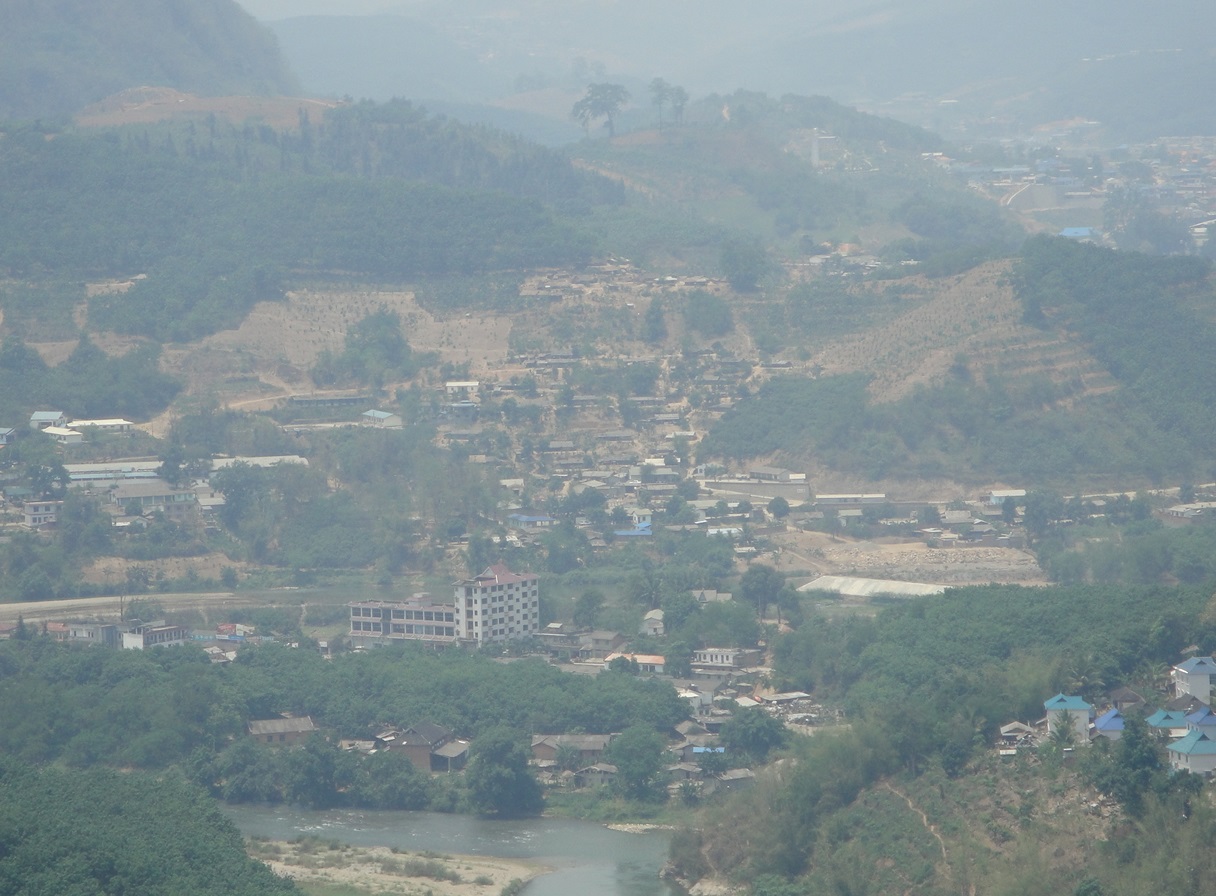}
}\hspace{-0.2cm}
\subfloat[Dehazed Output (Proposed)]
{
    \includegraphics[width=6.5cm]{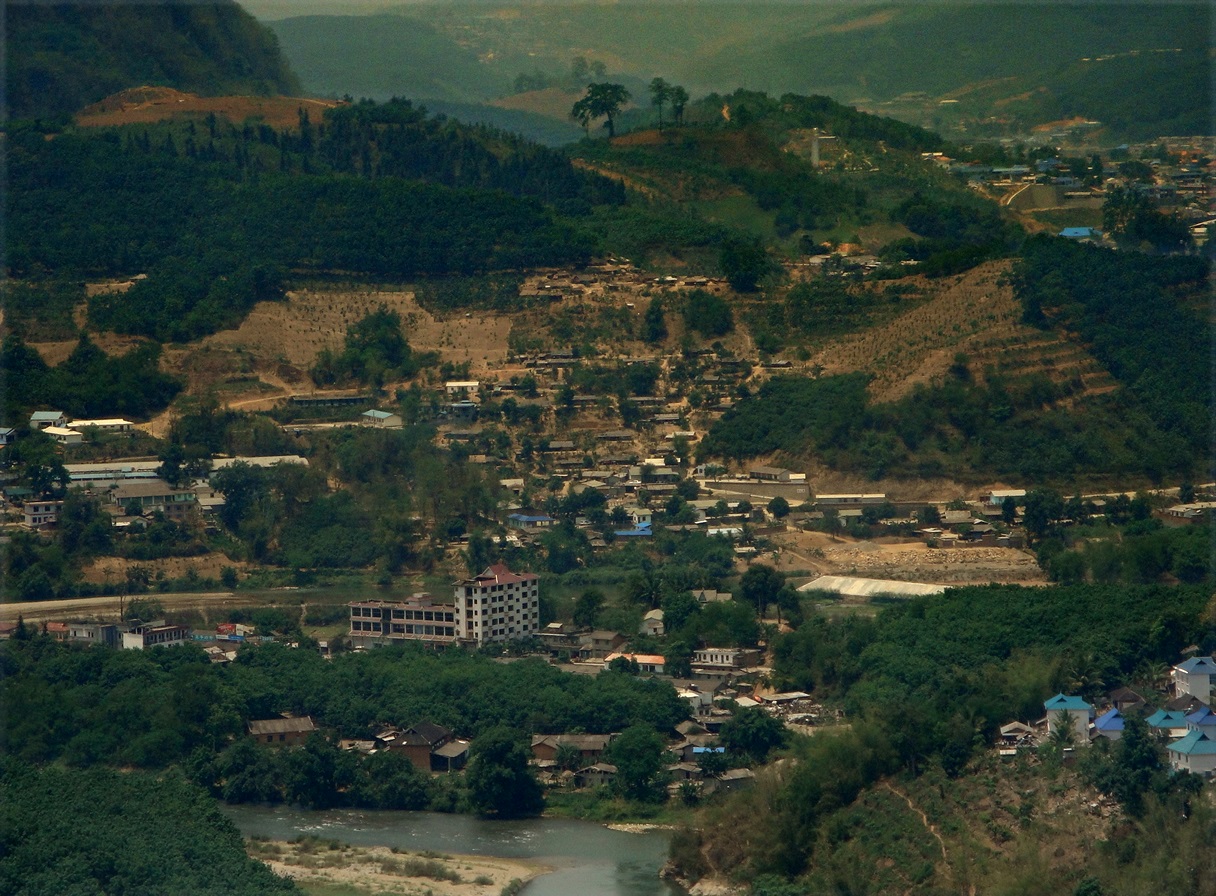}
}\hspace{-0.2cm}
\captionsetup{labelformat=empty}
\end{figure*}

\begin{figure*}
\centering
\captionsetup[subfloat]{labelformat=empty}
\subfloat[Hazy image]
{
    \includegraphics[width=6.5cm]{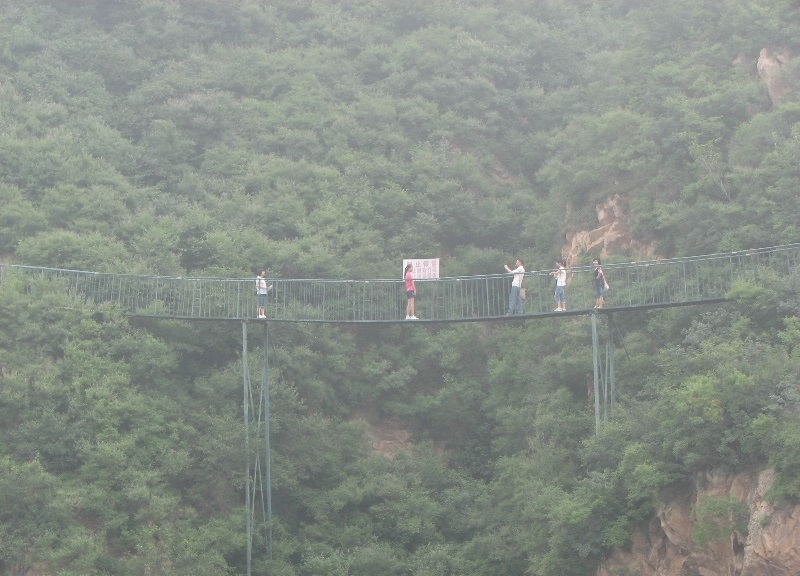}
}\hspace{-0.2cm}
\subfloat[Dehazed Output (Proposed)]
{
    \includegraphics[width=6.5cm]{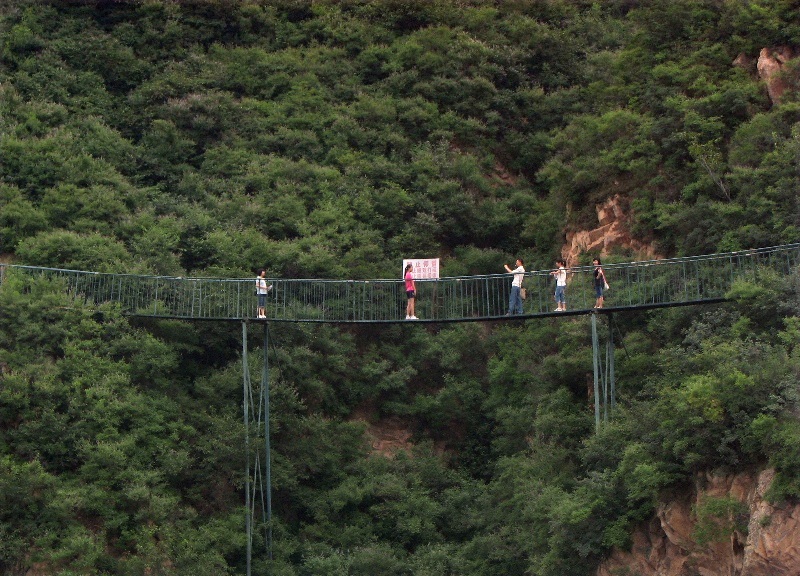}
}\hspace{-0.2cm}
\captionsetup{labelformat=empty}
\end{figure*}

\begin{figure*}
\centering \captionsetup[subfloat]{labelformat=empty}
\subfloat[Hazy image]
{
    \includegraphics[width=6.5cm]{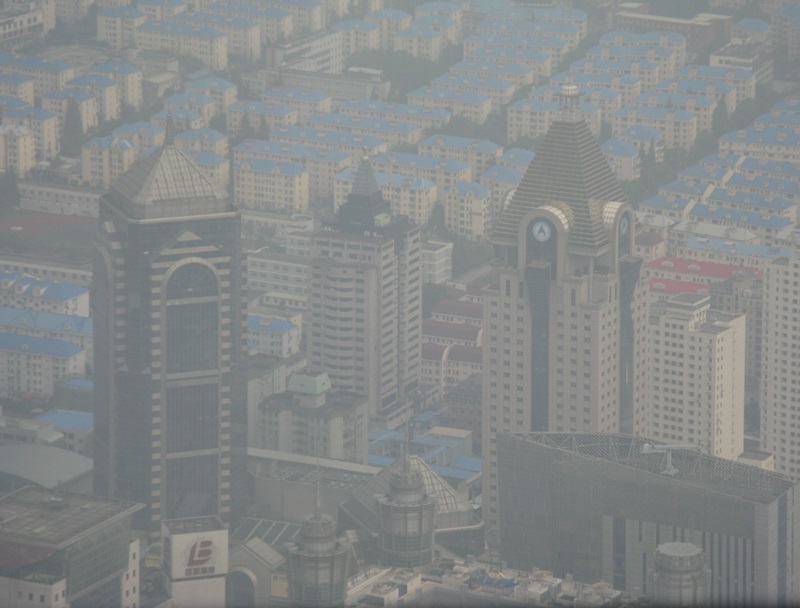}
}\hspace{-0.2cm}
\subfloat[Dehazed Output (Proposed)]
{
    \includegraphics[width=6.5cm]{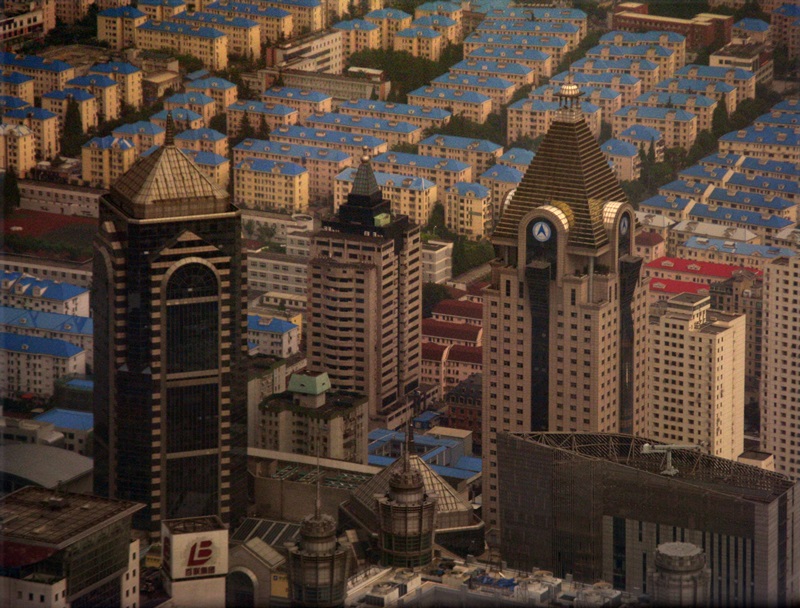}
}\hspace{-0.2cm}
\captionsetup{labelformat=empty}
\end{figure*}

\begin{figure*}
\centering \captionsetup[subfloat]{labelformat=empty}
\subfloat[Hazy image]
{
    \includegraphics[width=6.5cm]{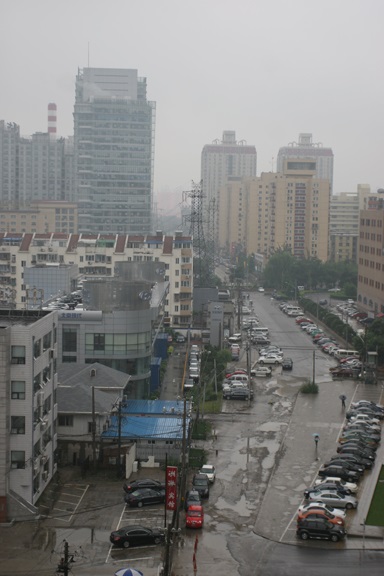}
}\hspace{-0.2cm}
\subfloat[Dehazed Output (Proposed)]
{
    \includegraphics[width=6.5cm]{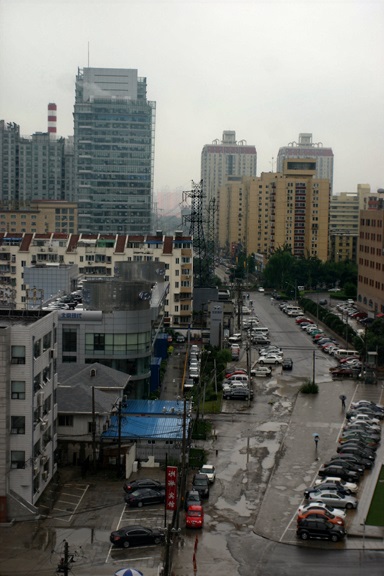}
}\hspace{-0.2cm}
\captionsetup{labelformat=empty}
\end{figure*}

\begin{figure*}
\centering \captionsetup[subfloat]{labelformat=empty}
\subfloat[Hazy image]
{
    \includegraphics[width=6.5cm]{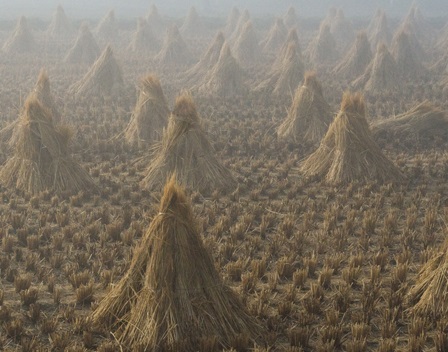}
}\hspace{-0.2cm}
\subfloat[Dehazed Output (Proposed)]
{
    \includegraphics[width=6.5cm]{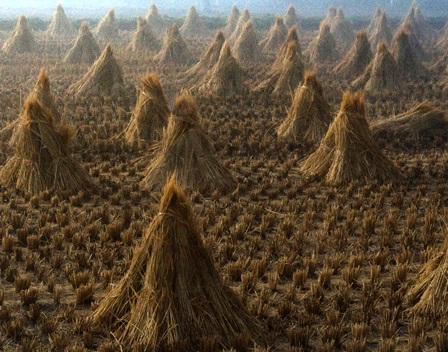}
}\hspace{-0.2cm}
\captionsetup{labelformat=empty}
\end{figure*}

\begin{figure*}
\centering \captionsetup[subfloat]{labelformat=empty}
\subfloat[Hazy image]
{
    \includegraphics[width=6.5cm]{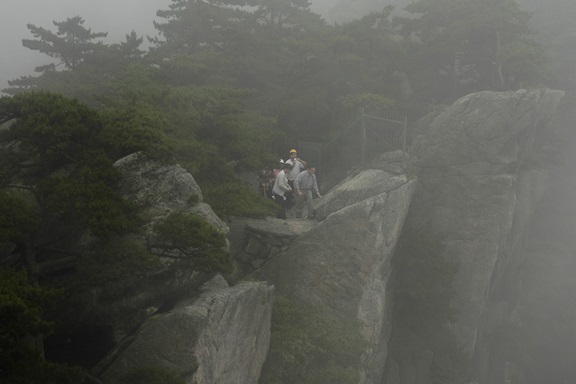}
}\hspace{-0.2cm}
\subfloat[Dehazed Output (Proposed)(Proposed)]
{
    \includegraphics[width=6.5cm]{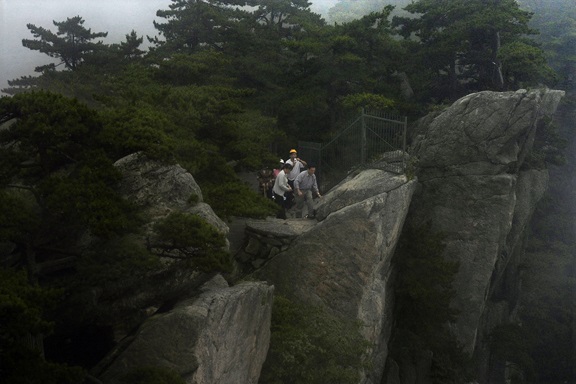}
}\hspace{-0.2cm}
\captionsetup{labelformat=empty}
\end{figure*}

\begin{figure*}
\centering \captionsetup[subfloat]{labelformat=empty}
\subfloat[Hazy image]
{
    \includegraphics[width=6.5cm]{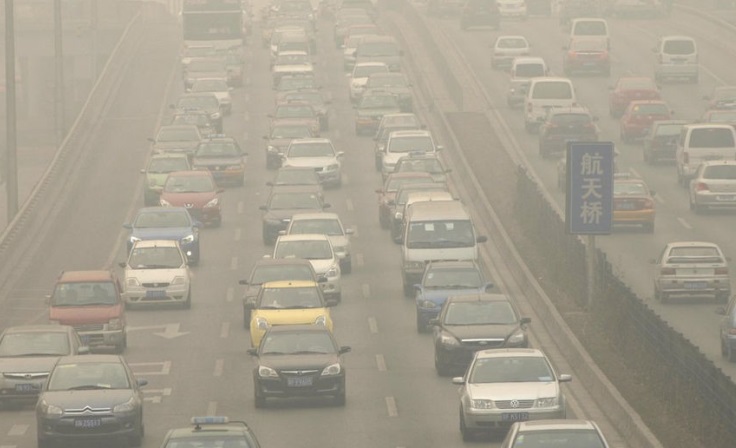}
}\hspace{-0.2cm}
\subfloat[Dehazed Output (Proposed)]
{
    \includegraphics[width=6.5cm]{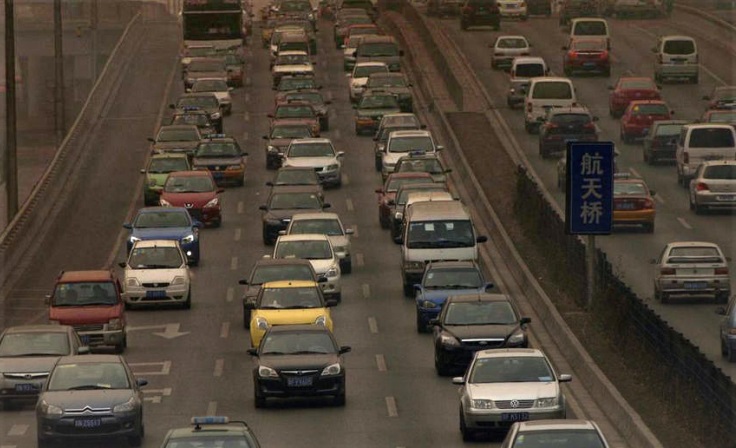}
}\hspace{-0.2cm}
\captionsetup{labelformat=empty}
\end{figure*}

\begin{figure*}
\centering \captionsetup[subfloat]{labelformat=empty}
\subfloat[Hazy image]
{
    \includegraphics[width=6.5cm]{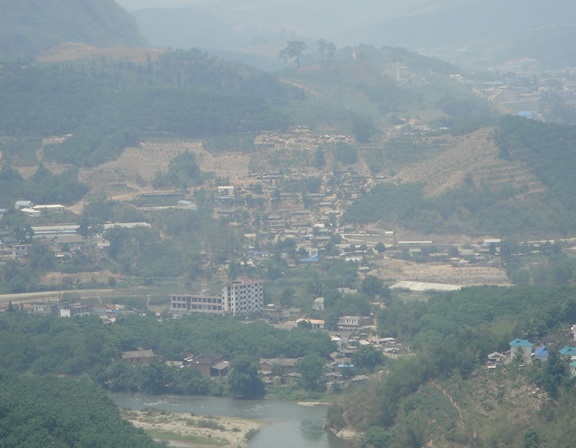}
}\hspace{-0.2cm}
\subfloat[Dehazed Output (Proposed)]
{
    \includegraphics[width=6.5cm]{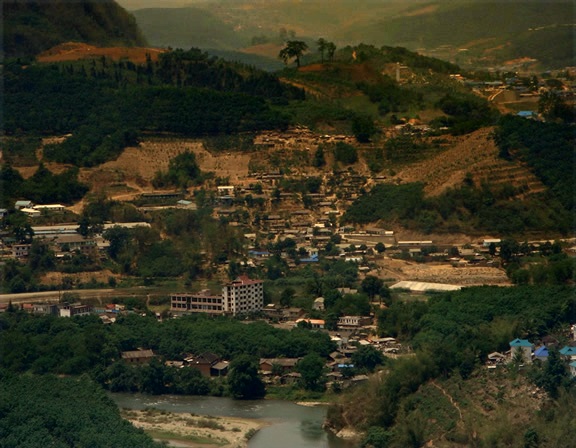}
}\hspace{-0.2cm}
\captionsetup{labelformat=empty}
\end{figure*}

\begin{figure*}
\centering \captionsetup[subfloat]{labelformat=empty}
\subfloat[Hazy image]
{
    \includegraphics[width=6.5cm]{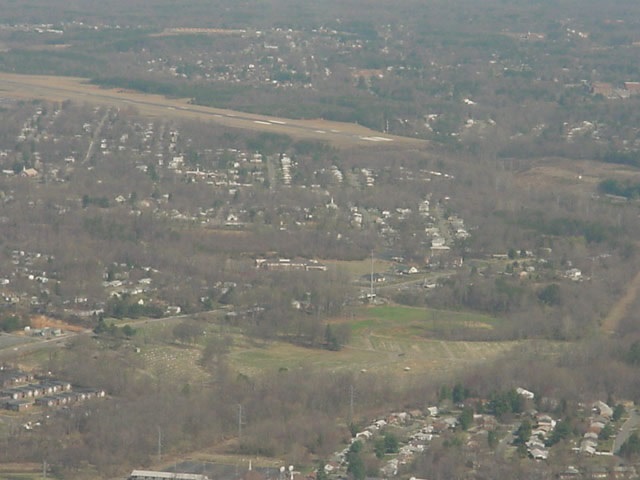}
}\hspace{-0.2cm}
\subfloat[Dehazed Output (Proposed)]
{
    \includegraphics[width=6.5cm]{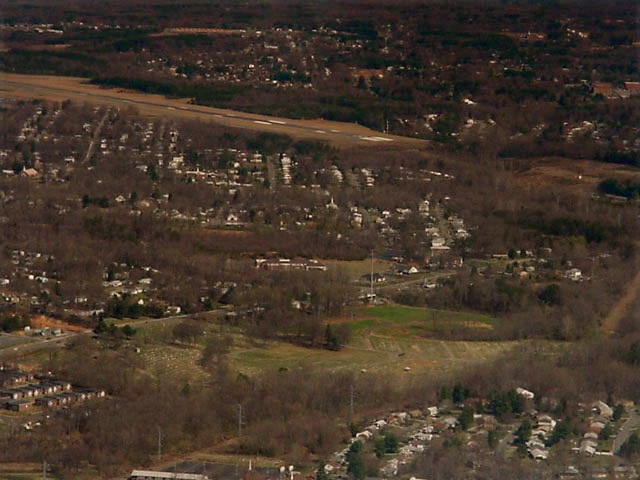}
}\hspace{-0.2cm}
\captionsetup{labelformat=empty}
\end{figure*}

\begin{figure*}
\centering \captionsetup[subfloat]{labelformat=empty}
\subfloat[Hazy image]
{
    \includegraphics[width=6.5cm]{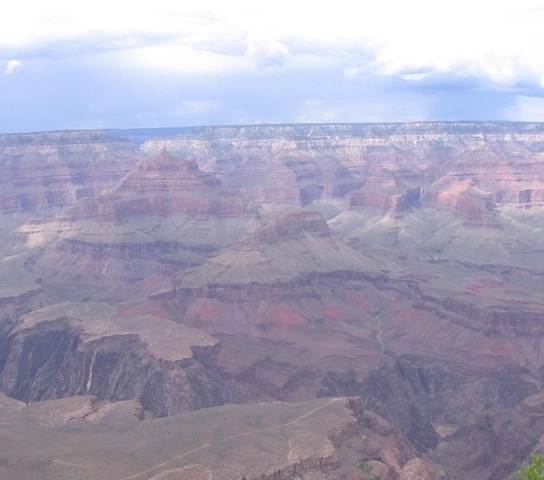}
}\hspace{-0.2cm}
\subfloat[Dehazed Output (Proposed)]
{
    \includegraphics[width=6.5cm]{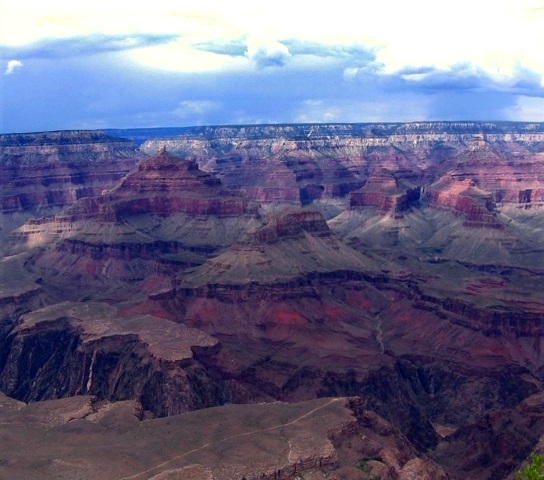}
}\hspace{-0.2cm}
\captionsetup{labelformat=empty}
\end{figure*}

\begin{figure*}
\centering \captionsetup[subfloat]{labelformat=empty}
\subfloat[Hazy image]
{
    \includegraphics[width=6.5cm]{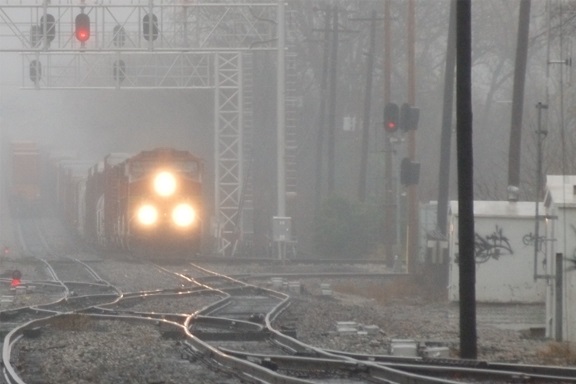}
}\hspace{-0.2cm}
\subfloat[Dehazed Output (Proposed)]
{
    \includegraphics[width=6.5cm]{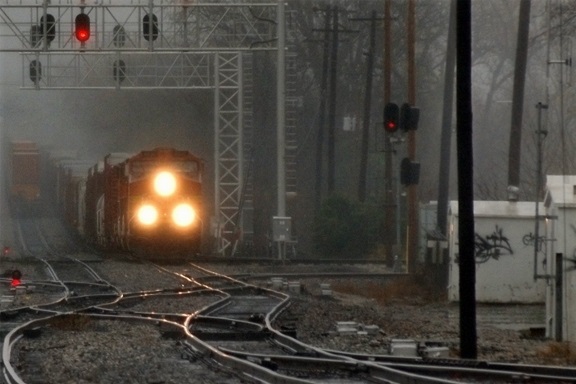}
}\hspace{-0.2cm}
\captionsetup{labelformat=empty}
\end{figure*}

\begin{figure*}
\centering \captionsetup[subfloat]{labelformat=empty}
\subfloat[Hazy image]
{
    \includegraphics[width=6.5cm]{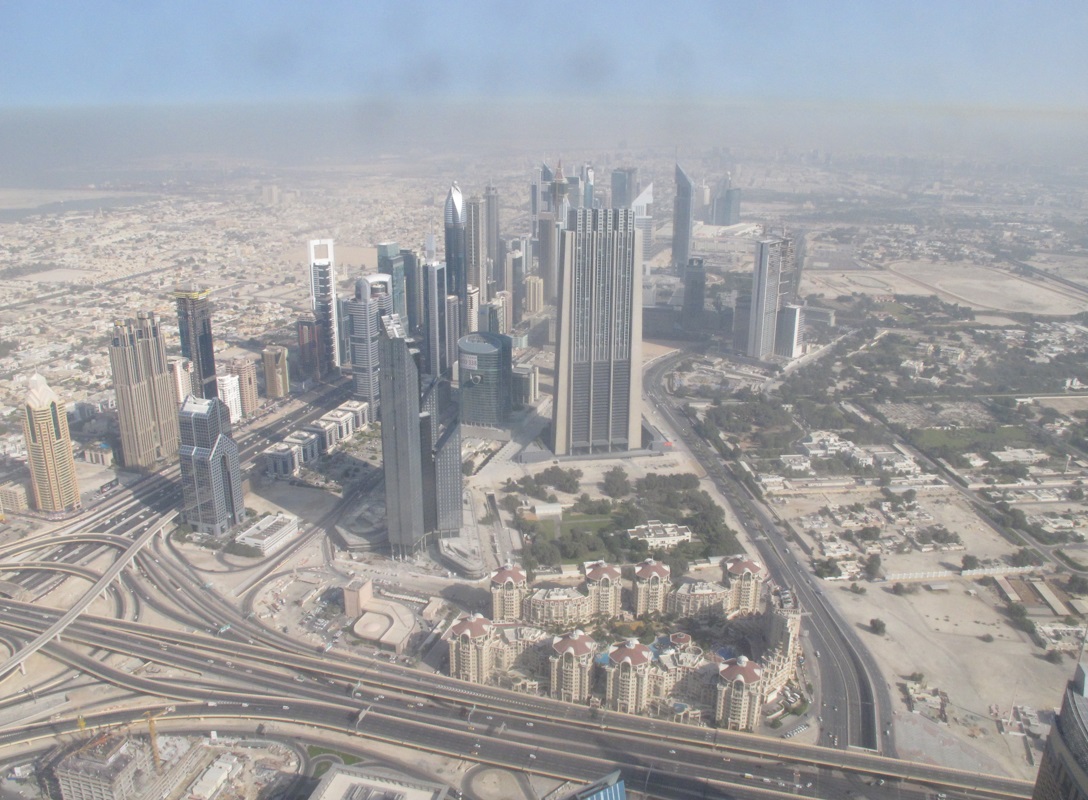}
}\hspace{-0.2cm}
\subfloat[Dehazed Output (Proposed)]
{
    \includegraphics[width=6.5cm]{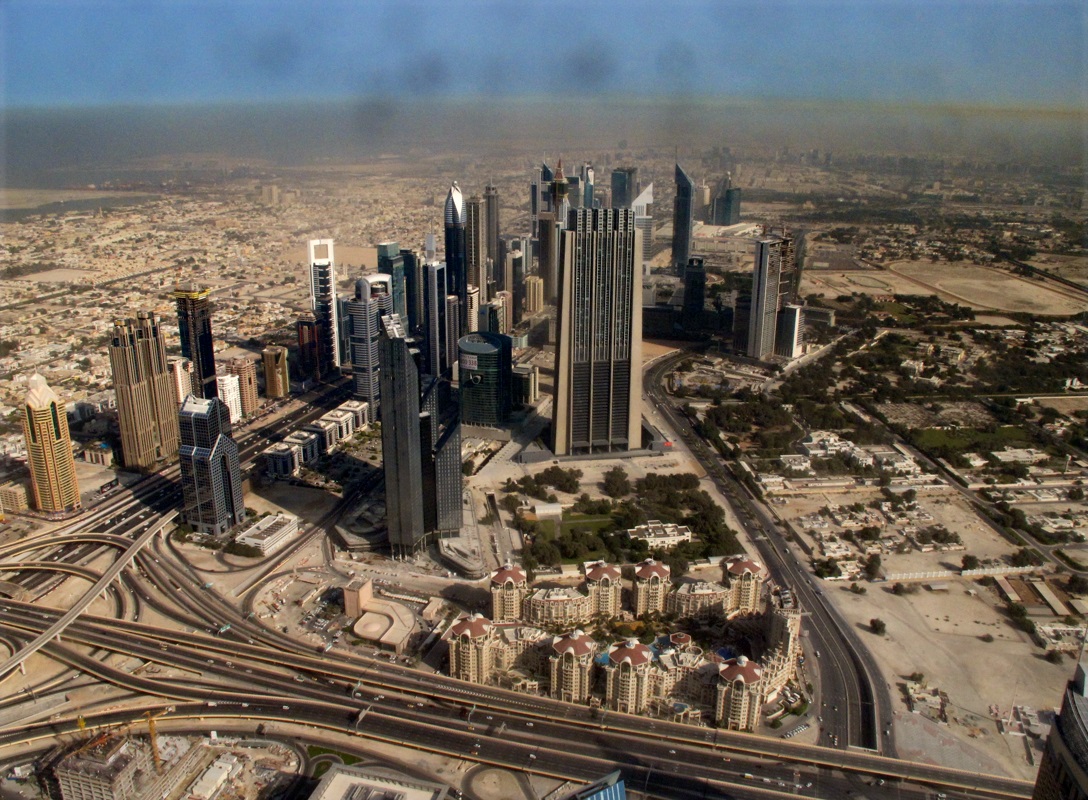}
}\hspace{-0.2cm}
\captionsetup{labelformat=empty}
\end{figure*}

\begin{figure*}
\centering \captionsetup[subfloat]{labelformat=empty}
\subfloat[Hazy image]
{
    \includegraphics[width=6.5cm]{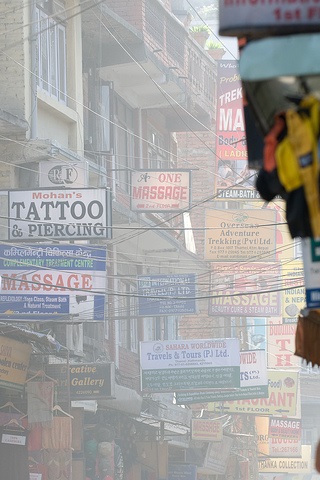}
}\hspace{-0.2cm}
\subfloat[Dehazed Output (Proposed)]
{
    \includegraphics[width=6.5cm]{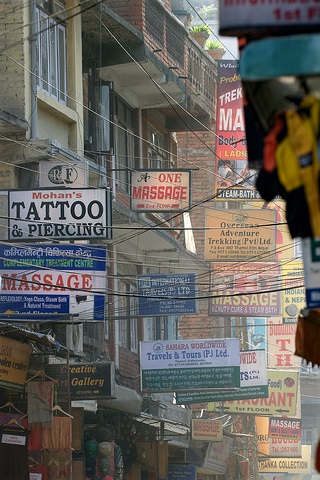}
}\hspace{-0.2cm}
\captionsetup{labelformat=empty}
\end{figure*}

\begin{figure*}
\centering \captionsetup[subfloat]{labelformat=empty}
\subfloat[Hazy image]
{
    \includegraphics[width=6.5cm]{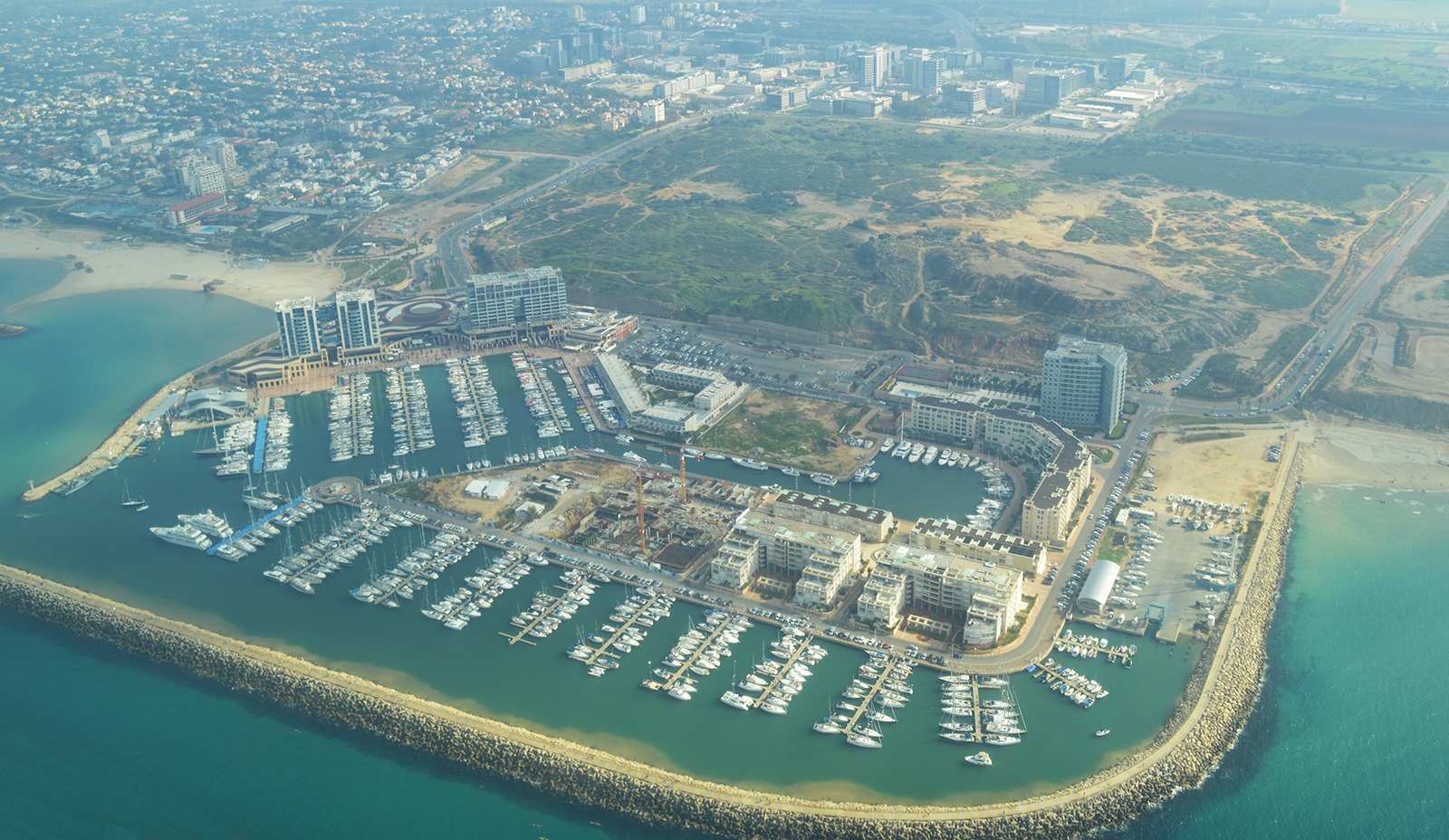}
}\hspace{-0.2cm}
\subfloat[Dehazed Output (Proposed)]
{
    \includegraphics[width=6.5cm]{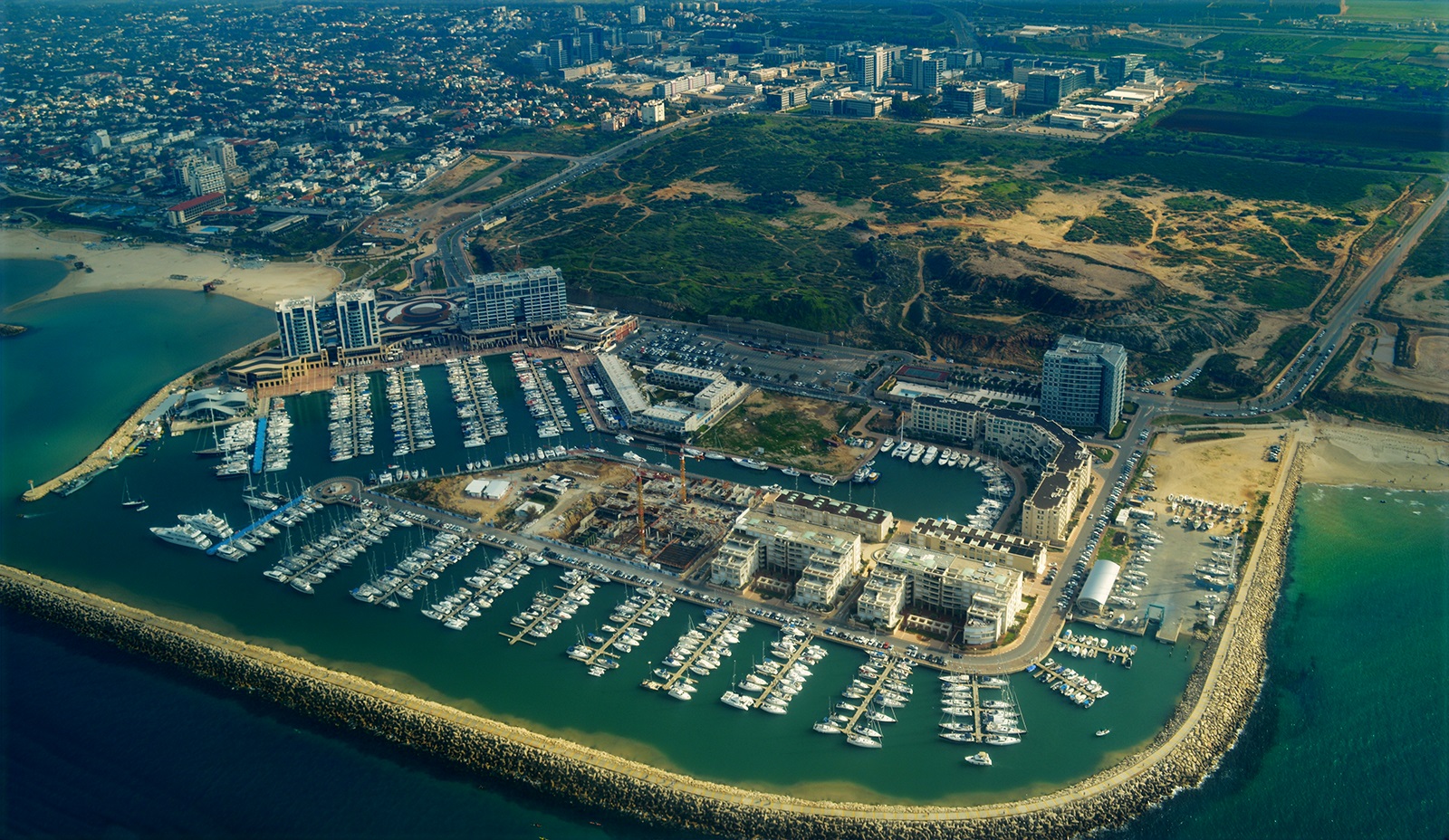}
}\hspace{-0.2cm}
\captionsetup{labelformat=empty}
\end{figure*}

\begin{figure*}
\centering \captionsetup[subfloat]{labelformat=empty}
\subfloat[Hazy image]
{
    \includegraphics[width=6.5cm]{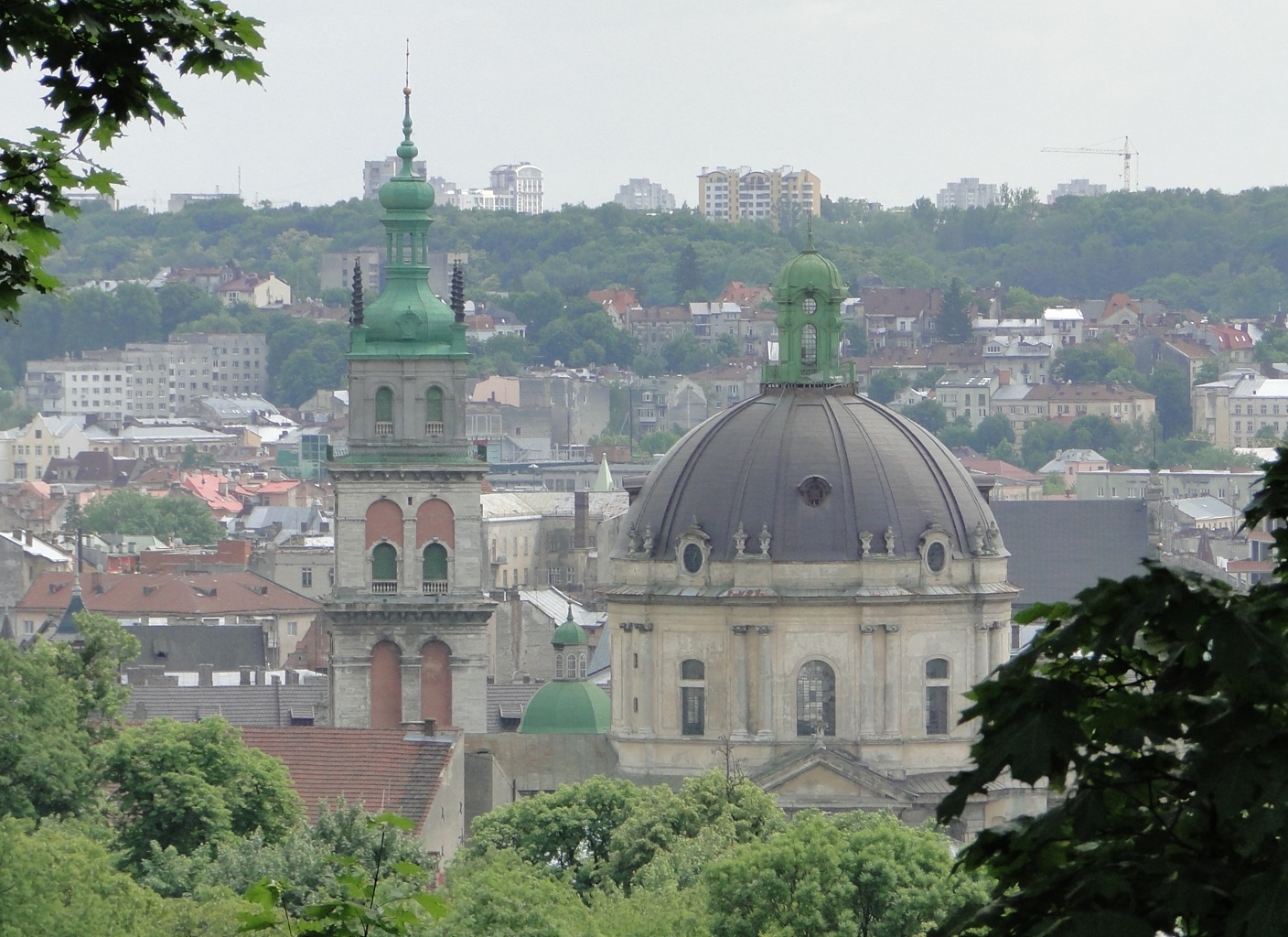}
}\hspace{-0.2cm}
\subfloat[Dehazed Output (Proposed)]
{
    \includegraphics[width=6.5cm]{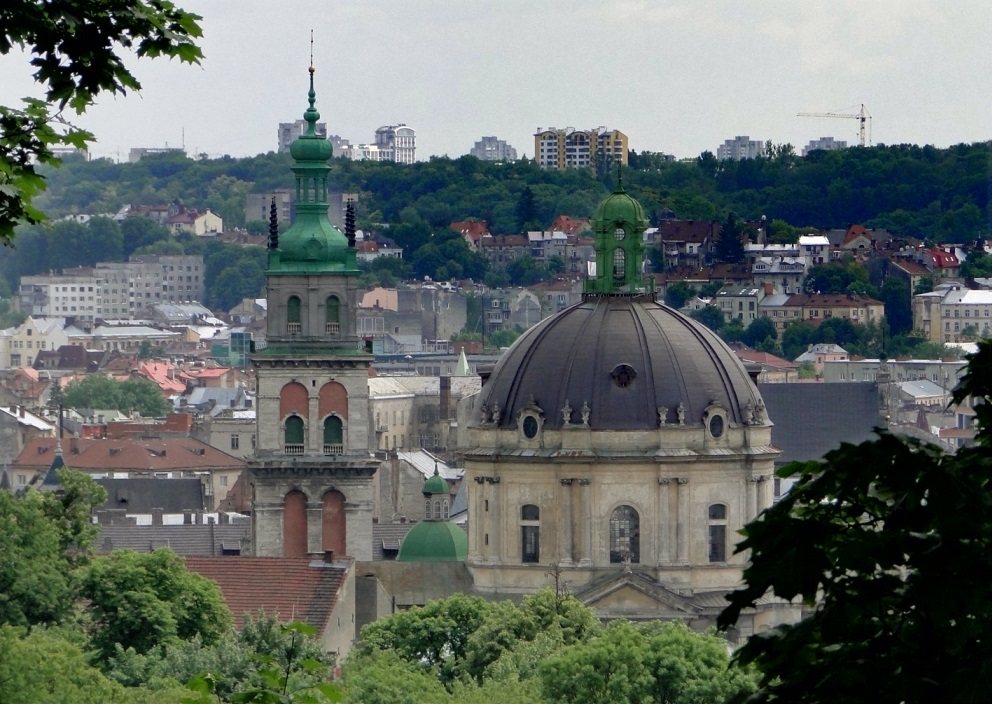}
}\hspace{-0.2cm}
\captionsetup{labelformat=empty}
\end{figure*}

\begin{figure*}
\centering \captionsetup[subfloat]{labelformat=empty}
\subfloat[Hazy image]
{
    \includegraphics[width=6.5cm]{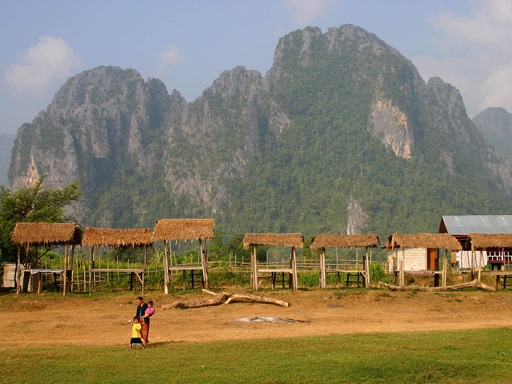}
}\hspace{-0.2cm}
\subfloat[Dehazed Output (Proposed)]
{
    \includegraphics[width=6.5cm]{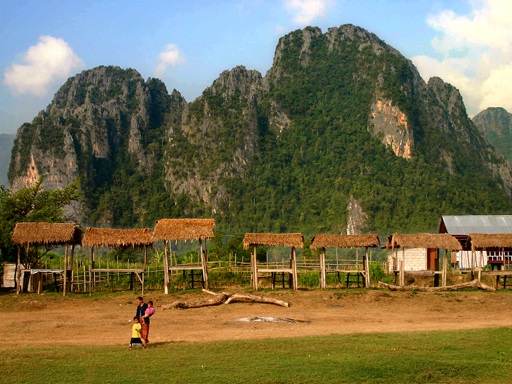}
}\hspace{-0.2cm}
\captionsetup{labelformat=empty}
\end{figure*}

\begin{figure*}
\centering \captionsetup[subfloat]{labelformat=empty}
\subfloat[Hazy image]
{
    \includegraphics[width=6.5cm]{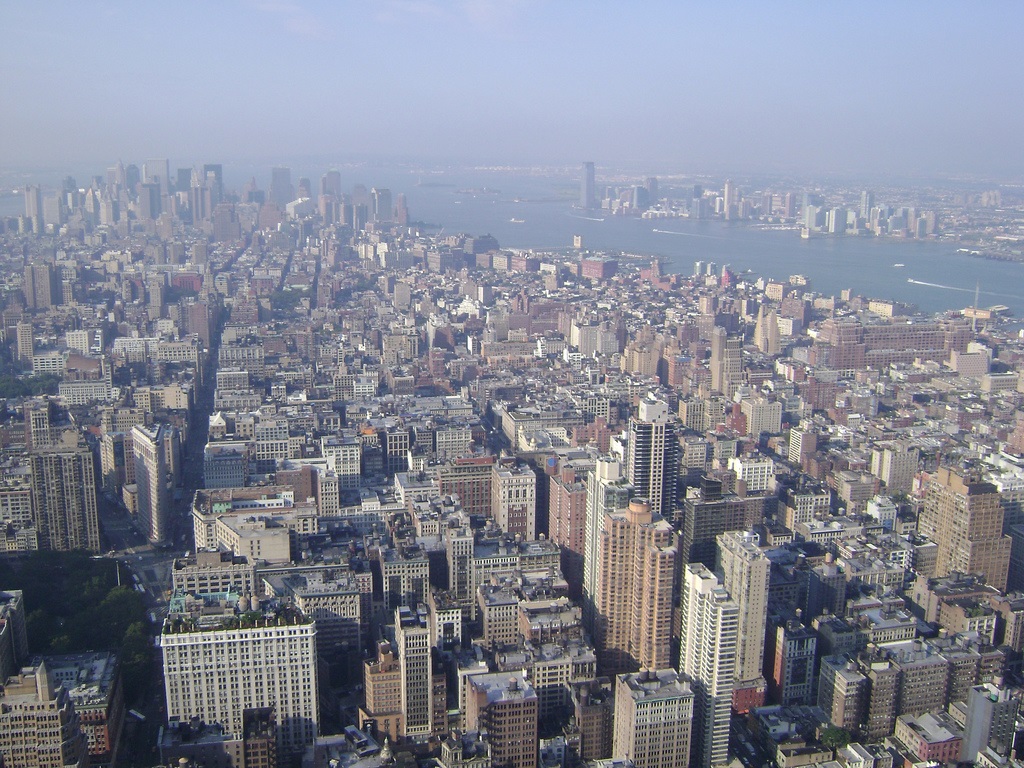}
}\hspace{-0.2cm}
\subfloat[Dehazed Output (Proposed)]
{
    \includegraphics[width=6.5cm]{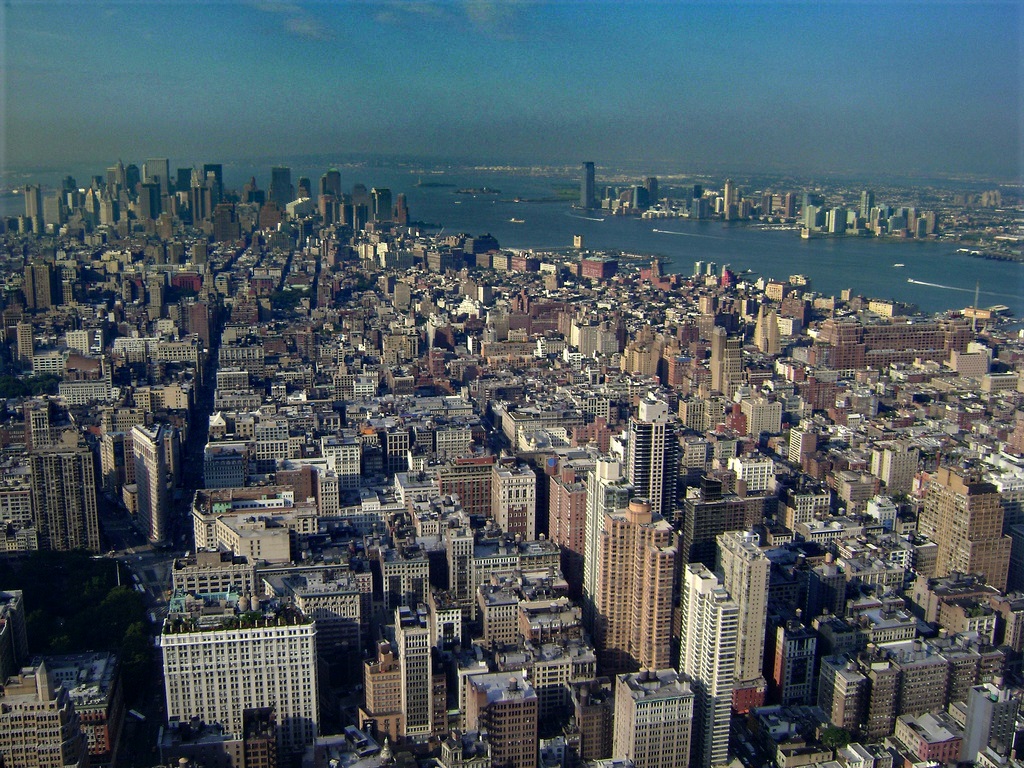}
}\hspace{-0.2cm}
\captionsetup{labelformat=empty}
\end{figure*}

\begin{figure*}
\centering \captionsetup[subfloat]{labelformat=empty}
\subfloat[Hazy image]
{
    \includegraphics[width=6.5cm]{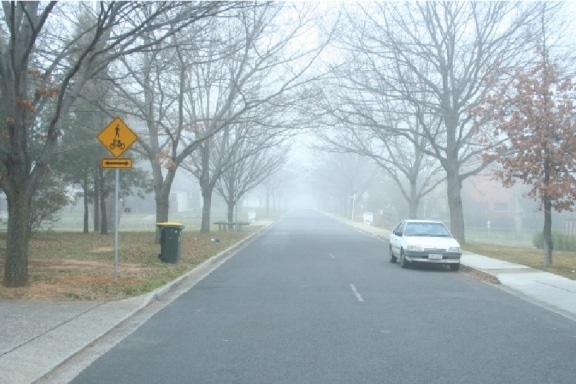}
}\hspace{-0.2cm}
\subfloat[Dehazed Output (Proposed)]
{
    \includegraphics[width=6.5cm]{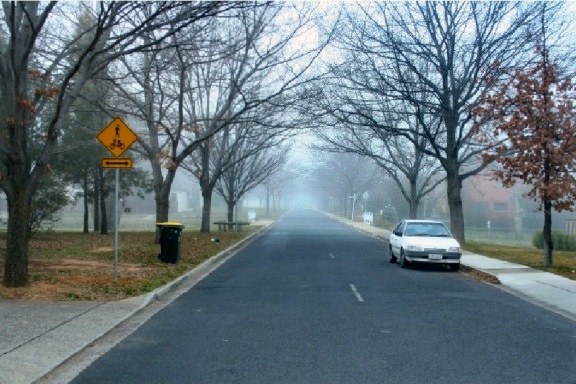}
}\hspace{-0.2cm}
\captionsetup{labelformat=empty}
\end{figure*}

\begin{figure*}
\centering \captionsetup[subfloat]{labelformat=empty}
\subfloat[Hazy image]
{
    \includegraphics[width=6.5cm]{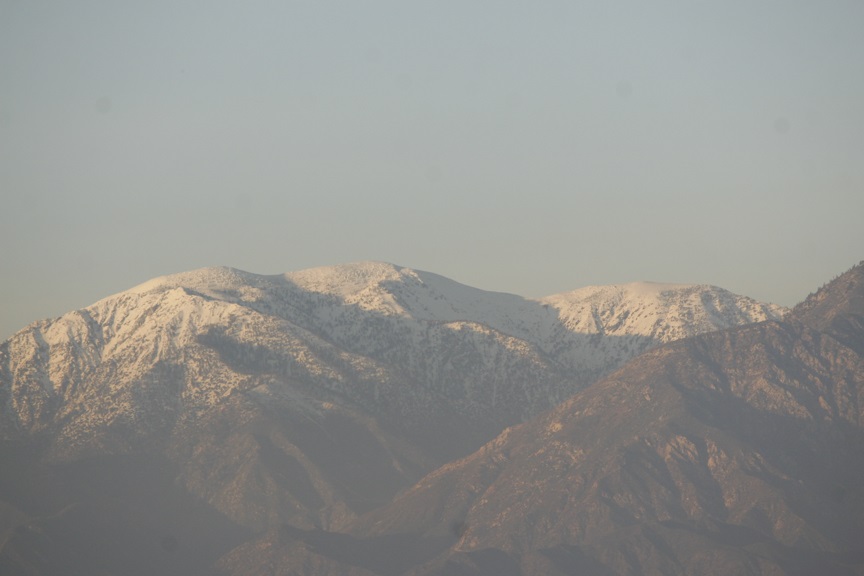}
}\hspace{-0.2cm}
\subfloat[Dehazed Output (Proposed)]
{
    \includegraphics[width=6.5cm]{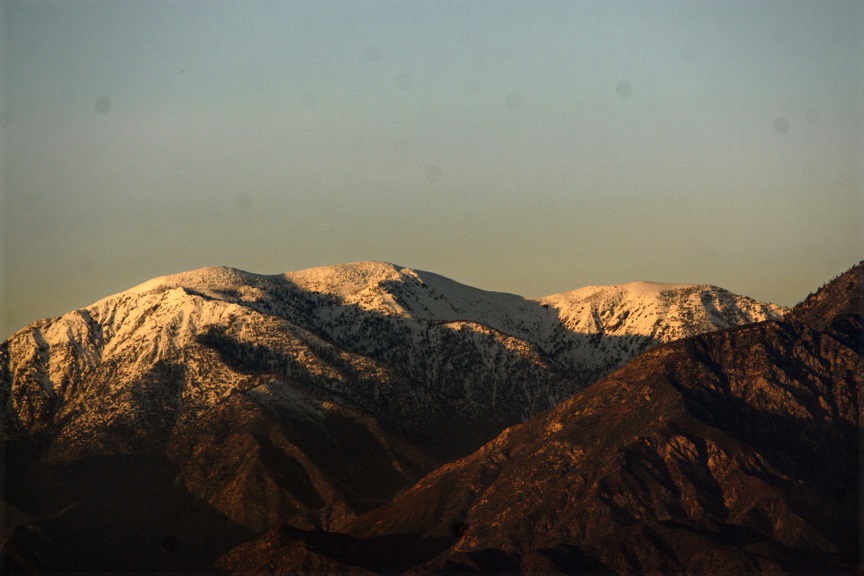}
}\hspace{-0.2cm}
\captionsetup{labelformat=empty}
\end{figure*}

\begin{figure*}
\centering \captionsetup[subfloat]{labelformat=empty}
\subfloat[Hazy image]
{
    \includegraphics[width=6.5cm]{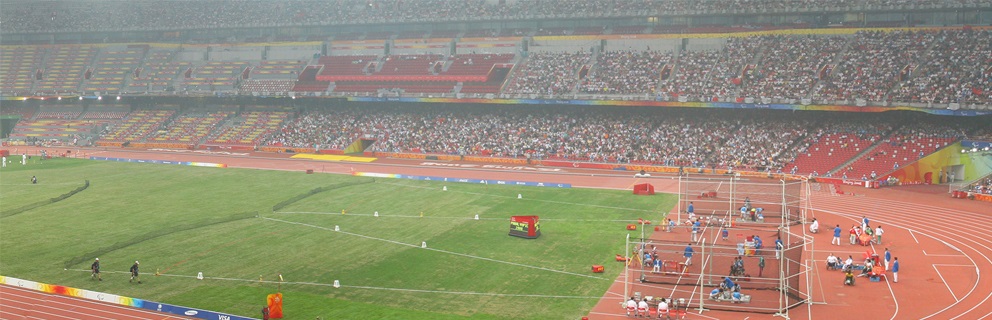}
}\hspace{-0.2cm}
\subfloat[Dehazed Output (Proposed)]
{
    \includegraphics[width=6.5cm]{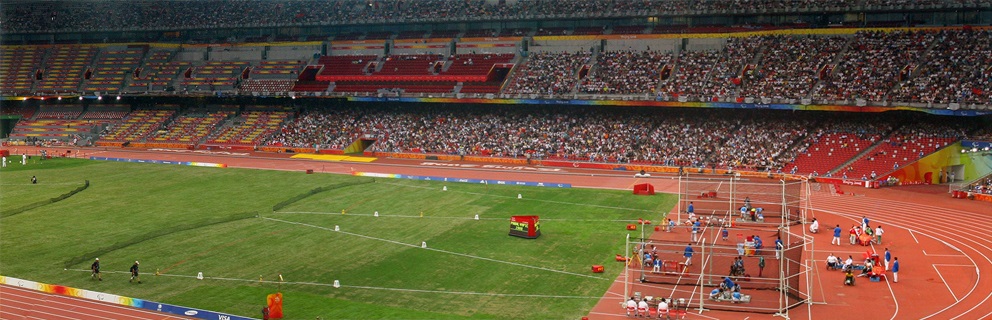}
}\hspace{-0.2cm}
\captionsetup{labelformat=empty}
\end{figure*}

\begin{figure*}
\centering \captionsetup[subfloat]{labelformat=empty}
\subfloat[Hazy image]
{
    \includegraphics[width=6.5cm]{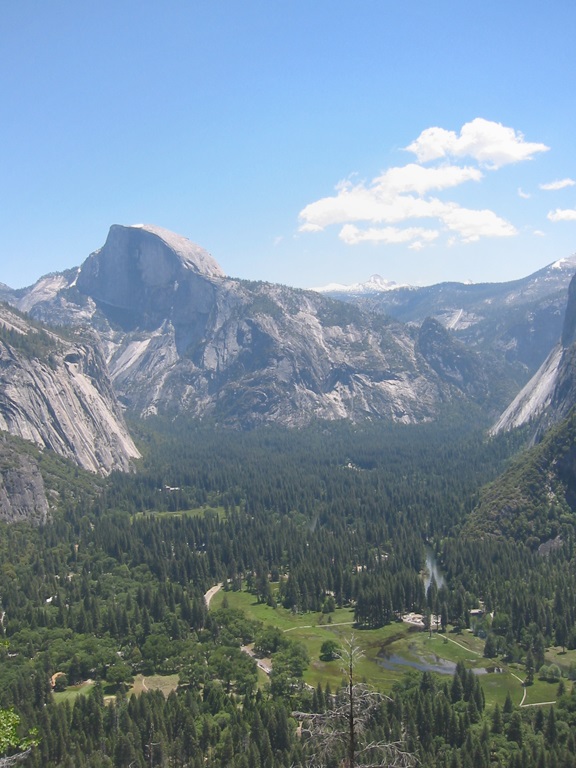}
}\hspace{-0.2cm}
\subfloat[Dehazed Output (Proposed)]
{
    \includegraphics[width=6.5cm]{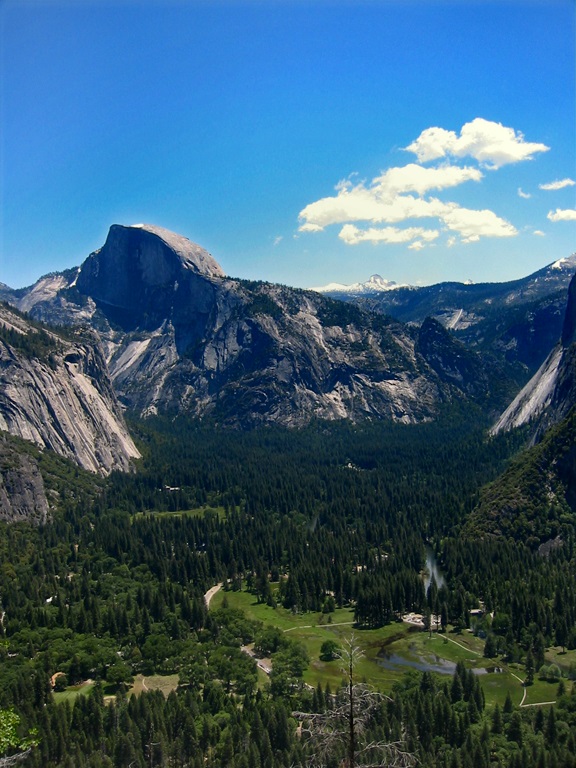}
}\hspace{-0.2cm}
\captionsetup{labelformat=empty}
\end{figure*}

\begin{figure*}
\centering
\subfloat[Hazy image]
{
    \includegraphics[width=6.5cm]{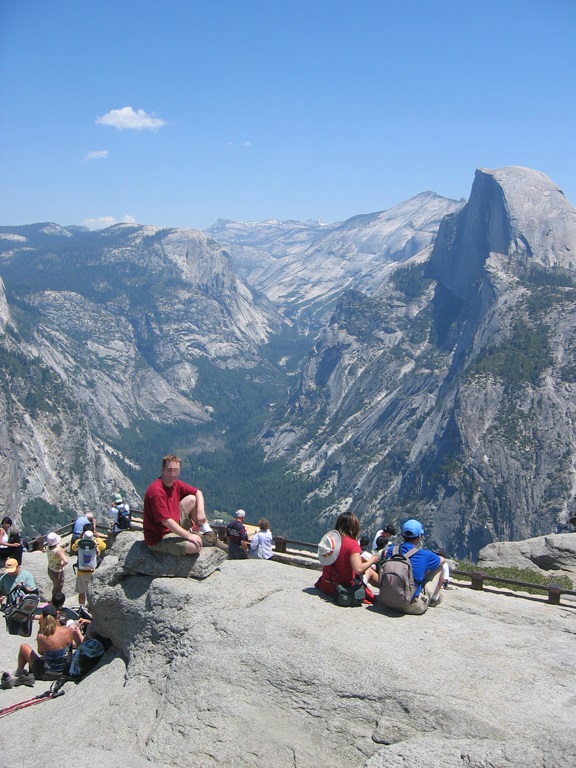}
}\hspace{-0.2cm}
\subfloat[Dehazed Output (Proposed)]
{
    \includegraphics[width=6.5cm]{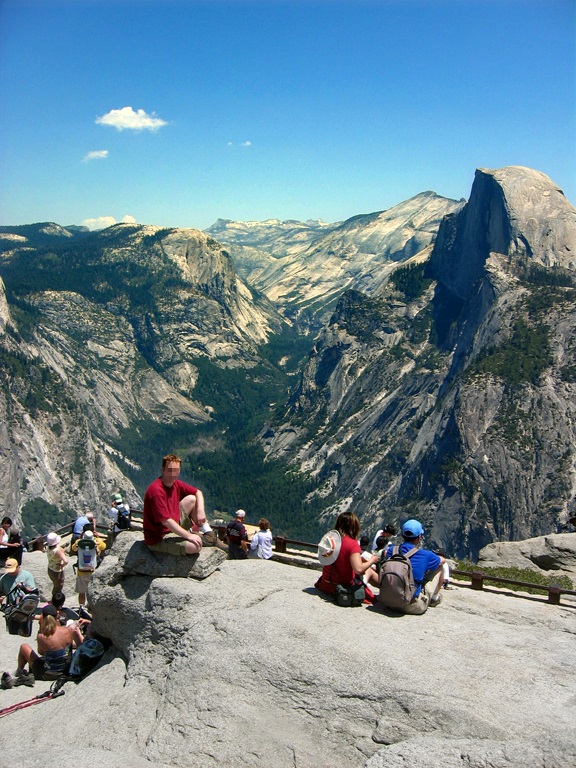}
}\hspace{-0.2cm}
\captionsetup{labelformat=empty}
\end{figure*}

\begin{figure*}
\centering
\subfloat[Hazy image]
{
    \includegraphics[width=6.5cm]{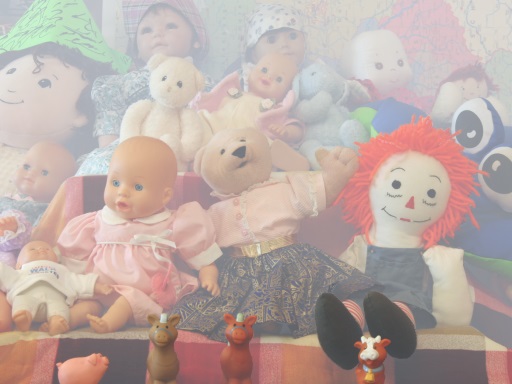}
}\hspace{-0.2cm}
\subfloat[Dehazed Output (Proposed)]
{	
    \includegraphics[width=6.5cm]{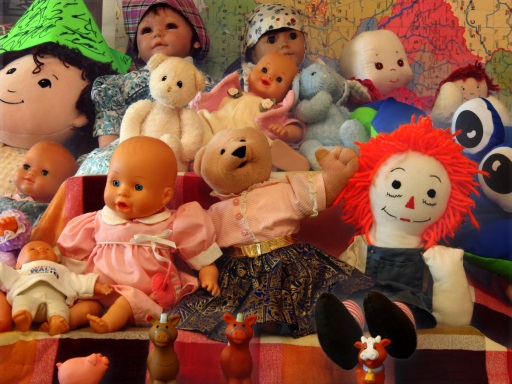}
}\hspace{-0.2cm}
\captionsetup{labelformat=empty}
\end{figure*}

\setcounter{figure}{6}
\begin{figure*}
\begin{center}
\end{center}
\caption{Dehazing performance of PUG-D(R6) on a few more real hazy images from \cite{choi2015referenceless, peng2019image, fattal2014dehazing, li2018benchmarking}.}
\label{figsupp/RealResults}
\end{figure*}

\bibliographystyle{myIEEEtran}
\bibliography{egbib}

\end{document}